\documentclass[letterpaper]{article} 
\usepackage{aaai23}  
\usepackage{times}  
\usepackage{helvet}  
\usepackage{courier}  
\usepackage[hyphens]{url}  
\usepackage{graphicx} 
\usepackage{natbib}  
\usepackage{caption} 
\usepackage{algorithm}
\usepackage{algorithmic}
\usepackage{amsmath,amssymb}
\usepackage{multirow}
\usepackage[misc]{ifsym}
\usepackage{newfloat}
\usepackage{listings}
\DeclareCaptionStyle{ruled}{labelfont=normalfont,labelsep=colon,strut=off} 
\floatstyle{ruled}
\newfloat{listing}{tb}{lst}{}
\floatname{listing}{Listing}
\nocopyright

\title{Symmetry-Aware Transformer-based Mirror Detection}
\author {
    Tianyu Huang\textsuperscript{\rm 1}, Bowen Dong\textsuperscript{\rm 1}, Jiaying Lin\textsuperscript{\rm 2}, Xiaohui Liu\textsuperscript{\rm 1}, Rynson W.H. Lau\textsuperscript{\rm 2}, Wangmeng Zuo\textsuperscript{\rm 1}\Letter
}
\affiliations {
    \textsuperscript{\rm 1} Harbin Institute of Technology, \textsuperscript{\rm 2} City University of Hong Kong\\
    \{tyhuang0428,cndongsky,lxh720199\}@gmail.com, jiayinlin5-c@my.cityu.edu.hk, rynson.lau@cityu.edu.hk, wmzuo@hit.edu.cn
}

\begin{document}

\maketitle

\begin{abstract}
Mirror detection aims to identify the mirror regions in the given input image.
Existing works mainly focus on integrating the semantic features and structural features to mine specific relations between mirror and non-mirror regions, or introducing mirror properties like depth or chirality to help analyze the existence of mirrors.
In this work, we observe that a real object typically forms a loose symmetry relationship with its corresponding reflection in the mirror, which is beneficial in distinguishing mirrors from real objects.
Based on this observation, we propose a dual-path Symmetry-Aware Transformer-based mirror detection Network (SATNet), which includes two novel modules: Symmetry-Aware Attention Module (SAAM) and Contrast and Fusion Decoder Module (CFDM).
Specifically, we first adopt a transformer backbone to model global information aggregation in images, extracting multi-scale features in two paths.
We then feed the high-level dual-path features to SAAMs to capture the symmetry relations.
Finally, we fuse the dual-path features and refine our prediction maps progressively with CFDMs to obtain the final mirror mask.
Experimental results show that SATNet outperforms both RGB and RGB-D mirror detection methods on all available mirror detection datasets.
Codes and trained models are available at \url{https://github.com/tyhuang0428/SATNet}.
\end{abstract}

\section{Introduction}
Mirrors are common objects in the human world, and their presence can affect the performance of a range of vision tasks.
For example, \citeauthor{zendel2017analyzing} propose a list of potential hazards within the CV domain, and the existence of mirrors is one of them.
However, mirror detection can be challenging by using some general detection methods from related tasks, such as salient object detection and semantic segmentation.
As such, it is necessary to treat mirror detection as an independent vision task, and previous works
have managed to tackle this issue from either relation-based frameworks or property-based paradigms.

\begin{figure}
    \centering
    \begin{tabular}{cccccc}
        \includegraphics[width=0.065\textwidth]{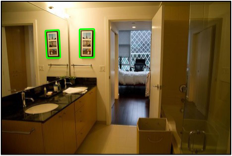} \hspace{-4mm} &
        \includegraphics[width=0.065\textwidth]{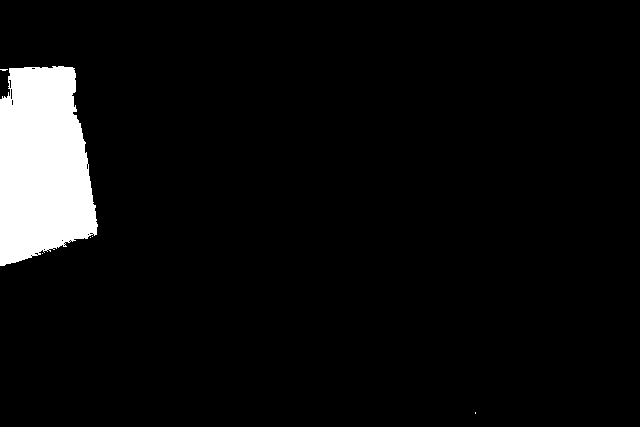} \hspace{-4mm} &
        \includegraphics[width=0.065\textwidth]{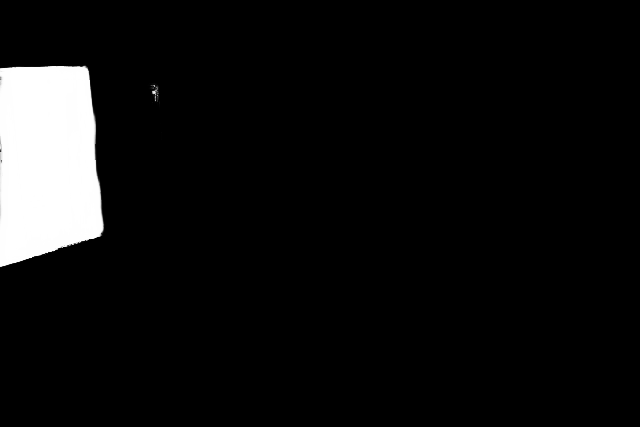} \hspace{-4mm} &
        \includegraphics[width=0.065\textwidth]{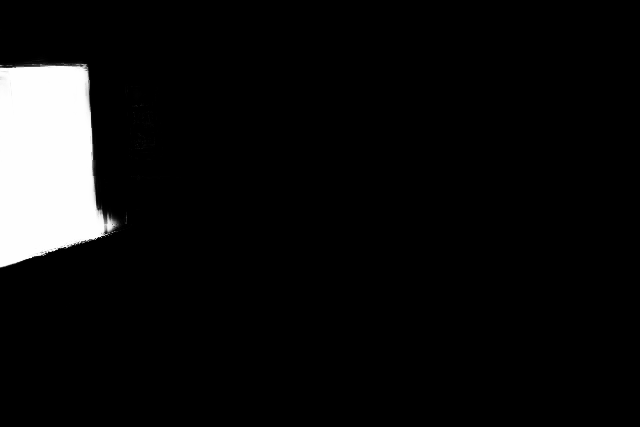} \hspace{-4mm} &
        \includegraphics[width=0.065\textwidth]{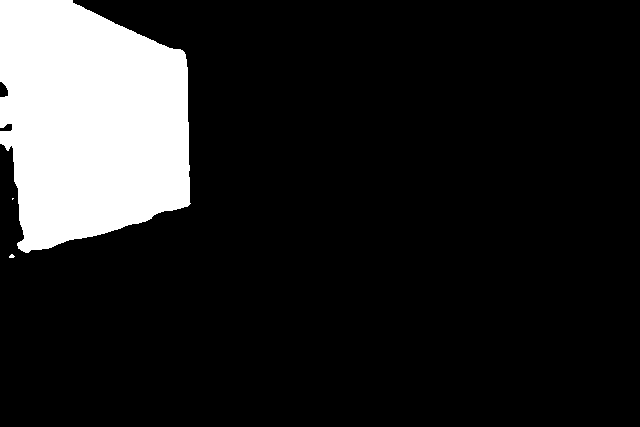} \hspace{-4mm} &
        \includegraphics[width=0.065\textwidth]{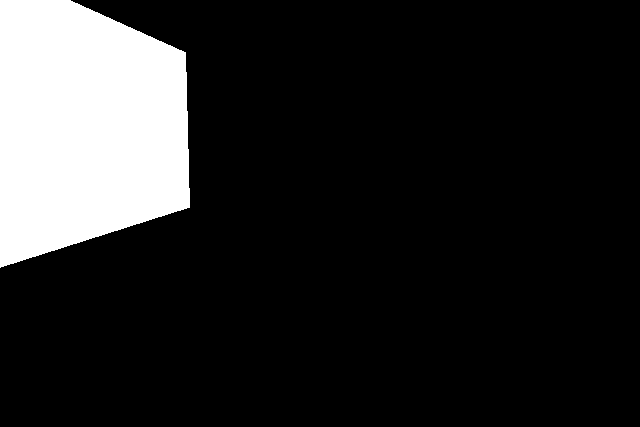} \\
        \includegraphics[width=0.065\textwidth]{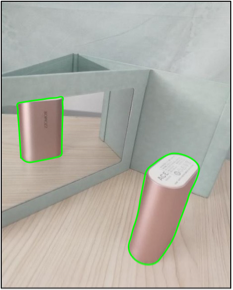} \hspace{-4mm} &
        \includegraphics[width=0.065\textwidth]{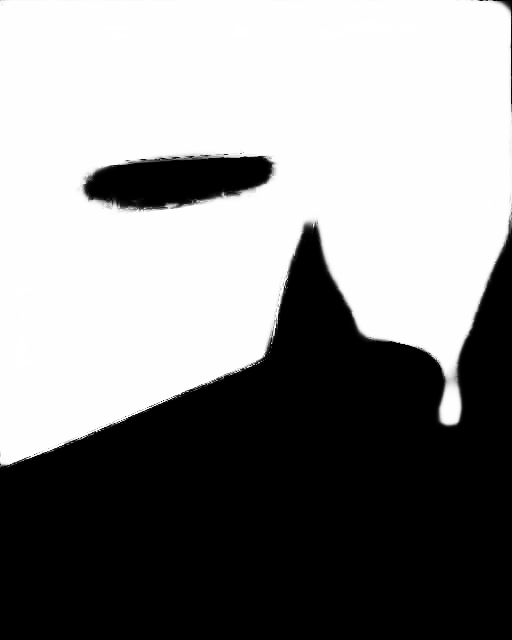} \hspace{-4mm} &
        \includegraphics[width=0.065\textwidth]{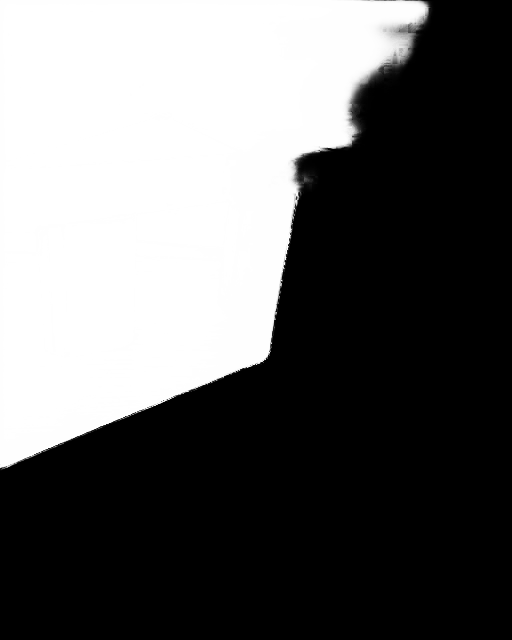} \hspace{-4mm} &
        \includegraphics[width=0.065\textwidth]{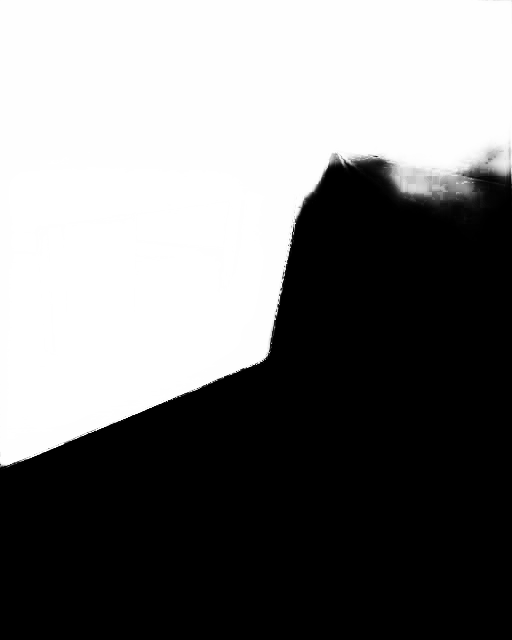} \hspace{-4mm} &
        \includegraphics[width=0.065\textwidth]{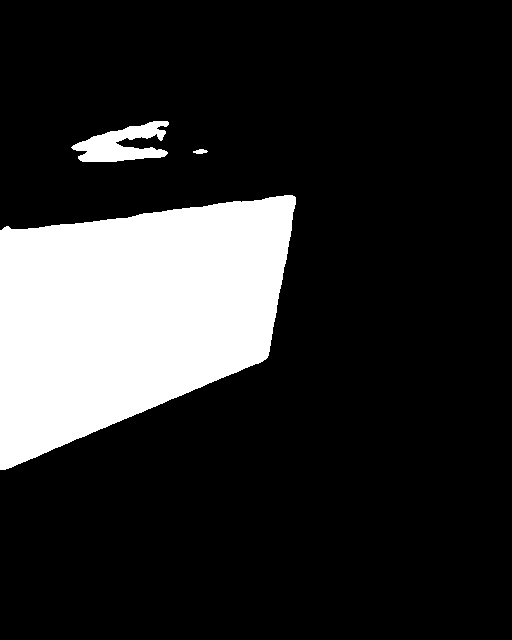} \hspace{-4mm} &
        \includegraphics[width=0.065\textwidth]{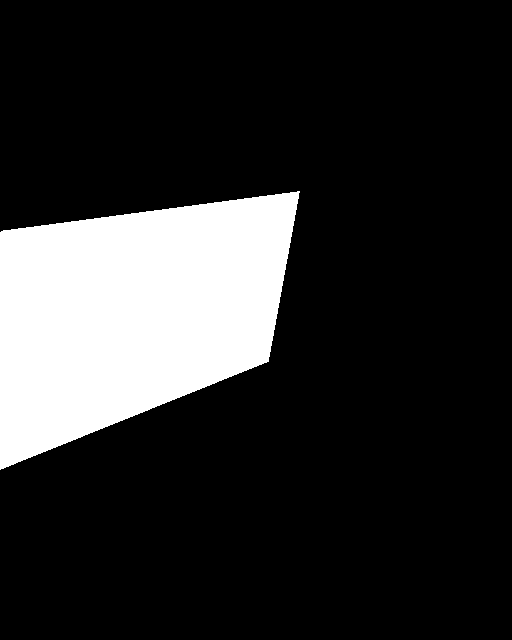} \\
        Image \hspace{-4mm} & MirrorNet \hspace{-4mm} & PMDNet \hspace{-4mm} & SANet \hspace{-4mm} & Ours \hspace{-4mm} & GT
    \end{tabular}
    \caption{Comparison of mirror detection among state-of-the-art methods. MirrorNet~\cite{yang2019my} cannot handle scenes with vague mirror boundaries. Although PMDNet~\cite{lin2020progressive} considers similarity semantics, it can hardly detect the symmetry pair (1st row), and can easily count part of the similar non-mirror regions into mirrors (2nd row). SANet~\cite{guan2022learning} only detects the mirror region above the sink (1st row), and it has worse prediction when semantic associations are lacking (2nd row). By modeling a loose symmetry relationship, SATNet succeeds in both cases.}
    \label{fig:vis1}
\end{figure}

Relations between mirror and non-mirror regions are counted in most mirror detection methods.
\citeauthor{yang2019my} propose to extract contextual discontinuities among regions, but it can only be effective when mirror boundaries are clear against backgrounds.
\citeauthor{lin2020progressive} propose to perceive similarity relationships for contents inside and outside mirrors, which may easily fail when similarities come from non-mirror regions.
\citeauthor{guan2022learning} propose to learn semantic associations in mirror scenes, while such relation quite relies on the environments nearby mirrors. It can only adapt to a few mirror cases, e.g., a mirror above a sink.
Considering mirror properties, \citeauthor{mei2021depth} and \citeauthor{tan2022mirror} regard depth and chirality as additional information for the detection, respectively.
However, these property aggregations only focus on mirror regions, dismissing the environmental semantics related to mirrors.

For a general solution, we need to fully leverage the relationship between mirror and non-mirror regions based on mirror properties.
Considering mirror reflection, symmetry relationships between mirror and non-mirror regions are supposed to be an essential cue for mirror detection. 
In Fig.~\ref{fig:vis1}(1st row), the right half of the mirror would not be missed if the mirror detection model could detect the mirror symmetry relationship of the two paintings.
In Fig.~\ref{fig:vis1}(2nd row), if the model recognizes the left power bank as the mirror region, it can then classify the corresponding real power bank on the right as a non-mirror region. 
However, this symmetry relationship is not a strict mirror symmetry and is highly dependent on the camera viewpoint.
The paintings inside and outside mirrors form a nearly perfect reflection symmetry pair in Fig.~\ref{fig:vis1}(1st row), while the power banks inside and outside mirrors are from different views in Fig.~\ref{fig:vis1}(2nd row). 
We cannot adopt reflection symmetry detection methods directly.
Instead, we observe that real-world object and their reflection in mirrors always maintain the semantic or luminance consistency with each other, even though they may not be strictly symmetric regarding the position or orientation.
%
That is, an object in a mirror should be a mirror reflection of an object in the real world from a certain view. 
We regard this kind of relationship as loose symmetry and aim to explore a new solution to model and leverage this loose symmetry relationship for mirror detection.

Taking loose symmetry into account, we present our Symmetry-Aware Transformer-based mirror detection Network (SATNet).
In particular, we introduce the first transformer baseline in mirror detection, considering the long-range dependencies that loose symmetry requires.
We construct a dual-path network to extract and enhance symmetric features, taking an input image as well as its corresponding horizontally flipped image as inputs. 
For modeling symmetry semantics, we propose a novel Symmetry-Aware Attention Module (SAAM) in high-level dual-path features.
For mirror region segmentation, we propose a novel Contrast and Fusion Decoder Module (CFDM), which constructs a pyramidal decoder to progressively fuse and refine dual-path features. 

To sum up, our main contributions include:
\begin{itemize}
	\item We observe that there are typically loose symmetry relationships between mirror and non-mirror regions. Based on this observation, we propose a novel dual-path Symmetry-Aware Transformer-based mirror detection network (SATNet) to learn symmetry relations for mirror detection. This is the first transformer pipeline in mirror detection.
	\item We present a novel Symmetry-Aware Attention Module (SAAM) to extract high-level symmetry semantics and a novel Contrast and Fusion Decoder Module (CFDM) to refine multi-scale mirror features.
	\item Our network SATNet achieves state-of-the-art results on various mirror detection datasets. Experimental results clearly demonstrate the benefit of loose symmetry relationships for mirror detection.
\end{itemize}

\section{Related Work}
\subsection{Mirror Detection}
The mirror detection task aims to identify the mirror regions of the given input image.
To tackle this problem, several methods attempt to model specific relations between mirror and non-mirror regions.
\citeauthor{yang2019my} proposed the first mirror detection network called MirrorNet, which focuses on perceiving the contrasting features between the contexts inside and outside mirrors.
\citeauthor{lin2020progressive} suggested a progressive mirror detection network PMDNet, designing a relational contextual contrasted local module to extract similarity features.
\citeauthor{guan2022learning} proposed to learn semantic association in mirror scenes, which may imply the existence of mirrors.
However, those methods can hardly adapt to general mirror detection cases as the relations they match are either too simple or too strict.
Recent methods take mirror properties into account.
\citeauthor{mei2021depth} introduced depth information to mirror detection as the depth value in mirror regions is irregular.
In contrast, the depth input is unreliable, and the method can be easily misled by depth.
\citeauthor{tan2022mirror} proposed a dense visual chirality discriminator to judge the possible mirror existence, while the improvement is limited since chirality information tends to be subtle when mirror contents are clean.
The leverage of mirror properties in these methods mainly depends on the semantics of mirror regions, dismissing the interaction with non-mirror regions.
Unlike existing works, we aim to utilize loose symmetry relationships between real-world objects and corresponding mirror regions to enhance the overall detection ability.

\subsection{Reflection Symmetry Detection}
Reflection symmetry detection aims to detect symmetry axes in given images. Early works in this task can be divided into two categories: keypoint matching detection and dense heatmap detection. 
\citeauthor{loy2006detecting} adopted SIFT~\cite{lowe2004distinctive} to compute matched keypoints, and generated potential symmetry axes accordingly. 
\citeauthor{cornelius2006detecting} took a single matching pair for hypothesizing with the local affine frame. 
For dense heatmap, \citeauthor{tsogkas2012learning} utilized pixel-level features to predict the symmetry area densely. 
\citeauthor{funk2017beyond} employed CNNs to extract the symmetric features directly. 
Recently, \citeauthor{Seo_2021_ICCV} proposed a novel polar matching convolution to encode the similarities among pixels. 
Contrary to the strict reflection symmetry, symmetry relationships in mirror cases are loosely defined. Therefore, reflection symmetry detection methods cannot be directly employed in mirror detection. To tackle this, we propose a dual-path  Transformer-based structure with attention mechanisms in high-level features to model the loose symmetry relationships.

\subsection{Salient Object Detection}
Salient object detection (SOD) aims to detect and segment the most distinct object in an input image.
Existing methods in RGB SOD are mainly based on the UNet structure~\cite{ronneberger2015u}, like~\cite{Wang_2017_ICCV,Pang_2020_CVPR}.
\citeauthor{deng2018r3net} adopt a recurrent network to refine the salient map progressively.
\citeauthor{Liu_2018_CVPR} adopt attention mechanisms to learn more dependencies among features.
Recently, RGB-D SOD has received considerable attention.
Several methods~\cite{song2017depth} treat depth as an additional dimension of the input features, while the others~\cite{fan2020rethinking} separately extract RGB and depth features and fuse them in the decoding process.
\citeauthor{Liu_2020_CVPR} proposed to fuse depth information with attention mechanisms.
\citeauthor{pang2020hierarchical} integrated RGB and depth through densely connected structures.
\citeauthor{Liu_2021_ICCV} proposed a vision transformer network, rethinking this field from an aspect of sequence-to-sequence architectures.
Albeit similar to mirror detection, SOD methods can hardly have a good performance on the mirror detection task as mirrors are not salient enough to detect in most cases.
SOD methods may wrongly detect some conspicuous objects inside mirrors.

\begin{figure*}[t]
  \centering
  \includegraphics[width=0.95\textwidth]{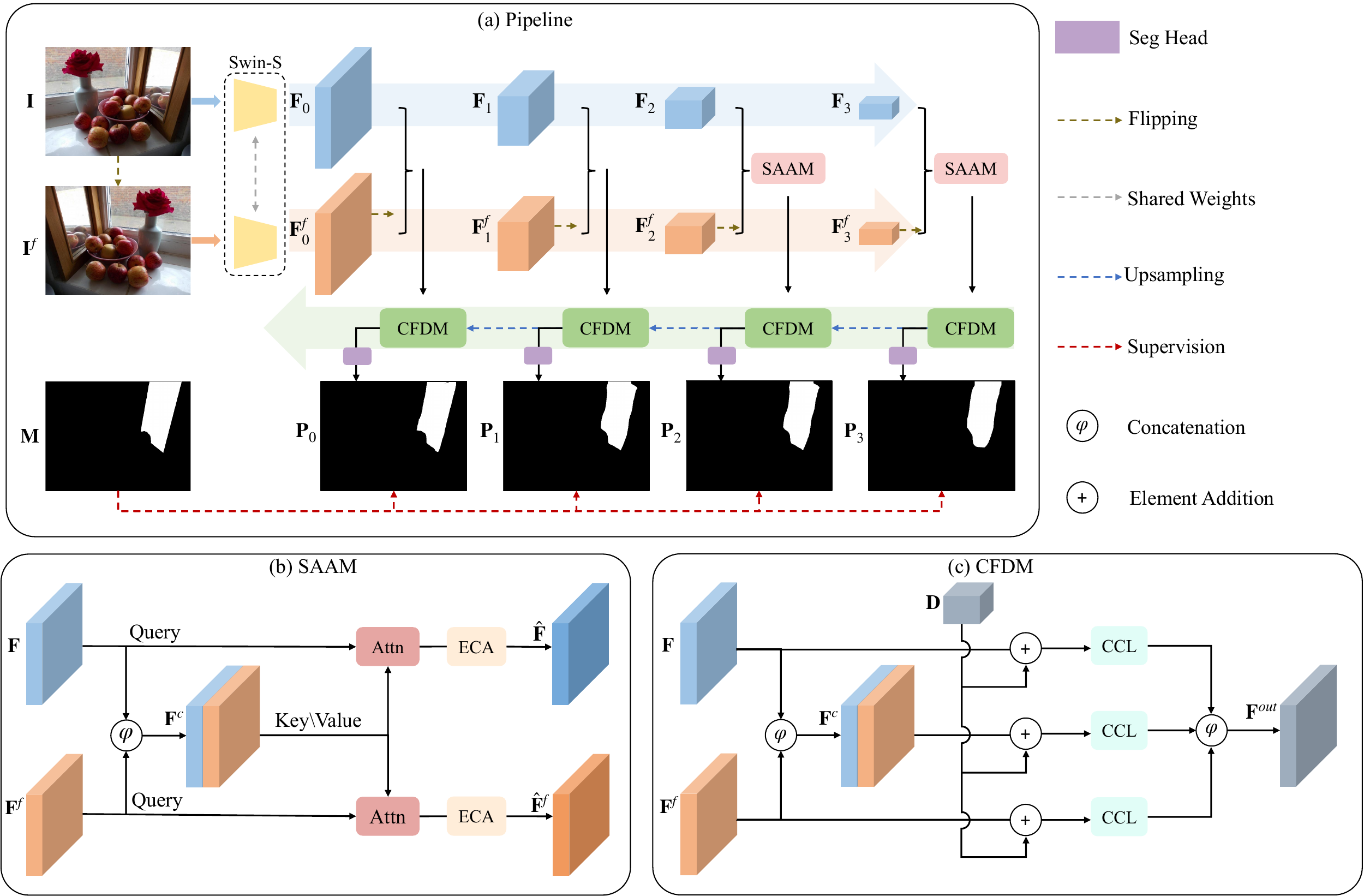}
  \caption{(a) Pipeline of our SATNet. 
(b) Symmetry-Aware Attention Module. (c) Contrast and FusionDecoder Module.}
  \label{fig:overview}
\end{figure*}

\section{Method}
Based on the idea of loose symmetry relationships, we propose a dual-path Symmetry-Aware Transformer-based network for mirror detection. 
Loose symmetry relationships can assist the detecting process in two aspects: the presences of loose symmetry relationships imply the possible existence of mirrors; differences between symmetric pairs indicate which part belongs to mirror regions. The dual-path structure and our novel Symmetry-Aware Attention Module are designed for the first aspect. Additionally, to better encode the symmetry features as well as recognize the corresponding mirror semantics, a transformer backbone and our Contrast and Fusion Decoder Module are proposed for the second aspect.

\subsection{Overview}
Fig.~\ref{fig:overview}(a) illustrates the pipeline of our SATNet.
Given an input image $\mathbf{I}$ as well as its flipped image $\mathbf{I}^{f}$, we feed them into a shared-weights transformer backbone to obtain multi-scale features $\{\mathbf{F}_{0},...,\mathbf{F}_{3}\}$ and corresponding flipped features $\{\mathbf{F}^{f}_{0},...,\mathbf{F}^{f}_{3}\}$, respectively.
For modeling symmetry relations, we select features from the highest two levels of both paths, and feed features in the same level into our Symmetry-Aware Attention Module (SAAM), fetching joint features $\mathbf{\hat{F}}$ and $\mathbf{\hat{F}}^{f}$.
Then, the multi-scale features $\{\mathbf{F}_{i} / \mathbf{\hat{F}}_{i}\}$ as well as the flipped features $\{\mathbf{F}^{f}_{i} / \mathbf{\hat{F}}^{f}_{i}\}$ are fed into corresponding Contrast and Fusion Decoder Module (CFDM), generating coarse output features $\mathbf{F}^{out}_{i}$ with different scales progressively.
For each $\mathbf{F}^{out}_{i}$ (except $\mathbf{F}^{out}_{0}$), we upsample it into the next decoder as the reference features $\mathbf{D}_{i-1}$ for further prediction refinement.
Meanwhile, we get the prediction map $\mathbf{P}_{i}$ in each decoder through a segmentation head and supervise it via the ground-truth mask $\mathbf{M}$.
Finally, our prediction result $\tilde{\mathbf{M}}$ is generated by the last decoder module.

\subsection{Dual-Path Structure}
In most cases, loose symmetry relationships are implicit under complex semantics. 
Such a relationship is too hidden to be perceived by existing baselines, which has been verified in our ablation study. 
To better perceive the relationship, dual enhancements are suggested. As a common method of data augmentation, horizontal flipping can modify the global semantics of natural images, while symmetry relationships still exist (they just display in the opposite direction). Thus, we introduce a dual-path network to extract symmetric features: Given $\mathbf{F}$ and $\mathbf{F}^{f}$ from both paths, we expect they differ from each other but have features of the same loose symmetry relationship.
When we concatenate them together as $\mathbf{F}^{c}$, symmetry semantics in the symmetric region can be enhanced. To extract the same symmetric features, input images $\mathbf{I}$ and $\mathbf{I}^f$ must be fed into the same backbone.
Our fusion function is defined as follows:
\begin{gather}\label{eq1}
    \varphi(\mathbf{a_1}, ..., \mathbf{a_n}) = \sigma(BN(\psi_{3\times3}(\psi_{1\times1}([\mathbf{a_1}, ..., \mathbf{a_n}])))), \\
    \label{eq2}
    \mathbf{F}^{c} = \varphi(\mathbf{F}, flip(\mathbf{F}^{f})),
\end{gather}
where $[\cdot,...,\cdot]$ denotes the concatenation operation on the channel dimension. $\psi_{w\times w}$ is a $w\times w$ convolution, BN denotes the Batch Normalization, $\sigma$ is the ReLU activation function, and flip is the horizontal flipping.
To align features in the spatial level during concatenation, we flip $\mathbf{F}^{f}$ back before SAAM or CFDM.

\subsection{Symmetry-Aware Attention Module}
Fig.~\ref{fig:overview}(b) shows the architecture of our symmetry-aware attention module.
With SAAM, we aim to perceive loose symmetry relationships in an image that indicates the possible existence of mirrors.
To this end, we use the attention mechanisms (i) to enhance the feature $\mathbf{F}$ of the input image as well as (ii) to obtain the symmetry-aware feature by modeling the dependency between the input and its flipped images.  
In general, the attention mechanism can model the dependencies among each position in a global manner~\cite{vaswani2017attention}, which can be formulated as,
\begin{equation}
\label{eq3}
	Attention(\mathbf{Q}, \mathbf{K}, \mathbf{V}) = Softmax(\frac{\mathbf{Q} \mathbf{K}^{T}}{\sqrt{d_k}})\mathbf{V},
\end{equation}
where $\mathbf{Q}$, $\mathbf{K}$ and $\mathbf{V}$ denote Query, Key, and Value, respectively.

Our SAAM takes both $\mathbf{F}^{c}$ as well as $\mathbf{F}$ and $\mathbf{F}^{f}$ as the input. 
Among them, $\mathbf{F}^{c}$ aggregates the features from $\mathbf{F}$ and $\mathbf{F}^{f}$ from both paths and is spatially consistent with $\mathbf{F}$, and thus can be treated as an augmented representation of $\mathbf{F}$. 
To exploit the attention to enhance the feature $\mathbf{F}$, we treat $\mathbf{F}$ as query and $\mathbf{F}^{c}$ as key and value, and further apply channel transformation with Efficient Channel Attention (ECA)~\cite{Wang_2020_CVPR} module right after the attention module to obtain the enhanced feature $\mathbf{\hat{F}}$,
\begin{gather}\label{eq4}
\begin{split}
	\mathbf{\hat{F}} = ECA(Attention(\mathbf{F}, \mathbf{F}^{c}, \mathbf{F}^{c})),
\end{split}
\end{gather}
where $ECA(\cdot)$ denotes the Efficient Channel Attention module.
To obtain the symmetry-aware feature, we treat $\mathbf{F}^{f}$ as query and $\mathbf{F}^{c}$ as key and value.
Note that $\mathbf{F}^{f}$ is extracted from the flipped image, and $\mathbf{F}^{c}$ is spatially consistent with the input image. 
Their similarity score can thus be treated as an indicator of loose mirror symmetry between parts from the input and its flipped images.
And the output of the attention module can then be regarded to be symmetry-aware.
Analogous to $\mathbf{\hat{F}}$, the symmetry-aware feature is obtained by,
\begin{gather}
\begin{split}
	\mathbf{\hat{F}}^{f} &= ECA(Attention(\mathbf{F}^{f}, \mathbf{F}^{c}, \mathbf{F}^{c})).
\end{split}
\end{gather}

\subsection{Contrast and Fusion Decoder Module}
%
Since MirrorNet~\cite{yang2019my}, Context Contrasted Local (CCL) decoder~\cite{Ding_2018_CVPR} has been widely adopted in mirror detection networks.
To better refine the prediction, edge extractors are joint as an extra supervision to previous methods~\cite{lin2020progressive,tan2022mirror} as well.
%
In this subsection, we further extend the CCL module to present our CFDM for handling multiple features.
With no edge information, our CFDM can outline precise mirror boundaries efficiently by refining multi-level features progressively in a top-down structure.

As shown in Fig.~\ref{fig:overview}(c), our CFDM takes $\mathbf{{F}}_i$ and $\mathbf{{F}}^f_i$ as the input when $i = 0, 1$, and $\mathbf{\hat{F}}_i$ and $\mathbf{\hat{F}}^f_i$ when $i = 2, 3$.
Without loss of generality, we use $\mathbf{{F}}_i$ and $\mathbf{{F}}^f_i$ as an example to explain the CFDM module. 
To begin with, we use Eq.~(\ref{eq2}) to obtain the fused feature $\mathbf{F}^c_i$. 
Denote by $\mathbf{F}^{out}_{i+1}$ the $(i+1)$-scale CFDM output. 
We then upsample $\mathbf{F}^{out}_{i+1}$ to obtain the higher-level feature map,
\begin{gather}
\mathbf{D}_{i} = U_2(\sigma(BN(\psi_{3 \times 3}(\mathbf{F}^{out}_{i+1}))),
\end{gather}
where $U_2$ denotes the bilinear upsampling operation.
Subsequently, the reference features for $(\mathbf{F}^{c}_{i}, \mathbf{F}_{i}, \mathbf{F}^{f}_{i})$ can be given by,
\begin{gather}
	\label{eq:add}
	(\mathbf{\tilde{F}}^{c}_{i}, \mathbf{\tilde{F}}_{i}, \mathbf{\tilde{F}}^{f}_{i}) = \begin{cases}
		(\mathbf{F}^{c}_{i}, \mathbf{F}_{i}, \mathbf{F}^{f}_{i}) \oplus (\mathbf{D}_{i}, \mathbf{D}_{i}, \mathbf{D}_{i}), & i < 3 \\
		(\mathbf{F}^{c}_{i}, \mathbf{F}_{i}, \mathbf{F}^{f}_{i}), & i = 3 \\
	\end{cases} 
\end{gather}
where $\oplus$ denotes the element-wise summation operator.

The three feature maps $\mathbf{\tilde{F}}^{c}_{i}, \mathbf{\tilde{F}}_{i}, \mathbf{\tilde{F}}^{f}_{i}$ are separately fed into the CCL module to extract contrastive semantics. Here we use $\mathbf{\tilde{F}}_{i}$ as an example, 
\begin{equation}
	CCL(\mathbf{\tilde{F}}_{i}) = \sigma(BN(f_l({\mathbf{\tilde{F}}_{i}})-f_{ct}({\mathbf{\tilde{F}}_{i}}))),
\end{equation}
where $f_l$ is the local feature extractor which contains a $3 \times 3$ convolution with a dilation rate of 1, BN, and ReLU in turn. 
Considering the changes on the receptive field, we set dilation rates to \{8, 6, 4, 2\} for layer~\{0, 1, 2, 3\}, respectively.
Finally, we concatenate those three CCL outputs together to get the output features $\mathbf{F}^{out}_{i}$ and the corresponding prediction map $\mathbf{P_i}$, which is given as follows,
\begin{gather}
\label{eq:prediction}
	\mathbf{F}^{out}_{i} = \varphi(CCL(\mathbf{\tilde{F}}^{c}_{i}), CCL(\mathbf{\tilde{F}}_{i}), CCL(\mathbf{\tilde{F}}^{f}_{i})), \\
	\label{eq:prediction2}\mathbf{P}_{i} = f_{seg}(\mathbf{F}^{out}_{i}),
\end{gather}
where $f_{seg}$ is a segmentation head whose output has two channels.
And the output of the last decoder layer $\mathbf{P}_{0}$ is adopted as the final prediction result $\tilde{\mathbf{M}}$ of our network.
%

\subsection{Transformer for Mirror Detection}
As for the feature extraction, loose symmetry is typically a long-range relationship,
which means our network needs a large receptive field to perceive it.
CNN-based methods utilize a couple of convolution kernels to fulfill local features aggregation.
However, the convolution with a small kernel size cannot construct global feature aggregation directly, which restricts the feature representation ability of those methods in complex scenarios.
In contrast, the self-attention module in transformers can model the long-range interaction explicitly, making vision transformer very competitive in several complex scene understanding tasks~\cite{Zheng_2021_CVPR}.
Swin Transformer~\cite{liu2021swin} proposes regular and shifted window self-attention modules to construct local and global feature aggregation with limited computation complexity while achieving state-of-the-art performance in scene parsing.
Thus, we adopt a transformer pipeline in mirror detection based on Swin Transformer.

\subsection{Loss Function}
%
Our learning objective is defined by considering all scales.
For each prediction map $\mathbf{P}_{i}$, we calculate the cross-entropy (CE) loss~\cite{de2005tutorial} between $\mathbf{P}_{i}$ and the ground-truth $\mathbf{M}$. 
The overall loss function $\mathcal{L}$ is then given as the summation of CE loss for each prediction map, 
\begin{equation}
	\mathcal{L} = \sum_{i=0}^{3}w_{i}\mathcal{L}_{\text{ce}}(\mathbf{P}_{i}, \mathbf{M}),
	\label{eq:loss}
\end{equation}
where $w_i$ is the corresponding weight for the $i$-th layer.
We empirically set the weight $w_i$ as [1.25, 1.25, 1.0, 1.5] according to the experimental results.

\section{Experiments}
\subsection{Datasets and Evaluation Metrics}
Following previous works~\cite{yang2019my,lin2020progressive}, we use Mirror Segmentation Dataset (MSD) and Progressive Mirror Dataset (PMD) to evaluate our method.
Besides, we adopt an RGB-D dataset RGBD-Mirror to make a comparison with the state-of-the-art RGB-D mirror detection method PDNet~\cite{mei2021depth}.

To assess mirror detection performance, we adopt three commonly used dense prediction evaluation metrics: intersection over union (IoU), F-measure $F_\beta$, and mean absolute error (MAE).
Please refer to the appendix for more details.

\subsection{Implementation Details}
We implement our network on PyTorch~\cite{paszke2019pytorch} and use the small version of Swin Transformer (namely Swin-S) pretrained on ImageNet-1k~\cite{deng2009imagenet} as the backbone of our network.
Note that dual-path features are fed into the same backbone and share the weights.
Following data augmentation methods adopted by previous works, we adopt random resize and crop as well as random horizontal flipping to augment training images. 
And for testing, we simply resize input images to $512 \times 512$ to evaluate our network. 
Our network is trained on 8 Tesla V100 GPUs with 2 images per GPU for 20K iterations.
During training, we use ADAM weight decay optimizer and set $\beta_1$, $\beta_2$, and the weight decay to 0.9, 0.999, and 0.01, respectively.
The learning rate is initialized to $6 \times 10^{-4}$ and decayed by the \emph{poly} strategy with the power of $1.0$.
It takes 6 hours to train our network, and testing on a single GPU needs 0.08s per image.

\subsection{Comparison with State-of-the-arts}

\begin{table}[t]
  \caption{Quantitative results of the state-of-the-art methods on MSD dataset and PMD dataset. Our method achieves the best performance in terms of all the evaluation metrics.}
  \centering
  \begin{tabular}{l|ccc|ccc}
	\hline
	\multicolumn{1}{l|}{\multirow{2}{*}{Method}} & \multicolumn{3}{c|}{MSD} & \multicolumn{3}{c}{PMD} \\
	\cline{2-7}
	\multicolumn{1}{l|}{} & $IoU\uparrow$ \hspace{-4mm} & $F_\beta\uparrow$ \hspace{-4mm} & $MAE\downarrow$ \hspace{-3mm} & $IoU\uparrow$ \hspace{-4mm} & $F_\beta\uparrow$ \hspace{-4mm} & $MAE\downarrow$ \\
	\hline
	CPDNet & 57.58 \hspace{-4mm} & 0.743 \hspace{-4mm} & 0.115 \hspace{-3mm} & 60.04 \hspace{-4mm} & 0.733 \hspace{-4mm} & 0.041 \\ 
	MINet & 66.39 \hspace{-4mm} & 0.823 \hspace{-4mm} & 0.087 \hspace{-3mm} & 60.83 \hspace{-4mm} & 0.798 \hspace{-4mm} & 0.037 \\
	LDF & 72.88 \hspace{-4mm} & 0.843 \hspace{-4mm} & 0.068 \hspace{-3mm} & 63.31 \hspace{-4mm} & 0.796 \hspace{-4mm} & 0.037 \\
	VST & 79.09 \hspace{-4mm} & 0.867 \hspace{-4mm} & 0.052 \hspace{-3mm} & 59.06 \hspace{-4mm} & 0.769 \hspace{-4mm} & 0.035 \\
	\hline
	MirrorNet & 78.88 \hspace{-4mm} & 0.856 \hspace{-4mm} & 0.066 \hspace{-3mm} & 58.51 \hspace{-4mm} & 0.741 \hspace{-4mm} & 0.043 \\
	PMDNet & 81.54 \hspace{-4mm} & 0.892 \hspace{-4mm} & 0.047 \hspace{-3mm} & 66.05 \hspace{-4mm} & 0.792 \hspace{-4mm} & 0.032 \\
	SANet & 79.85 \hspace{-4mm} & 0.879 \hspace{-4mm} & 0.054 \hspace{-3mm} & 66.84 \hspace{-4mm} & 0.837 \hspace{-4mm} & 0.032 \\
	VCNet & 80.08 \hspace{-4mm} & 0.898 \hspace{-4mm} & 0.044 \hspace{-3mm} & 64.02 \hspace{-4mm} & 0.815 \hspace{-4mm} & 0.028 \\ 
	\hline
	Ours & \textbf{85.41} \hspace{-4mm} & \textbf{0.922} \hspace{-4mm} & \textbf{0.033} \hspace{-3mm} & \textbf{69.38} \hspace{-4mm} & \textbf{0.847} \hspace{-4mm} & \textbf{0.025} \\
	\hline
  \end{tabular}
  \label{tab:msd&pmd}
\end{table}

\begin{table}[t]
  \caption{Quantitative results of the state-of-the-art methods on RGBD-Mirror dataset. w/ Depth denotes the usage of depth information in a corresponding method. Our method outperforms all the competing methods, even though we do not use depth information.}
  \centering
  \begin{tabular}{l|c|ccc}
	\hline
	\multicolumn{1}{l|}{\multirow{2}{*}{Method}} & \multicolumn{1}{c|}{\multirow{2}{*}{w/ Depth}} & \multicolumn{3}{c}{RGBD-Mirror} \\
	\cline{3-5}
	\multicolumn{1}{l|}{} & \multicolumn{1}{l|}{} & $IoU\uparrow$ & $F_\beta\uparrow$ & $MAE\downarrow$ \\
	\hline
	JL-DCF & \checkmark & 69.65 & 0.844 & 0.056 \\
	DANet & \checkmark & 67.81 & 0.835 & 0.060 \\
	BBSNet & \checkmark & 74.33 & 0.868 & 0.046 \\
	VST & \checkmark & 70.20 & 0.851 & 0.052 \\
	\hline
	PDNet & & 73.57 & - & 0.053 \\
	PDNet & \checkmark & 77.77 & 0.878 & 0.041 \\
	SANet & & 74.99 & 0.873 & 0.048 \\
	VCNet & & 73.01 & 0.849 & 0.052 \\
	\hline
	Ours & & \textbf{78.42} & \textbf{0.906} & \textbf{0.031} \\
	\hline
  \end{tabular}
  \label{tab:rgbd}
\end{table}

\begin{figure*}
    \centering
    \begin{tabular}{ccccccccccc}
        \includegraphics[width=0.08\textwidth]{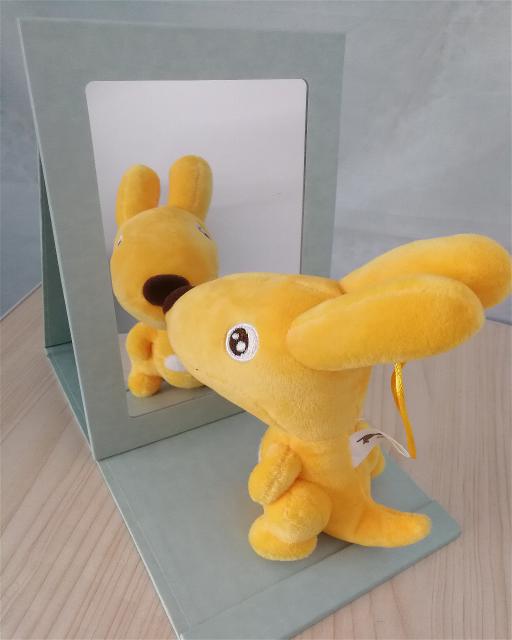} \hspace{-4mm} &
        \includegraphics[width=0.08\textwidth]{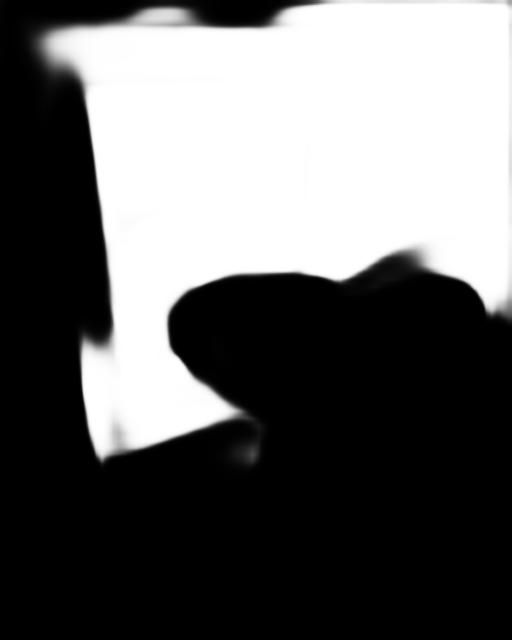} \hspace{-4mm} &
        \includegraphics[width=0.08\textwidth]{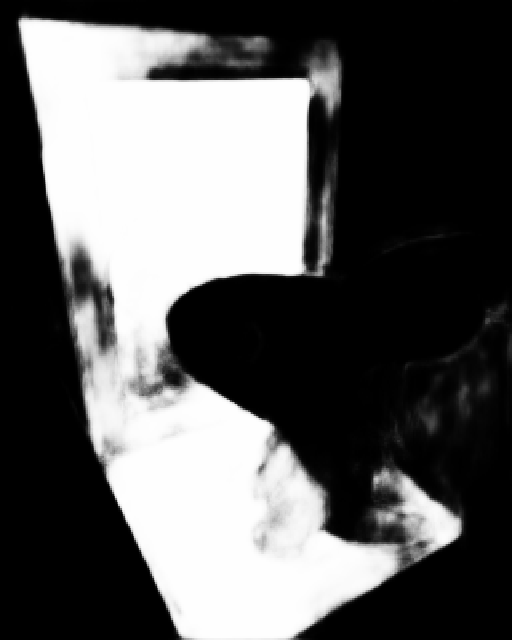} \hspace{-4mm} &
        \includegraphics[width=0.08\textwidth]{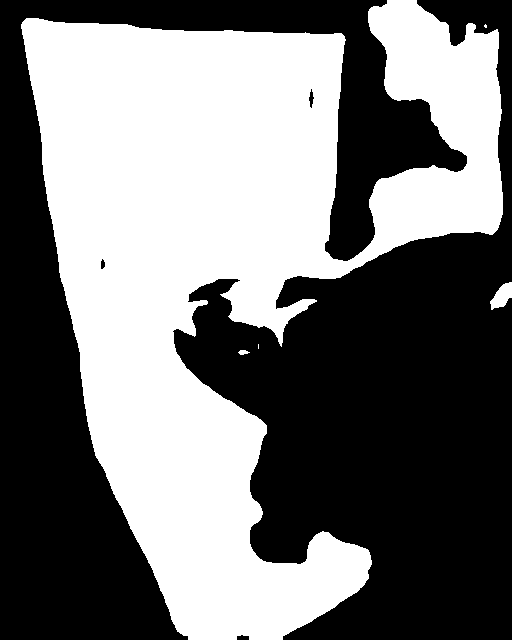} \hspace{-4mm} &
        \includegraphics[width=0.08\textwidth]{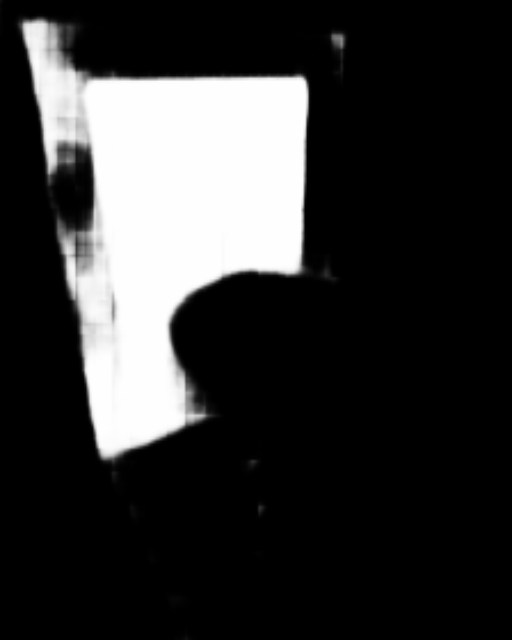} \hspace{-4mm} &
        \includegraphics[width=0.08\textwidth]{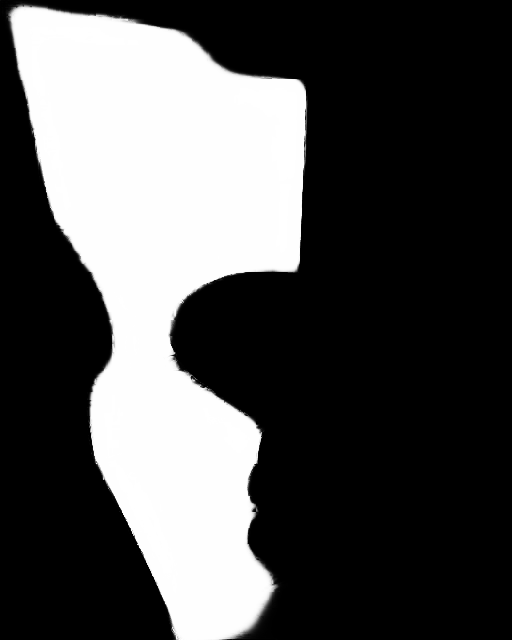} \hspace{-4mm} &
        \includegraphics[width=0.08\textwidth]{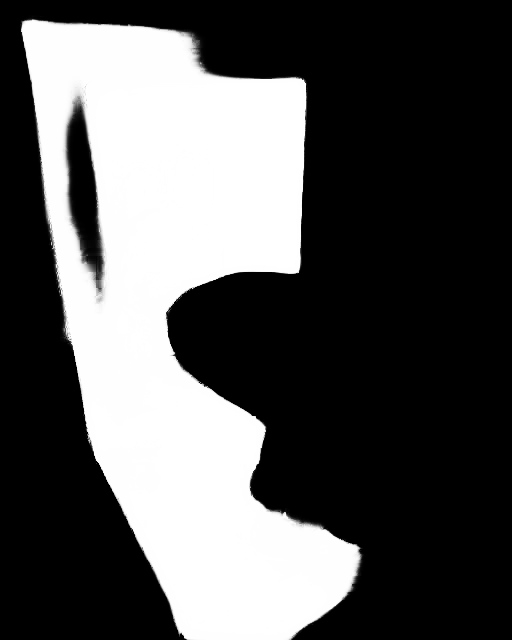} \hspace{-4mm} &
        \includegraphics[width=0.08\textwidth]{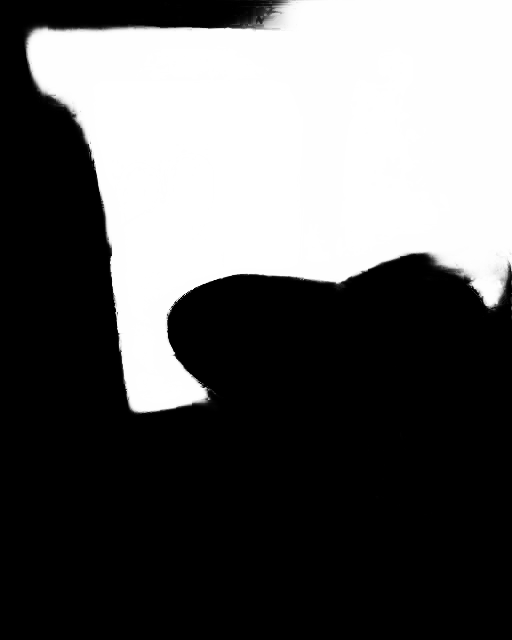} \hspace{-4mm} &
        \includegraphics[width=0.08\textwidth]{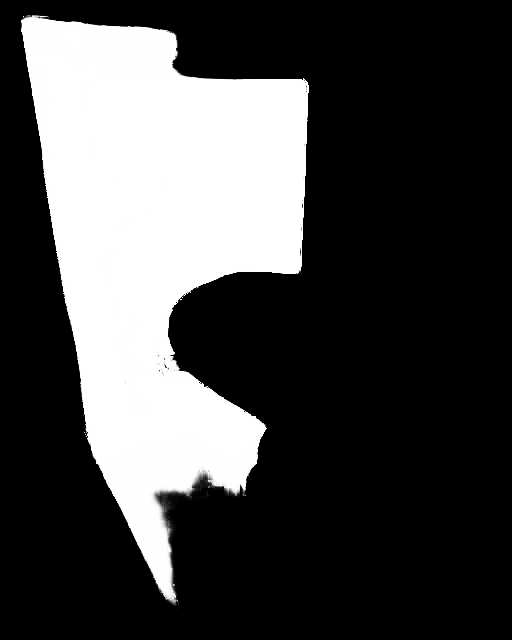} \hspace{-4mm} &
        \includegraphics[width=0.08\textwidth]{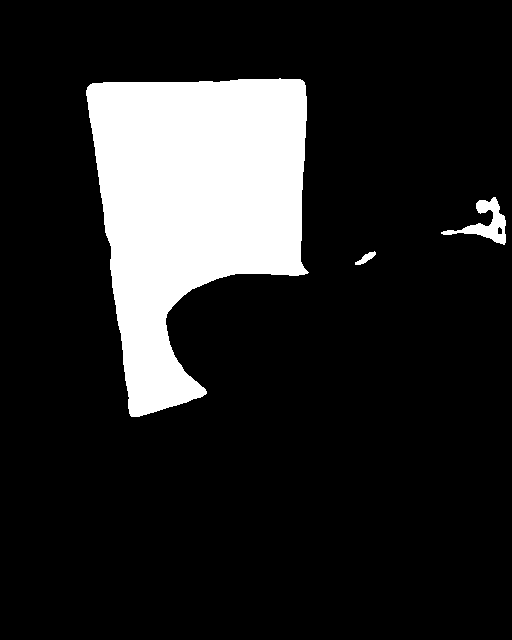} \hspace{-4mm} &
        \includegraphics[width=0.08\textwidth]{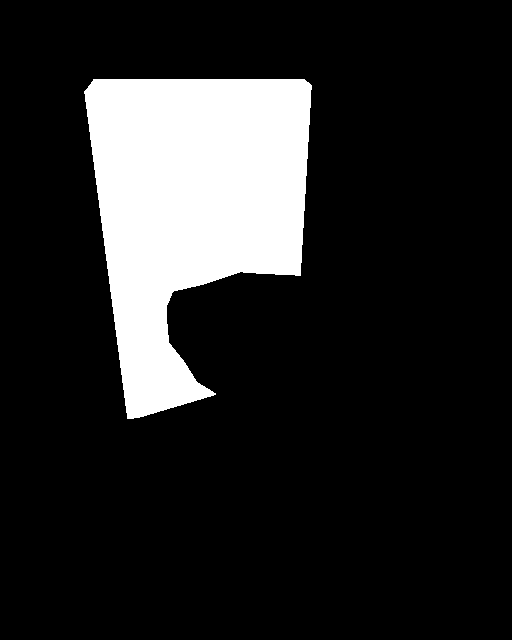} \\
        \includegraphics[width=0.08\textwidth]{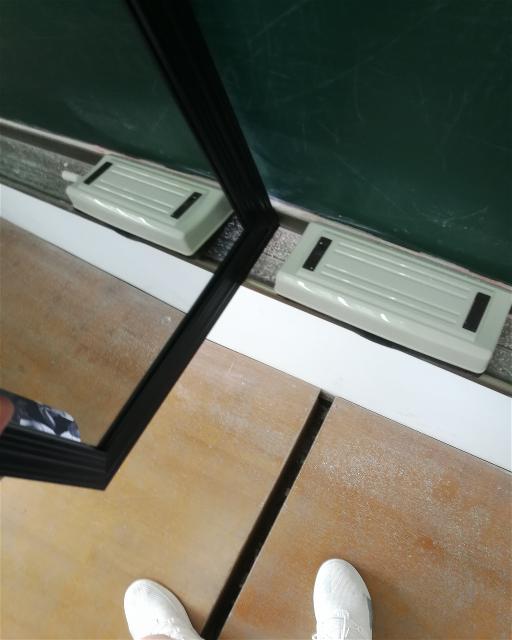} \hspace{-4mm} &
        \includegraphics[width=0.08\textwidth]{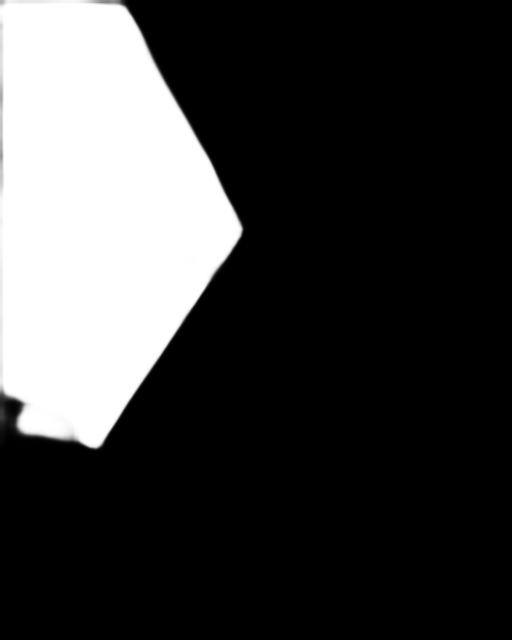} \hspace{-4mm} &
        \includegraphics[width=0.08\textwidth]{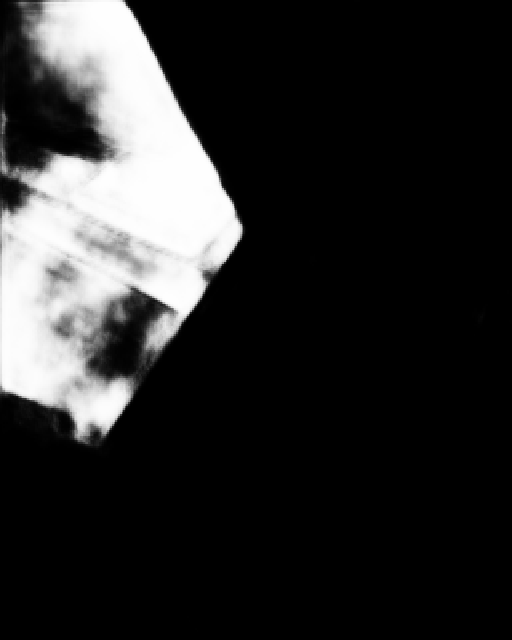} \hspace{-4mm} &
        \includegraphics[width=0.08\textwidth]{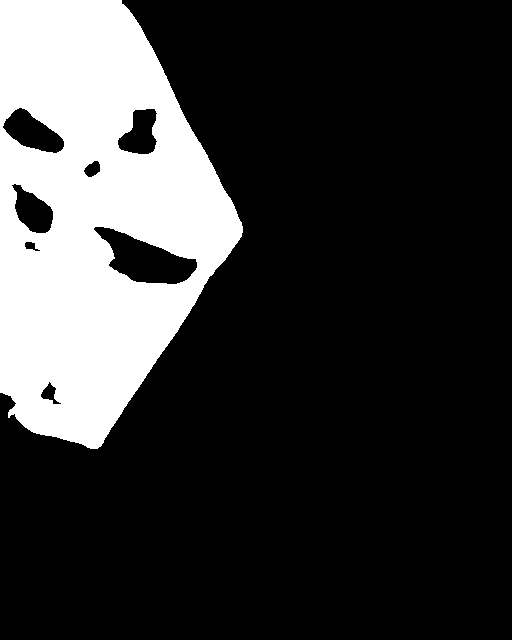} \hspace{-4mm} &
        \includegraphics[width=0.08\textwidth]{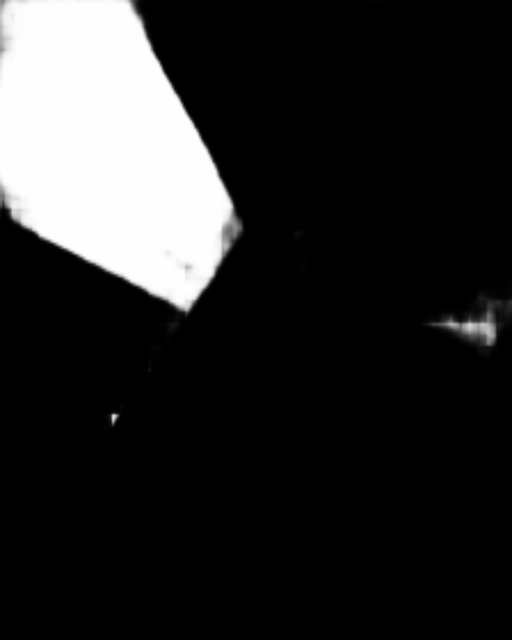} \hspace{-4mm} &
        \includegraphics[width=0.08\textwidth]{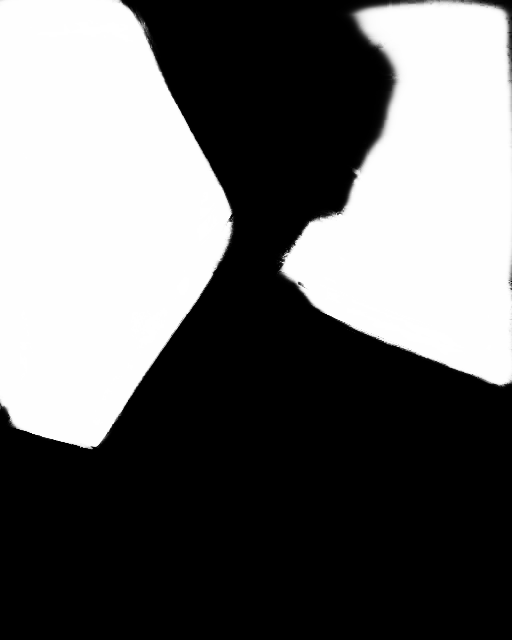} \hspace{-4mm} &
        \includegraphics[width=0.08\textwidth]{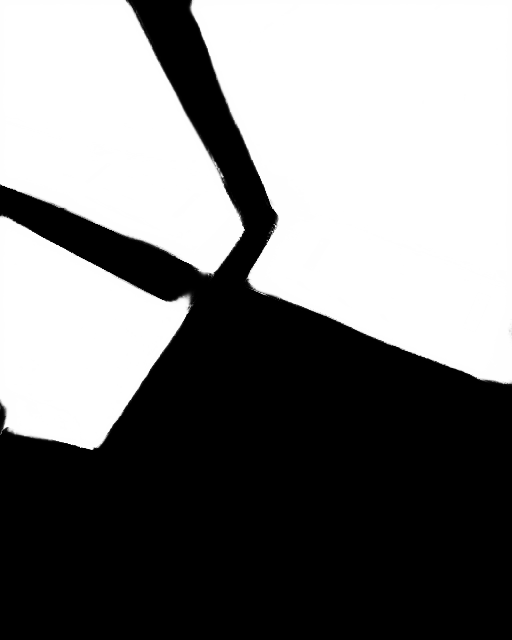} \hspace{-4mm} &
        \includegraphics[width=0.08\textwidth]{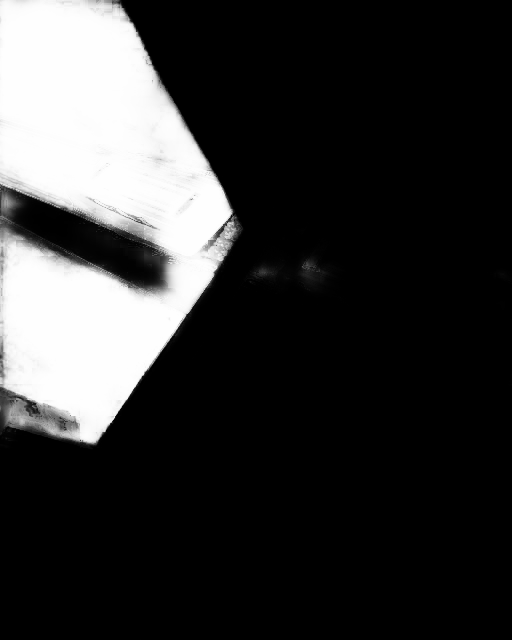} \hspace{-4mm} &
        \includegraphics[width=0.08\textwidth]{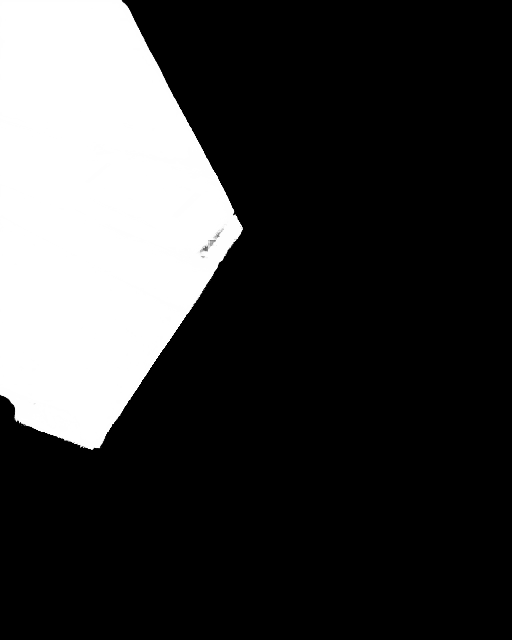} \hspace{-4mm} &
        \includegraphics[width=0.08\textwidth]{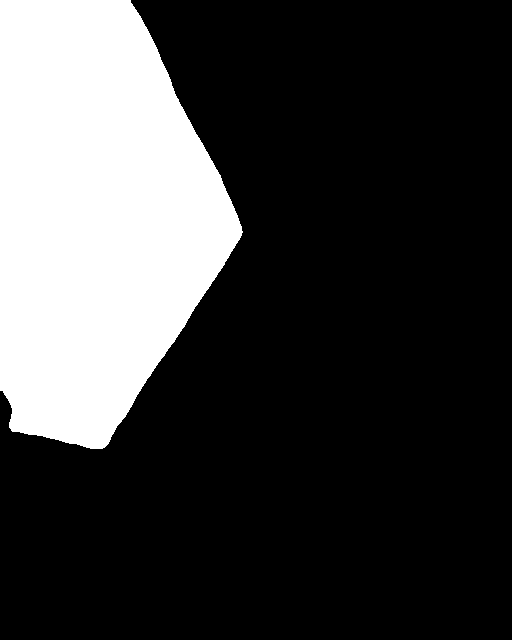} \hspace{-4mm} &
        \includegraphics[width=0.08\textwidth]{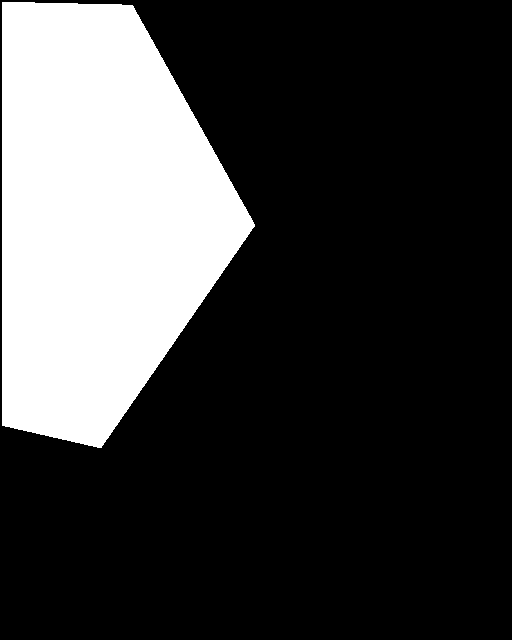} \\
        \includegraphics[width=0.08\textwidth]{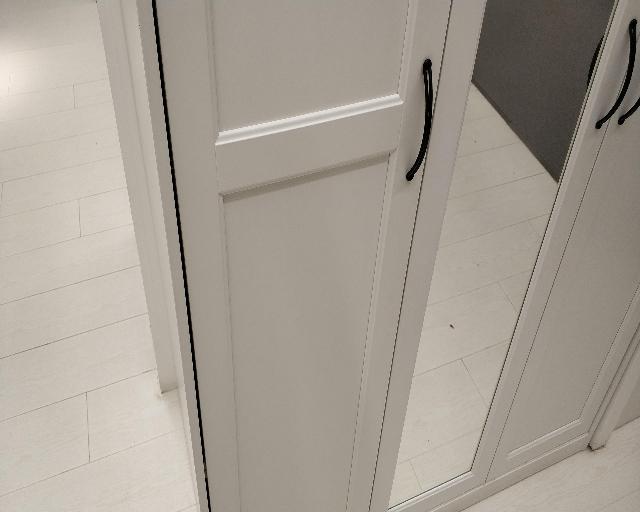} \hspace{-4mm} &
        \includegraphics[width=0.08\textwidth]{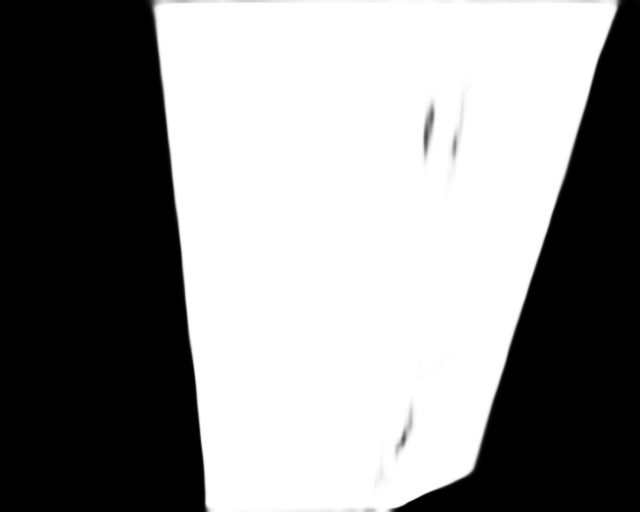} \hspace{-4mm} &
        \includegraphics[width=0.08\textwidth]{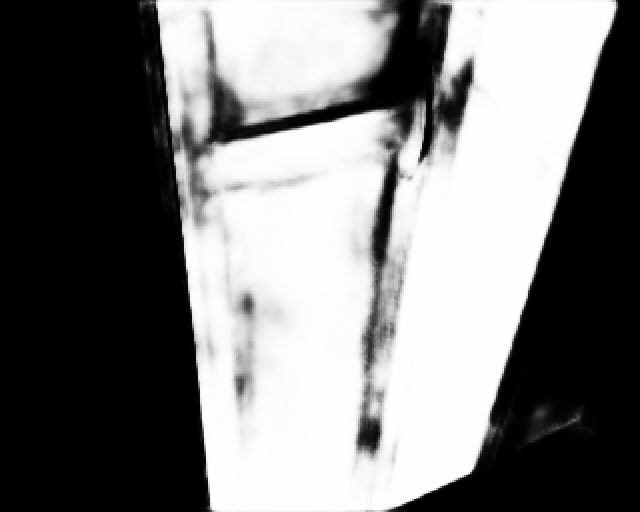} \hspace{-4mm} &
        \includegraphics[width=0.08\textwidth]{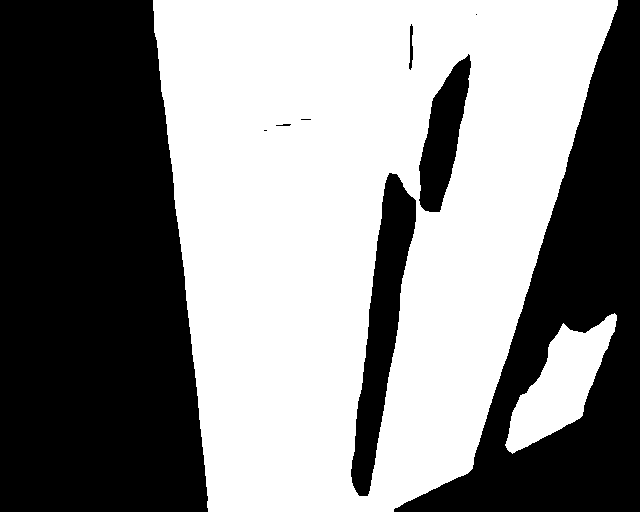} \hspace{-4mm} &
        \includegraphics[width=0.08\textwidth]{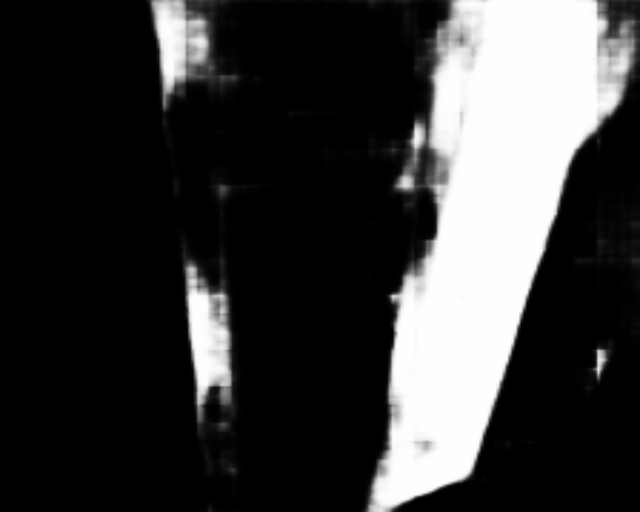} \hspace{-4mm} &
        \includegraphics[width=0.08\textwidth]{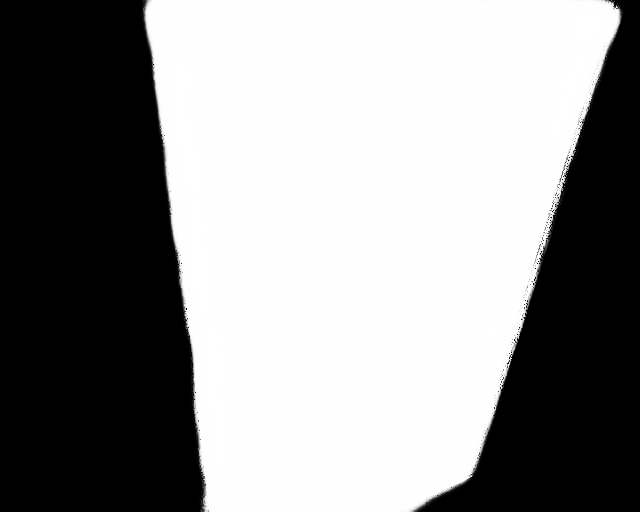} \hspace{-4mm} &
        \includegraphics[width=0.08\textwidth]{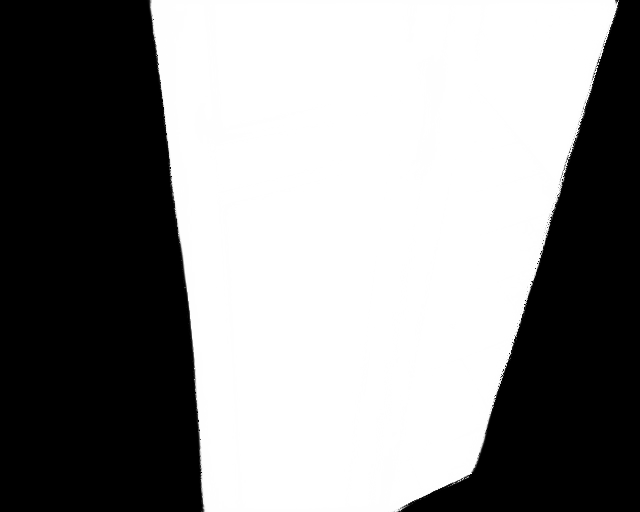} \hspace{-4mm} &
        \includegraphics[width=0.08\textwidth]{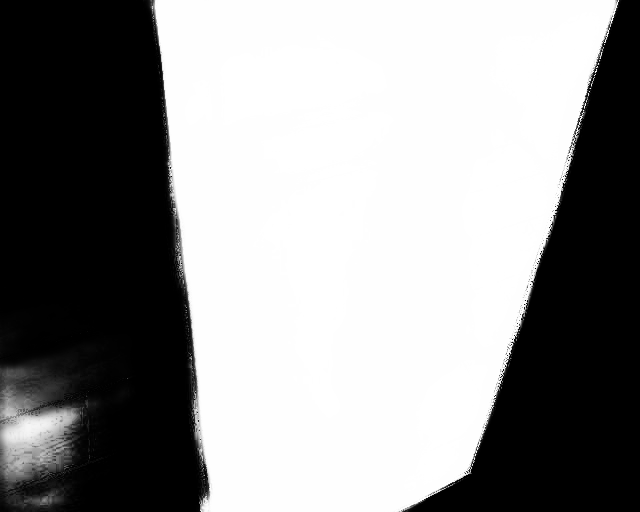} \hspace{-4mm} &
        \includegraphics[width=0.08\textwidth]{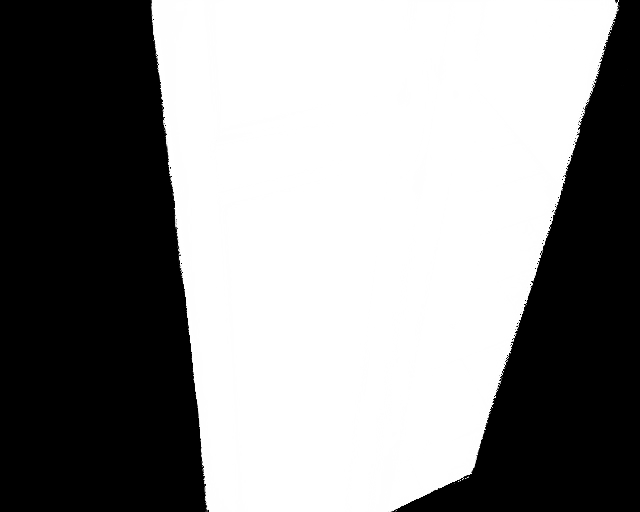} \hspace{-4mm} &
        \includegraphics[width=0.08\textwidth]{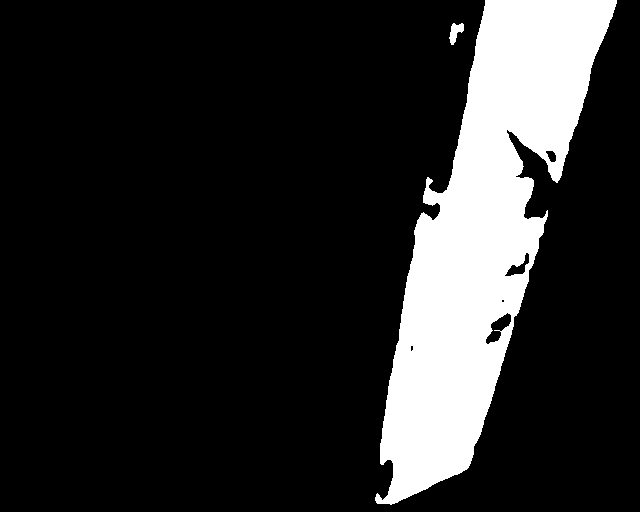} \hspace{-4mm} &
        \includegraphics[width=0.08\textwidth]{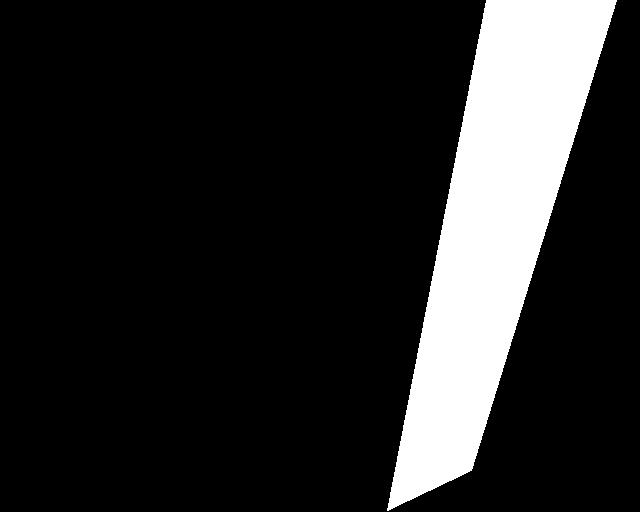} \\
        Image \hspace{-4mm} & CPDNet \hspace{-4mm} & MINet \hspace{-4mm} & LDF \hspace{-4mm} & VST \hspace{-4mm} & MirrorNet \hspace{-4mm} & PMDNet \hspace{-4mm} & SANet \hspace{-4mm} & VCNet \hspace{-4mm} & Ours \hspace{-4mm} & GT
    \end{tabular}
    \caption{Visualization results on MSD and PMD datasets. The first two rows are examples of loose symmetry relationships. The last row has scenes where mirrors are similar to their surroundings.}
    \label{fig:msd&pmd}
\end{figure*}

To evaluate SATNet, we extensively compare it with several state-of-the-art methods.
As shown in Table~\ref{tab:msd&pmd}, we select 8 state-of-the-art methods for the comparison on MSD dataset and PMD dataset, including 4 RGB salient object detection methods CPDNet~\cite{wu2019cascaded}, MINet~\cite{pang2020multi}, LDF~\cite{Wei_2020_CVPR}, and VST~\cite{Liu_2021_ICCV}, and 4 mirror detection methods MirrorNet~\cite{yang2019my}, PMDNet~\cite{lin2020progressive}, SANet~\cite{guan2022learning}, and VCNet~\cite{tan2022mirror}.
Our network outperforms other methods in terms of all the evaluation metrics.
Fig.~\ref{fig:msd&pmd} provides the visualized comparison with those methods.
The first two rows are examples of loose symmetry relationships.
Our network can precisely distinguish real-world objects from their mirror reflections.
In the first row, the cartoon toy and its reflection in mirrors cannot construct an apparent reflection symmetry, but our network can still perceive which part is in mirrors.
Albeit PMDNet~\cite{lin2020progressive} has a specific module for modeling similarity relationships, it fails in handling an easy case in the second row, in which a chalk eraser is reflected in the mirror.
The last row has scenes where mirrors are similar to their surroundings.
Our method can well exclude the non-mirror region, while the competing methods tend to classify the similar area as the mirror region, especially the four mirror detection methods.
The results show that symmetry-awareness is beneficial for mirror detection, and our method can utilize the symmetry information well.

\begin{figure*}
    \centering
    \begin{tabular}{ccccccccccc}
        \includegraphics[width=0.08\textwidth]{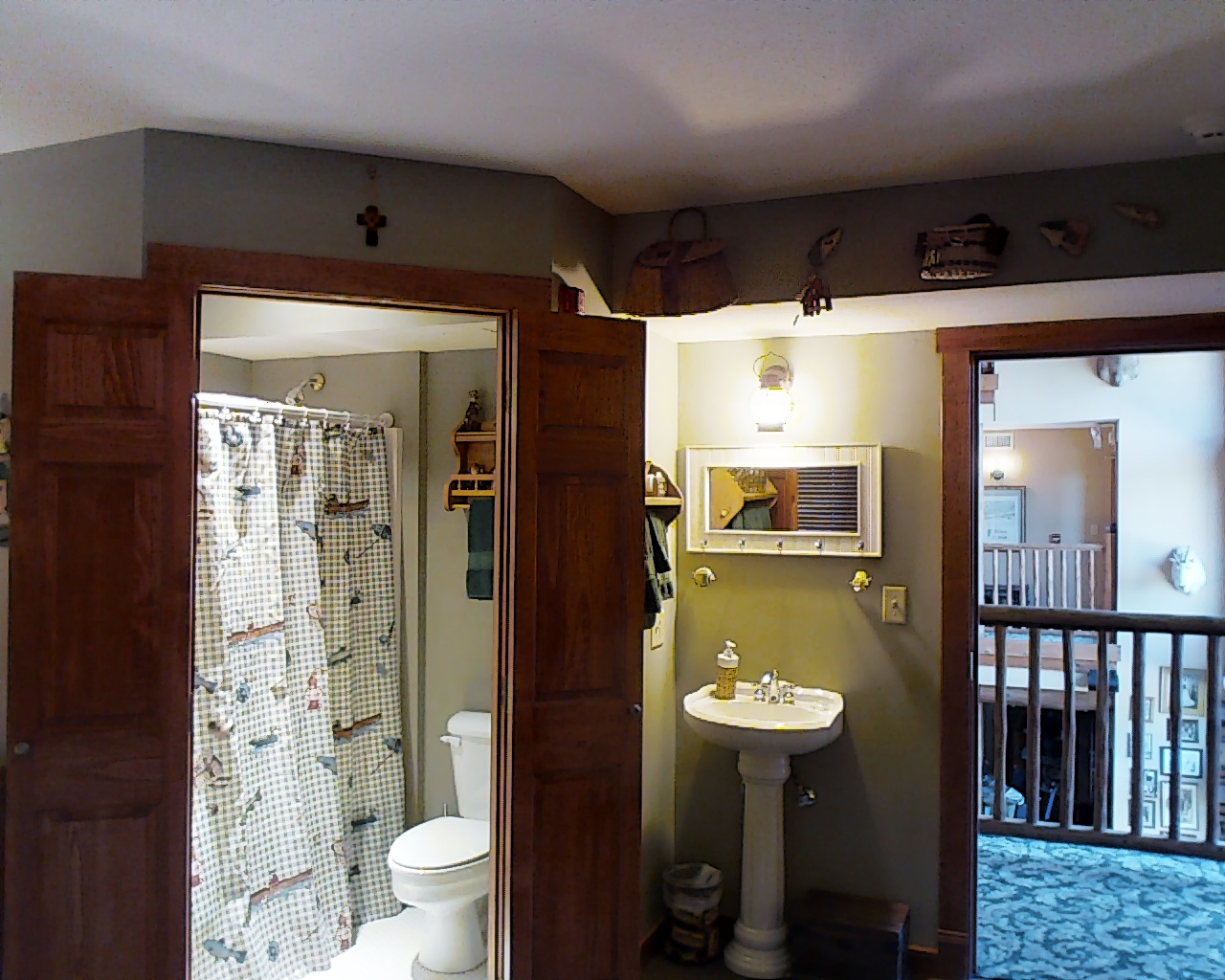} \hspace{-4mm} &
        \includegraphics[width=0.08\textwidth]{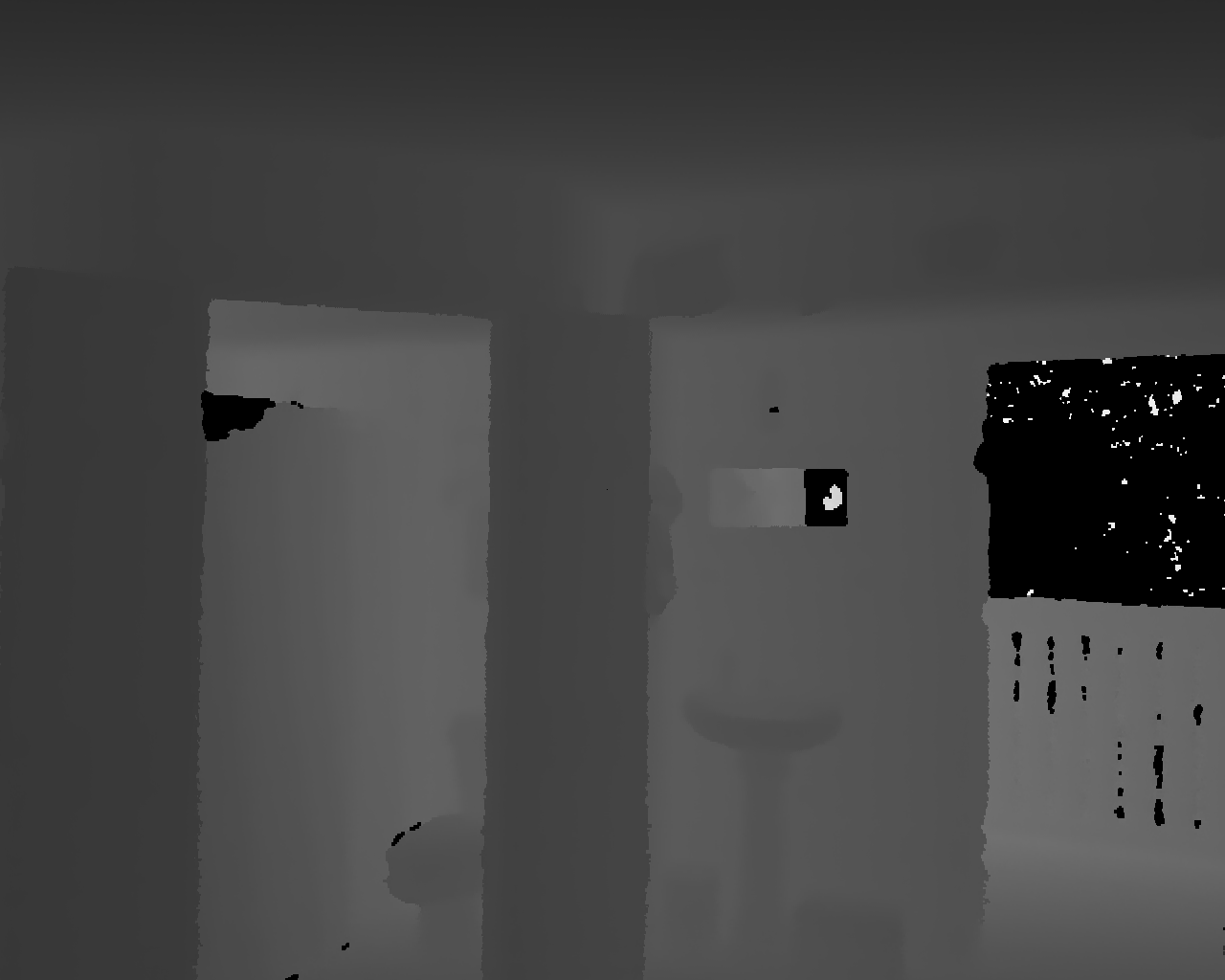} \hspace{-4mm} &
        \includegraphics[width=0.08\textwidth]{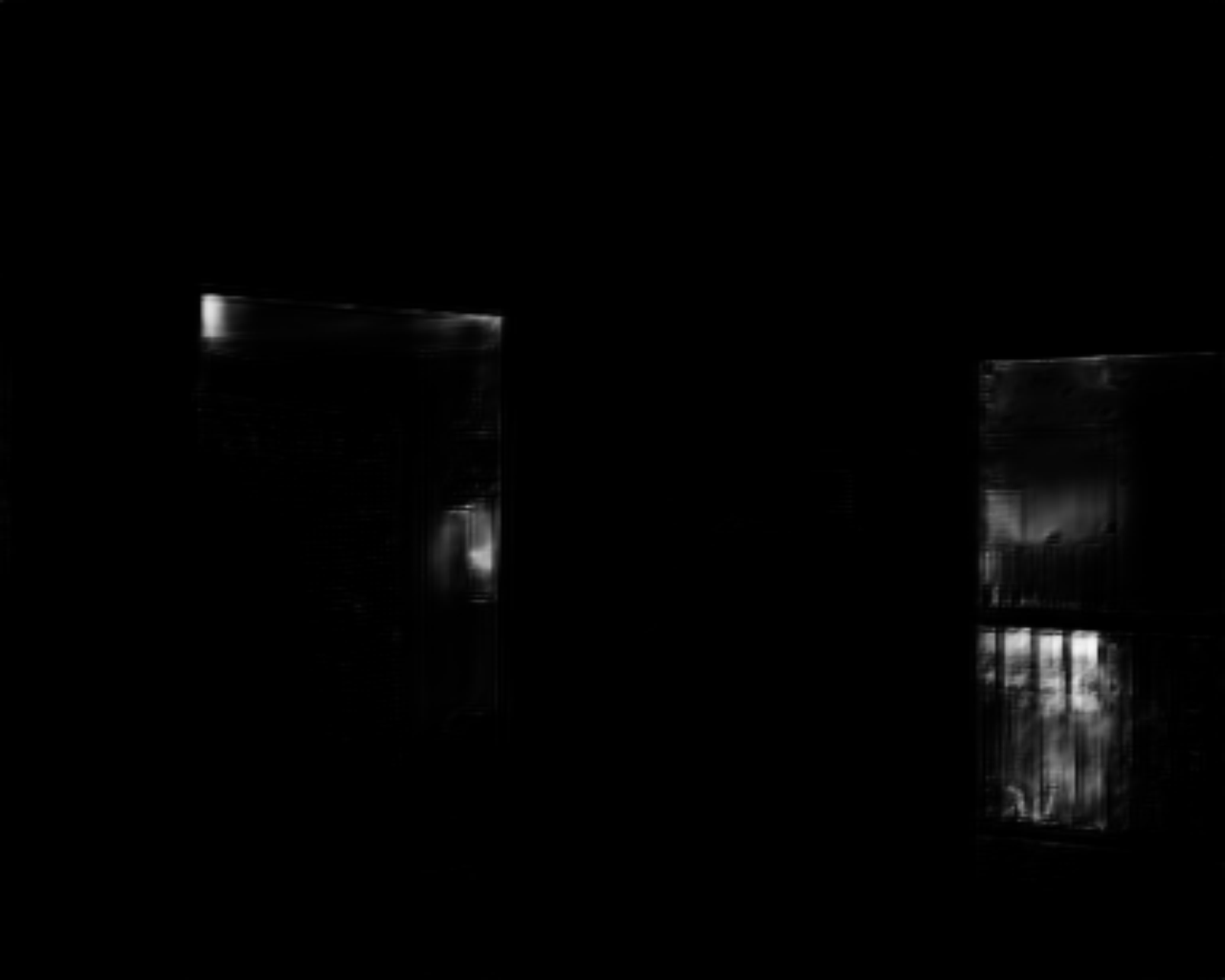} \hspace{-4mm} &
        \includegraphics[width=0.08\textwidth]{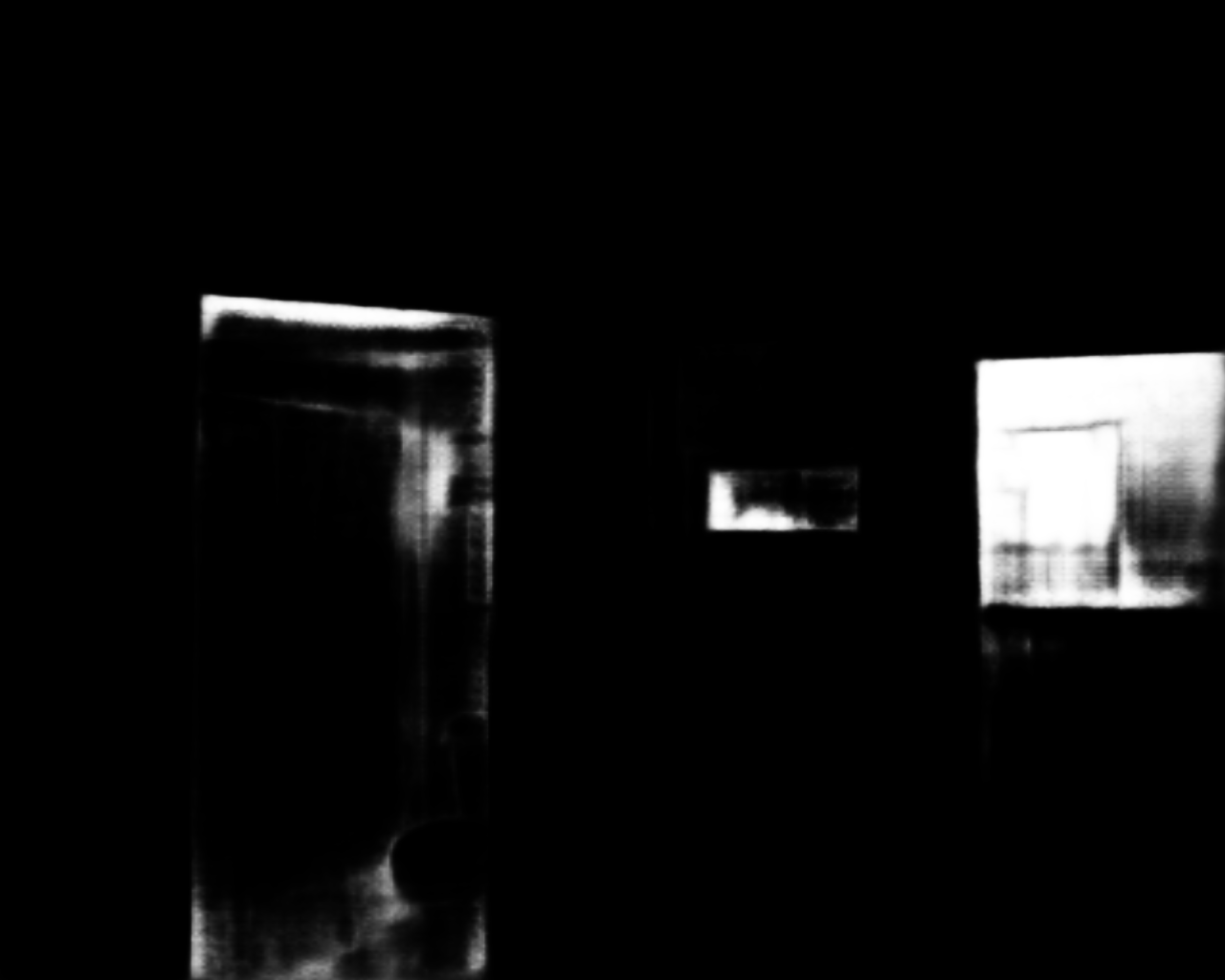} \hspace{-4mm} &
        \includegraphics[width=0.08\textwidth]{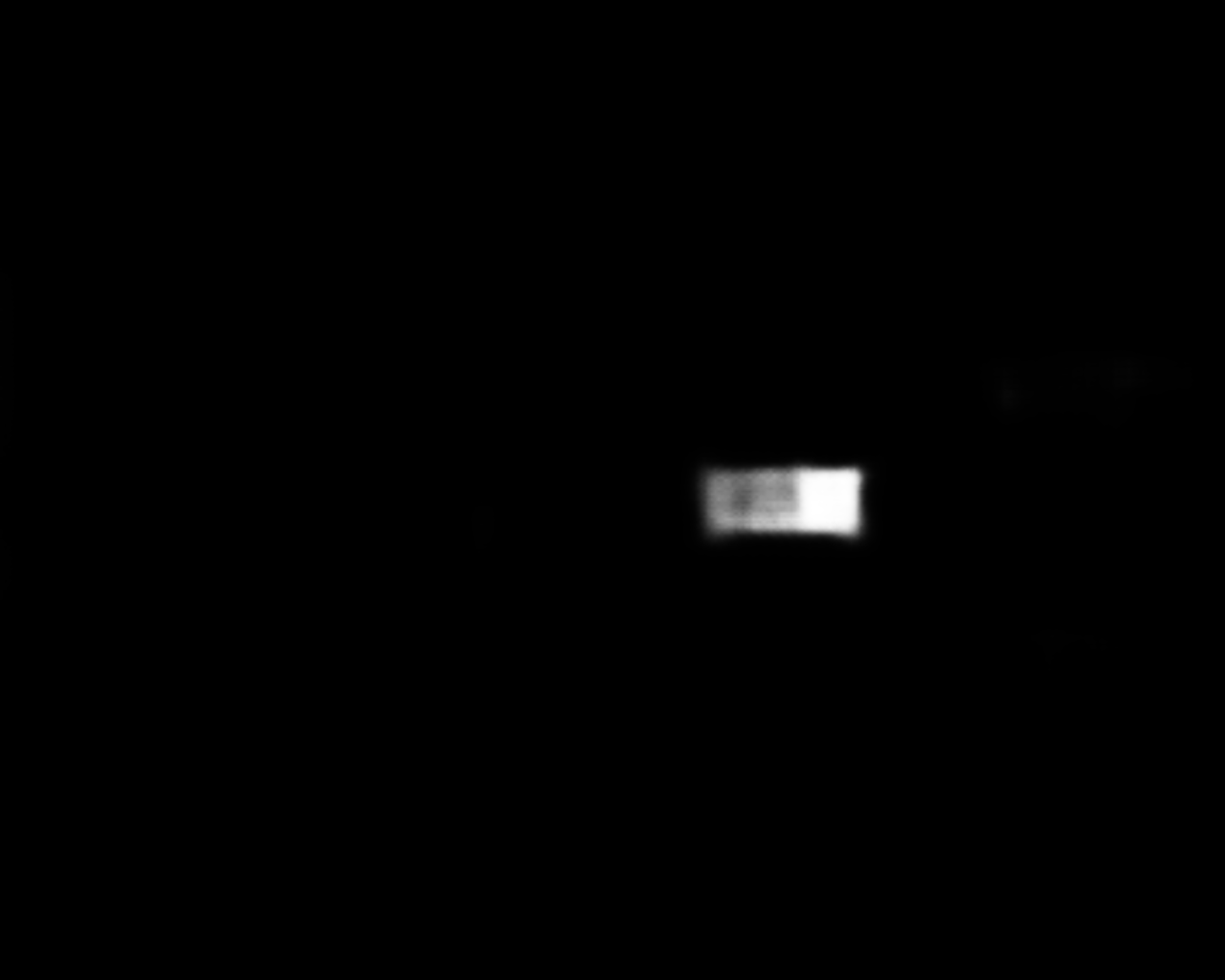} \hspace{-4mm} &
        \includegraphics[width=0.08\textwidth]{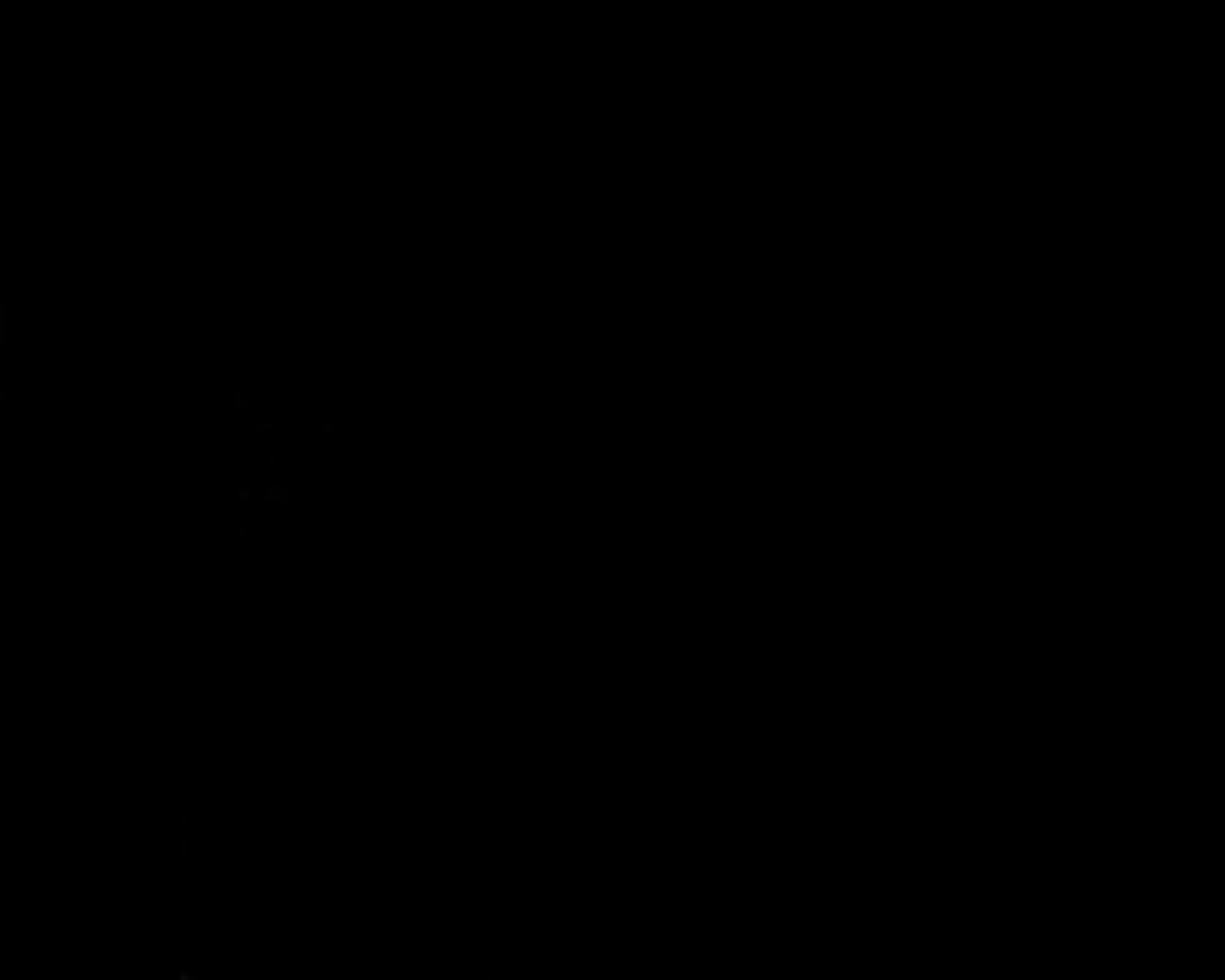} \hspace{-4mm} &
        \includegraphics[width=0.08\textwidth]{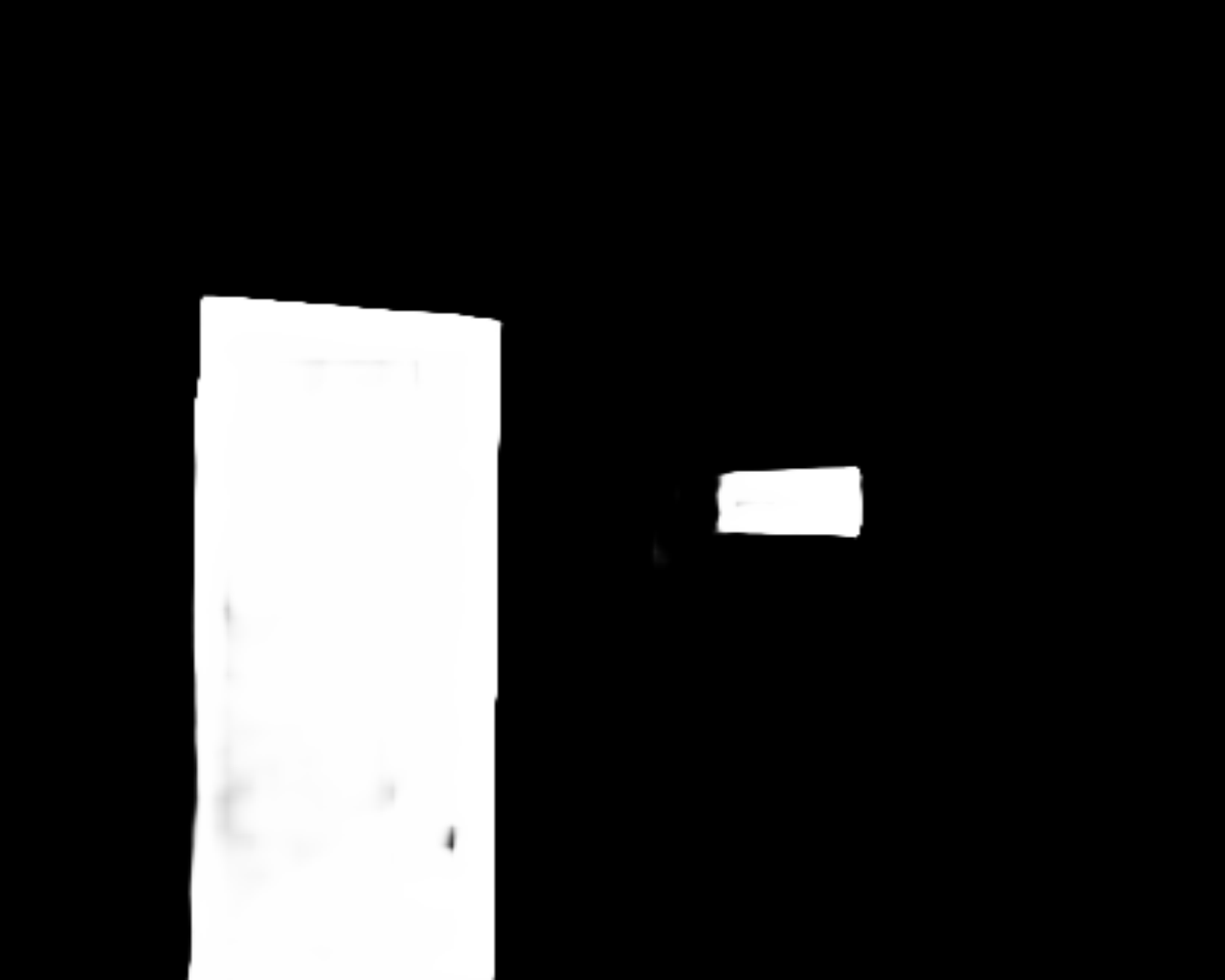} \hspace{-4mm} &
        \includegraphics[width=0.08\textwidth]{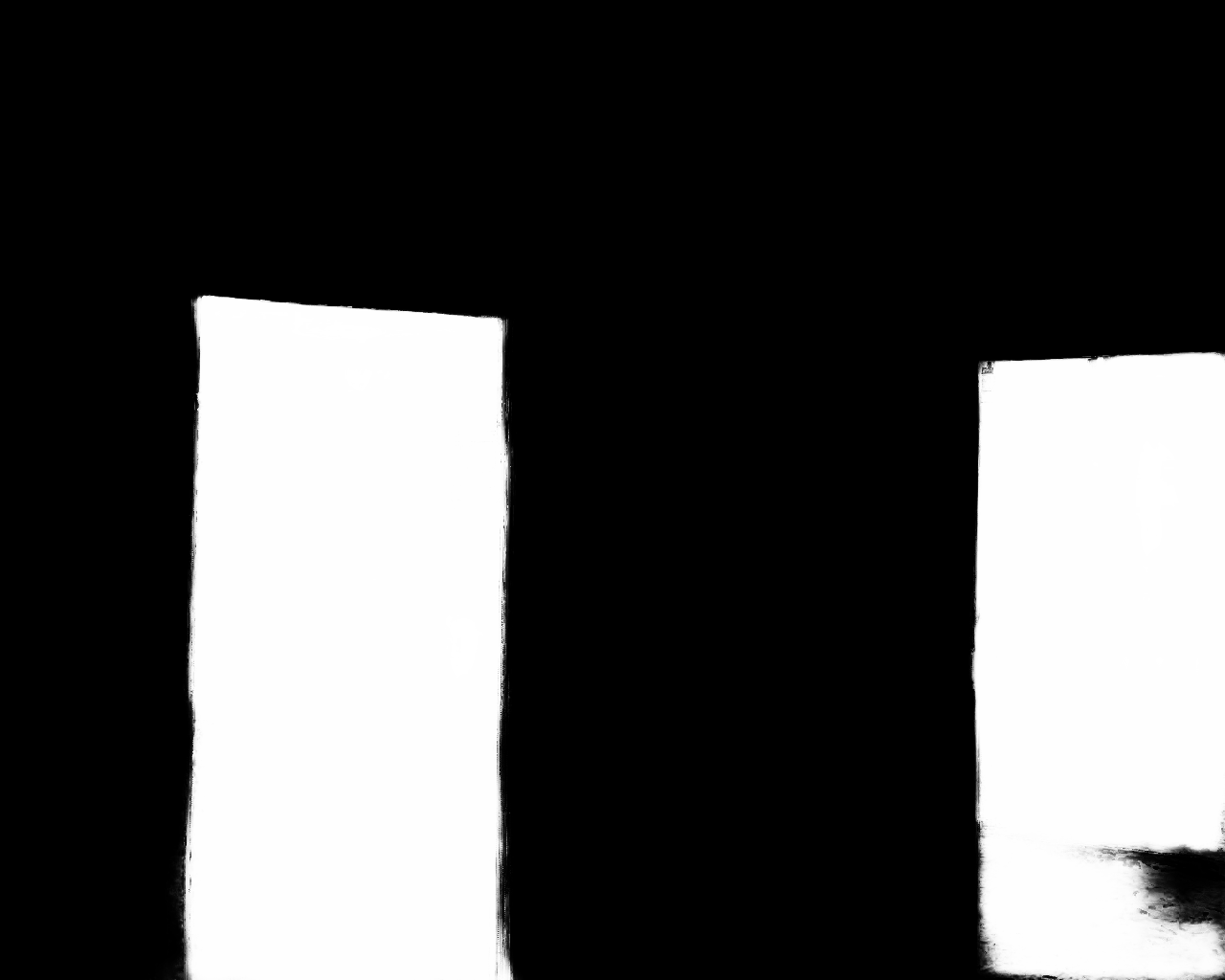} \hspace{-4mm} &
        \includegraphics[width=0.08\textwidth]{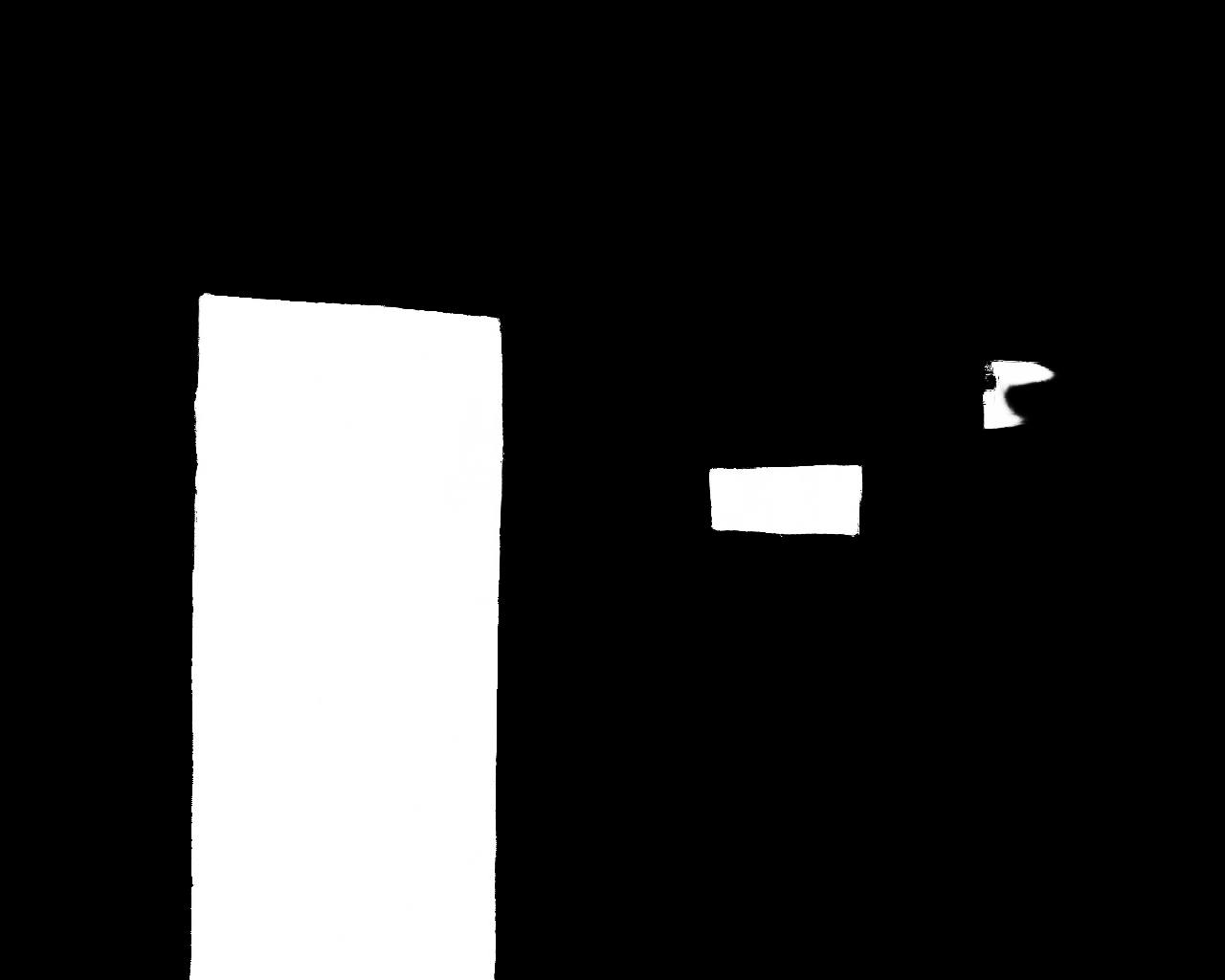} \hspace{-4mm} &
        \includegraphics[width=0.08\textwidth]{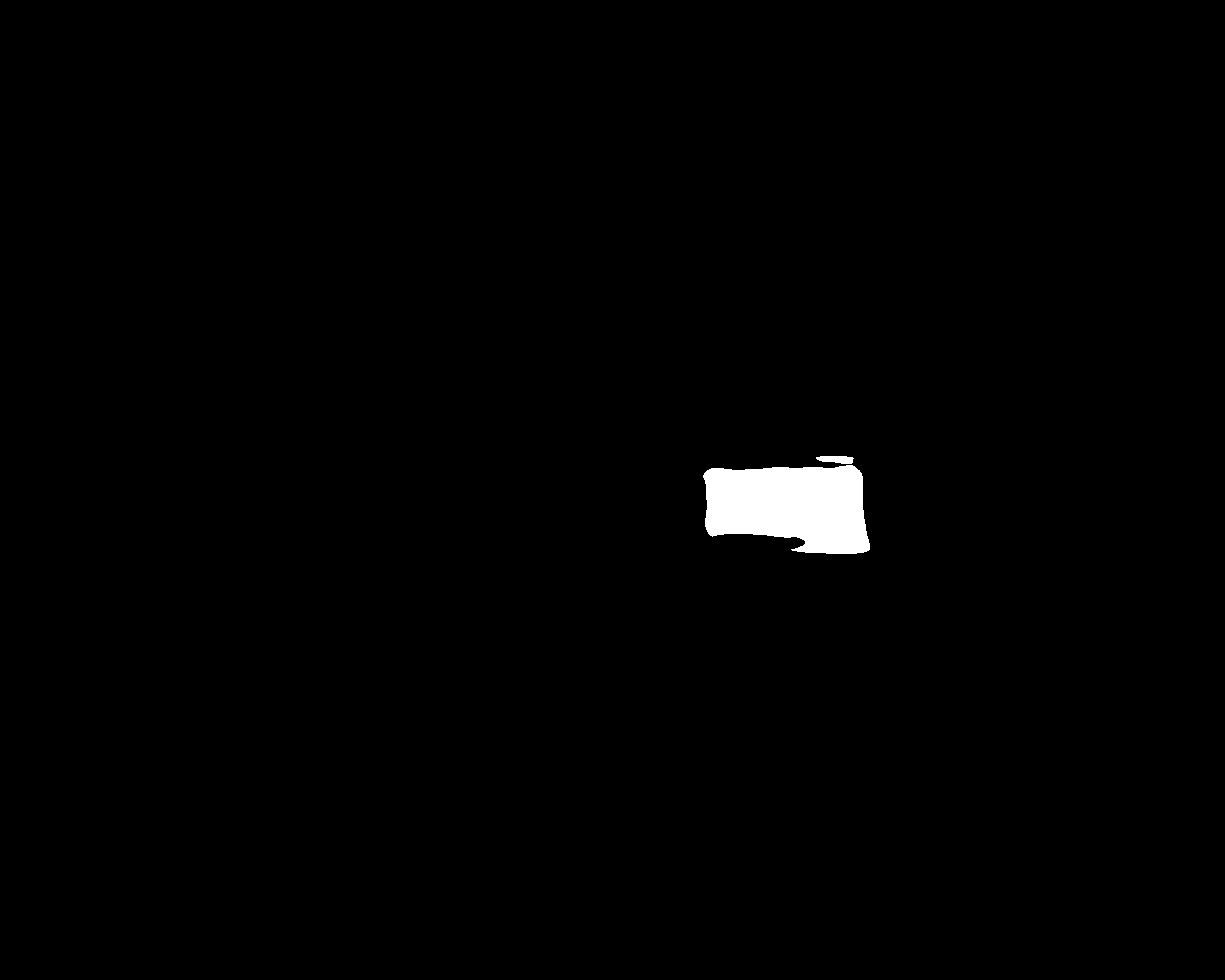} \hspace{-4mm} &
        \includegraphics[width=0.08\textwidth]{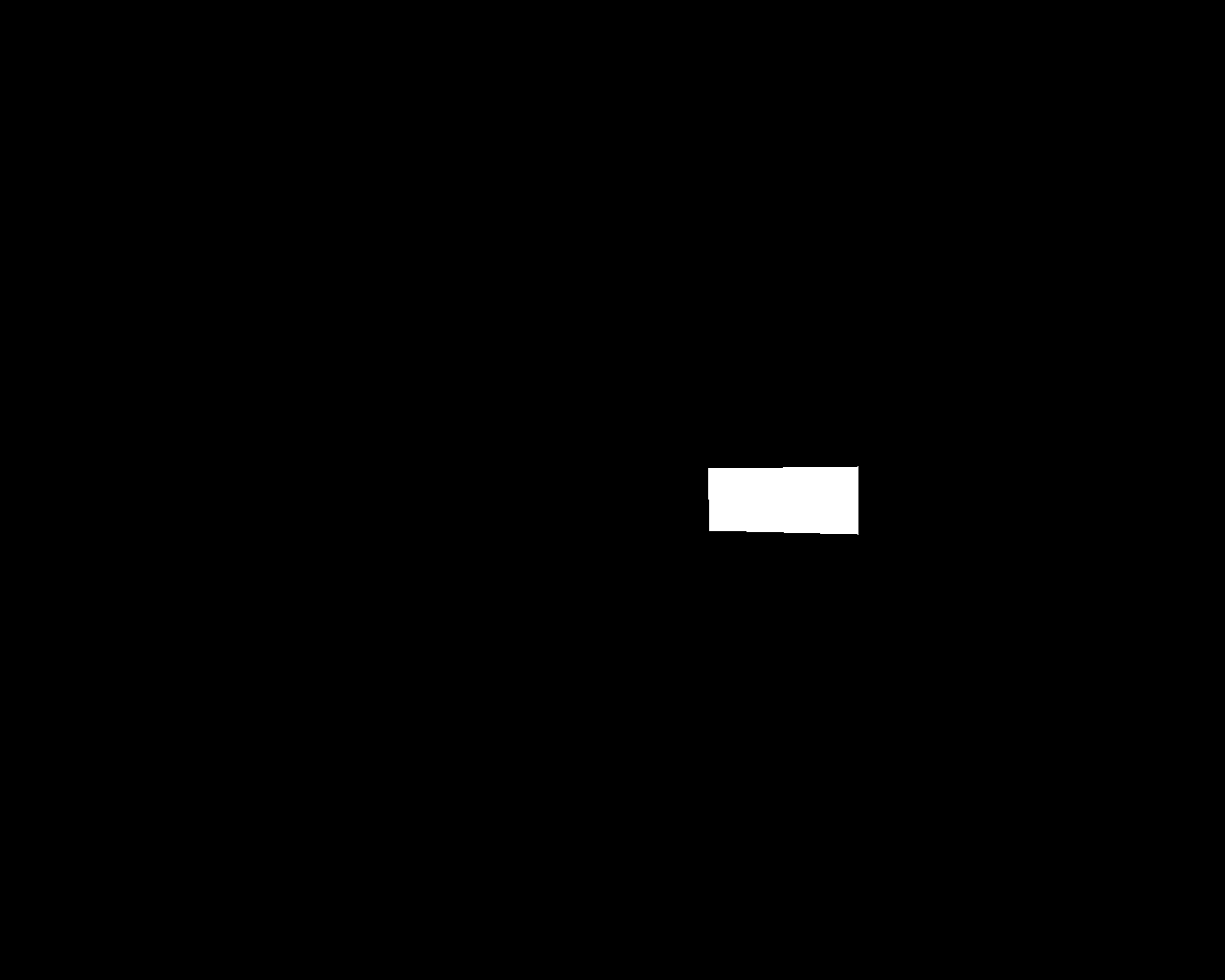} \\
        \includegraphics[width=0.08\textwidth]{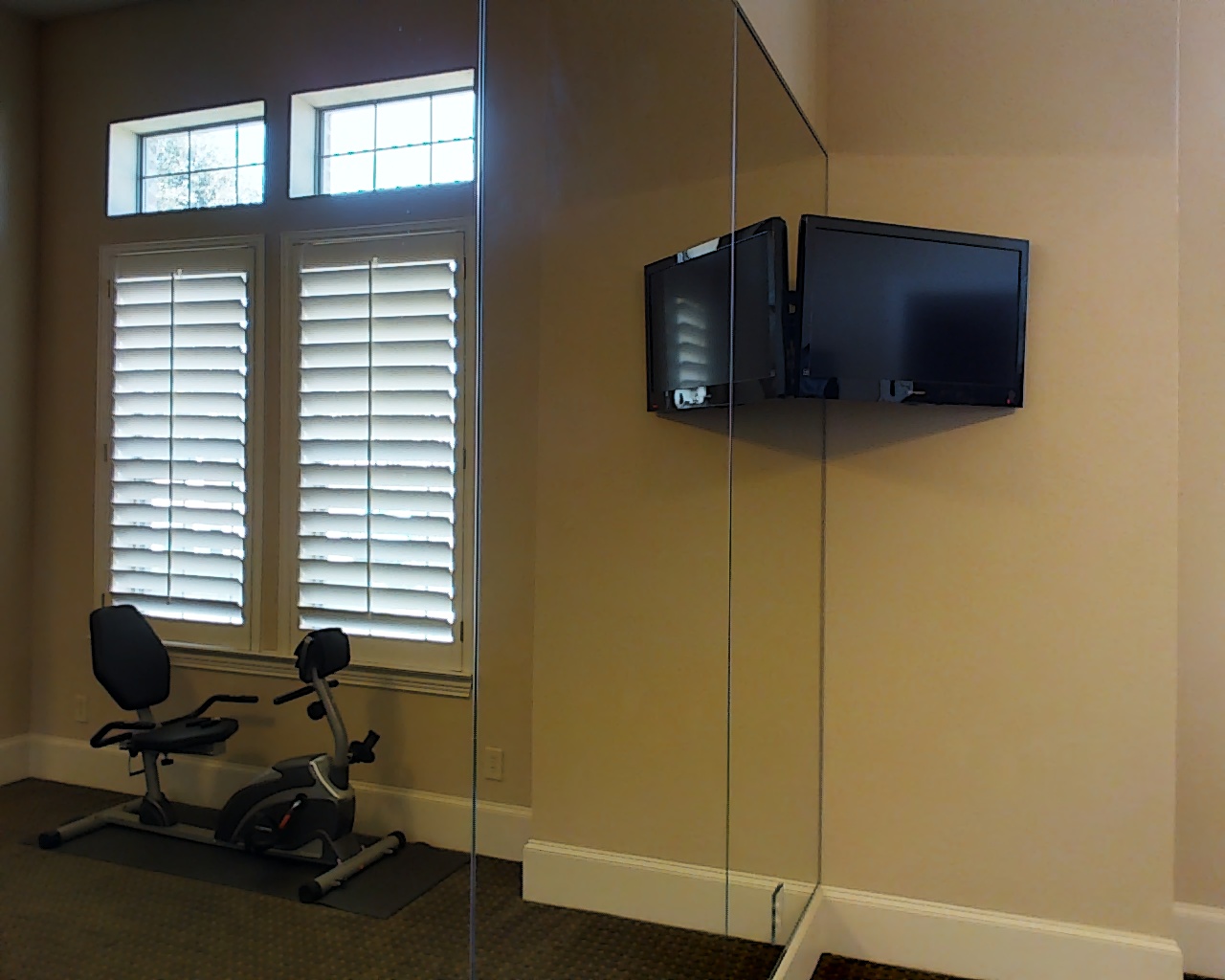} \hspace{-4mm} &
        \includegraphics[width=0.08\textwidth]{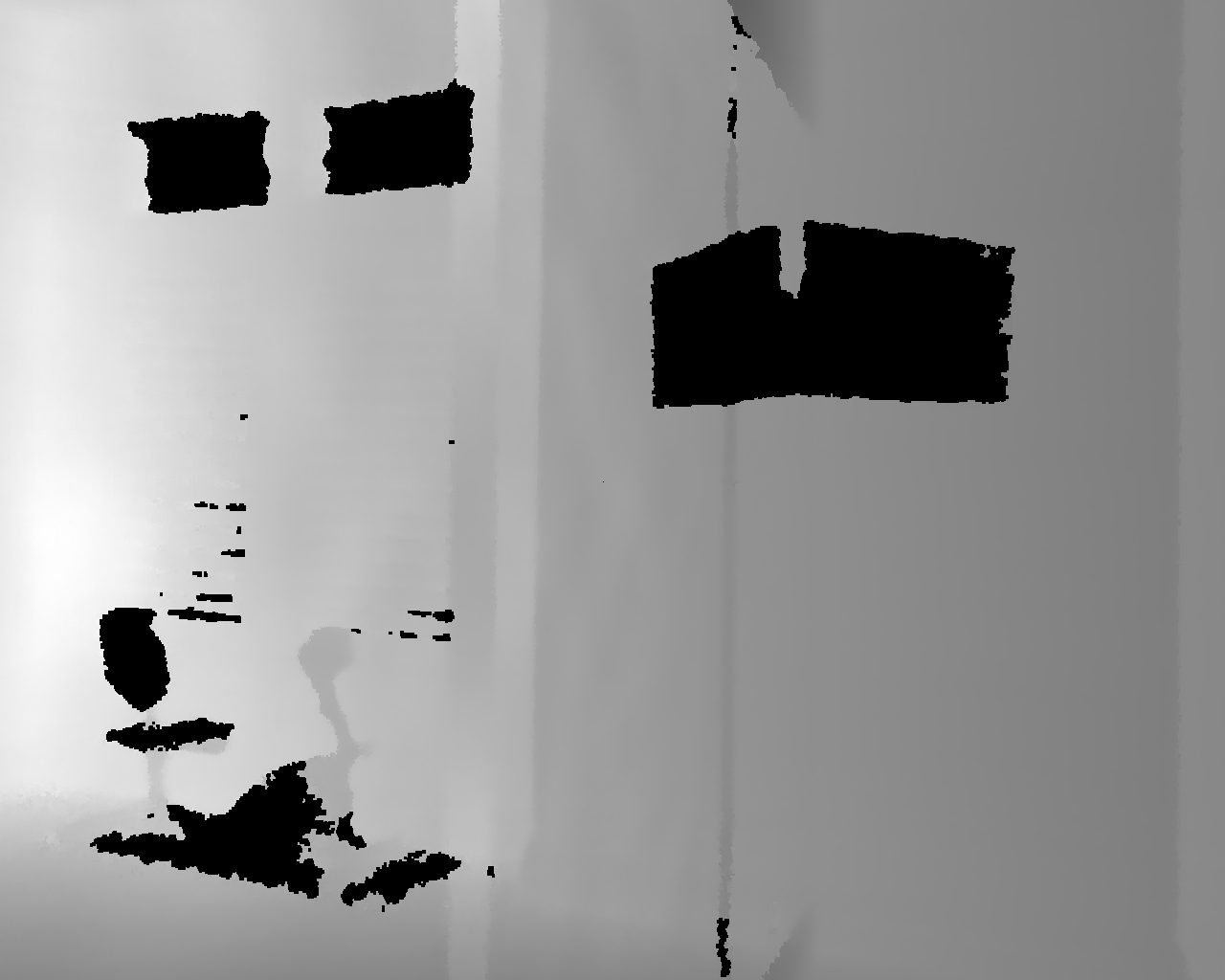} \hspace{-4mm} &
        \includegraphics[width=0.08\textwidth]{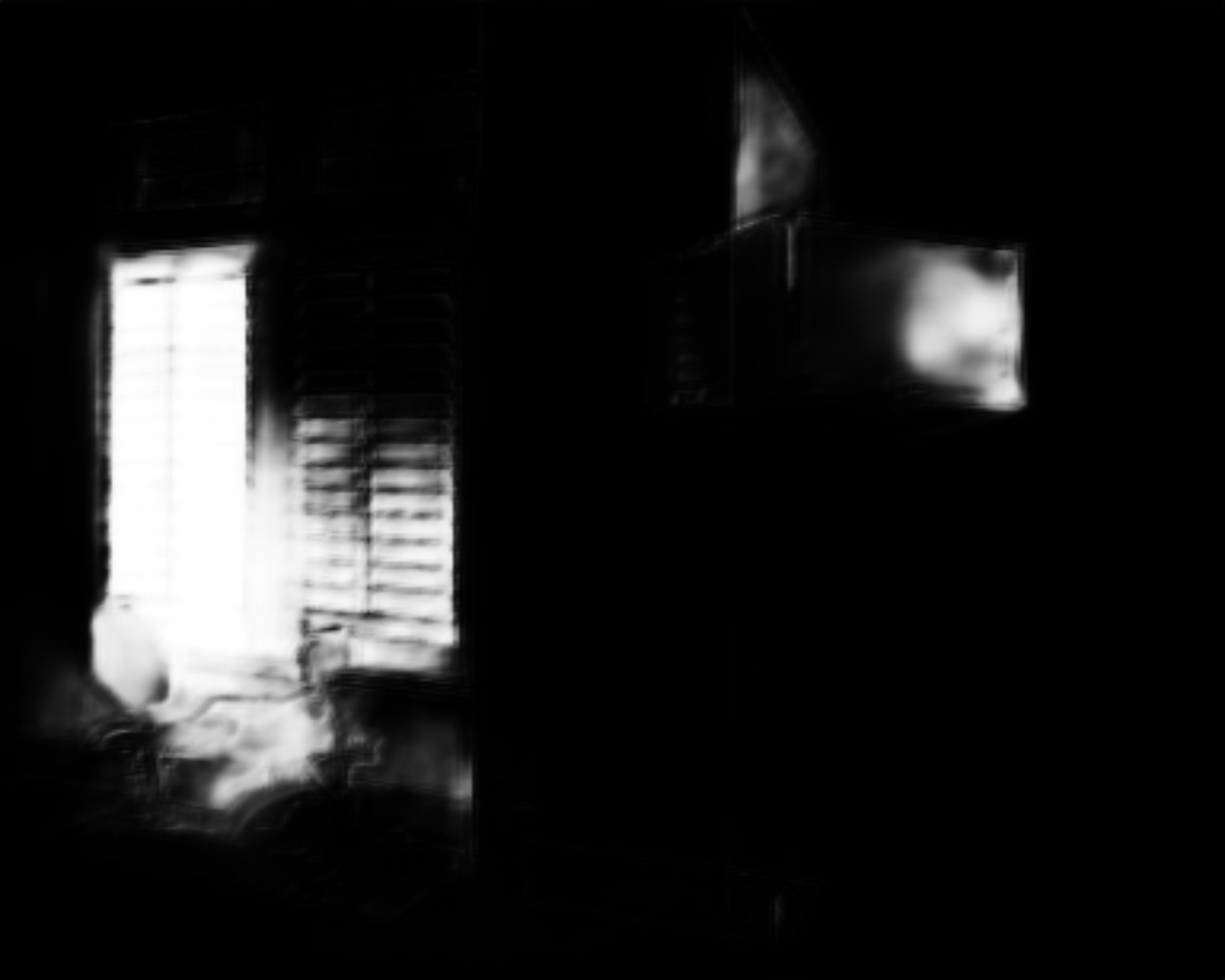} \hspace{-4mm} &
        \includegraphics[width=0.08\textwidth]{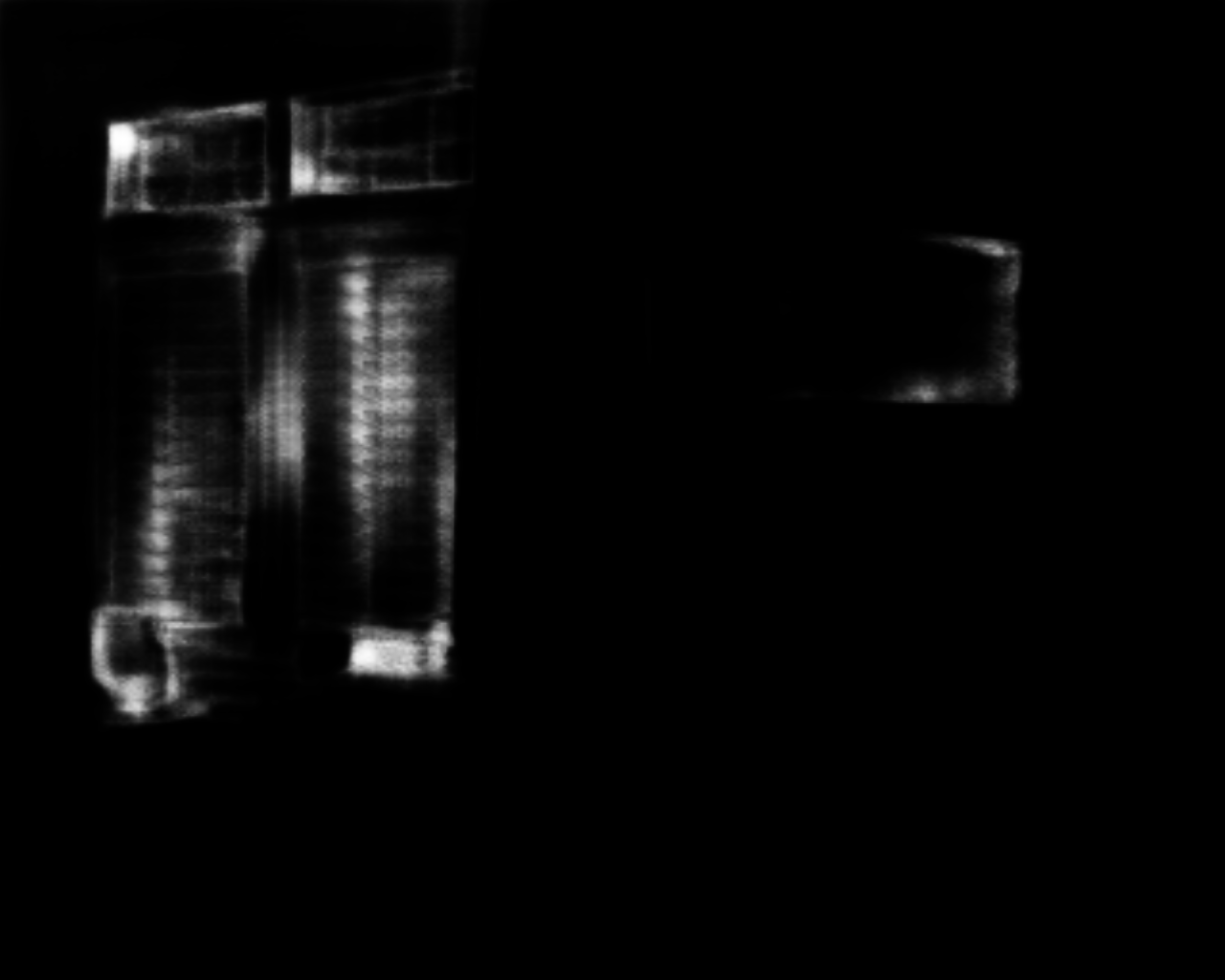} \hspace{-4mm} &
        \includegraphics[width=0.08\textwidth]{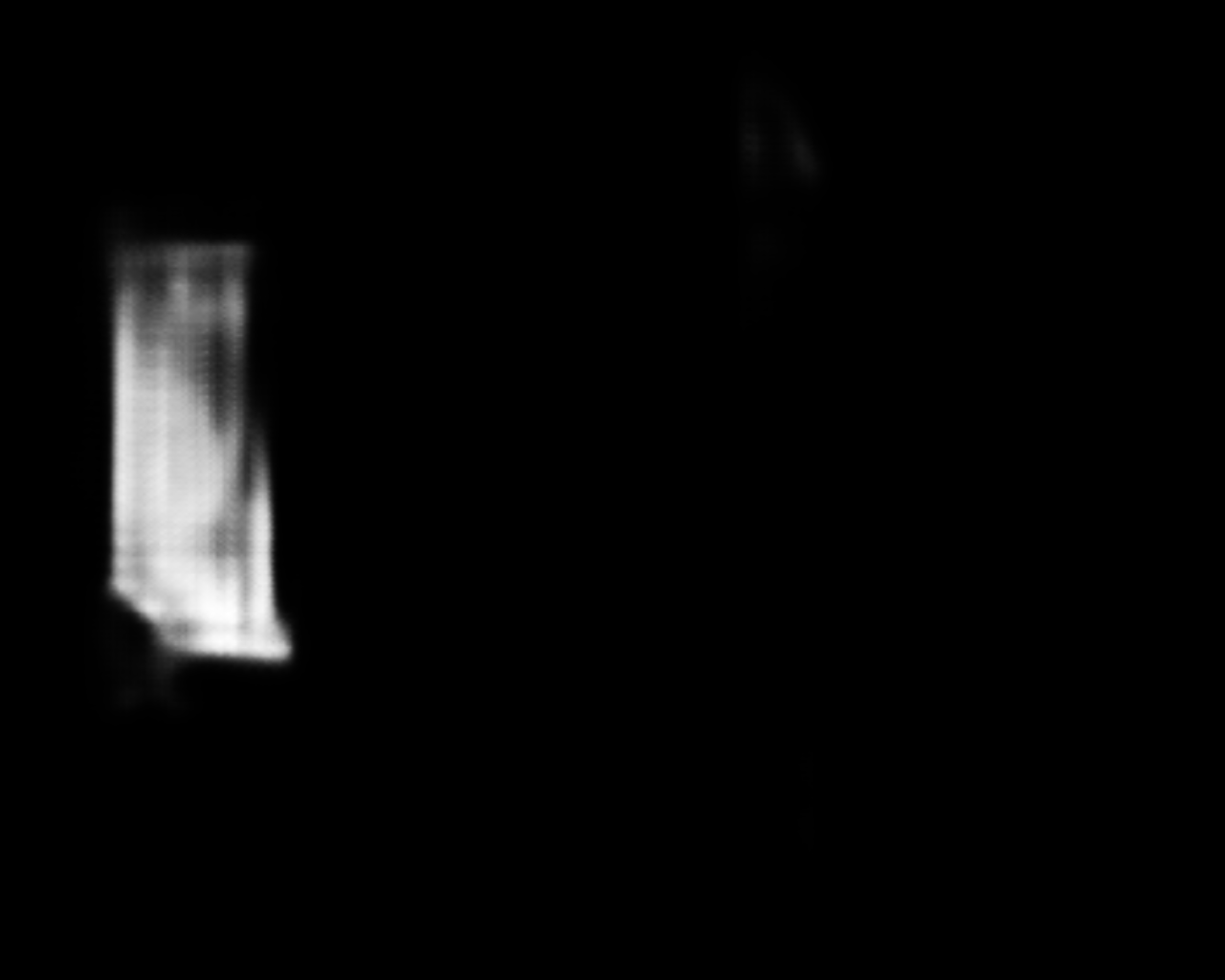} \hspace{-4mm} &
        \includegraphics[width=0.08\textwidth]{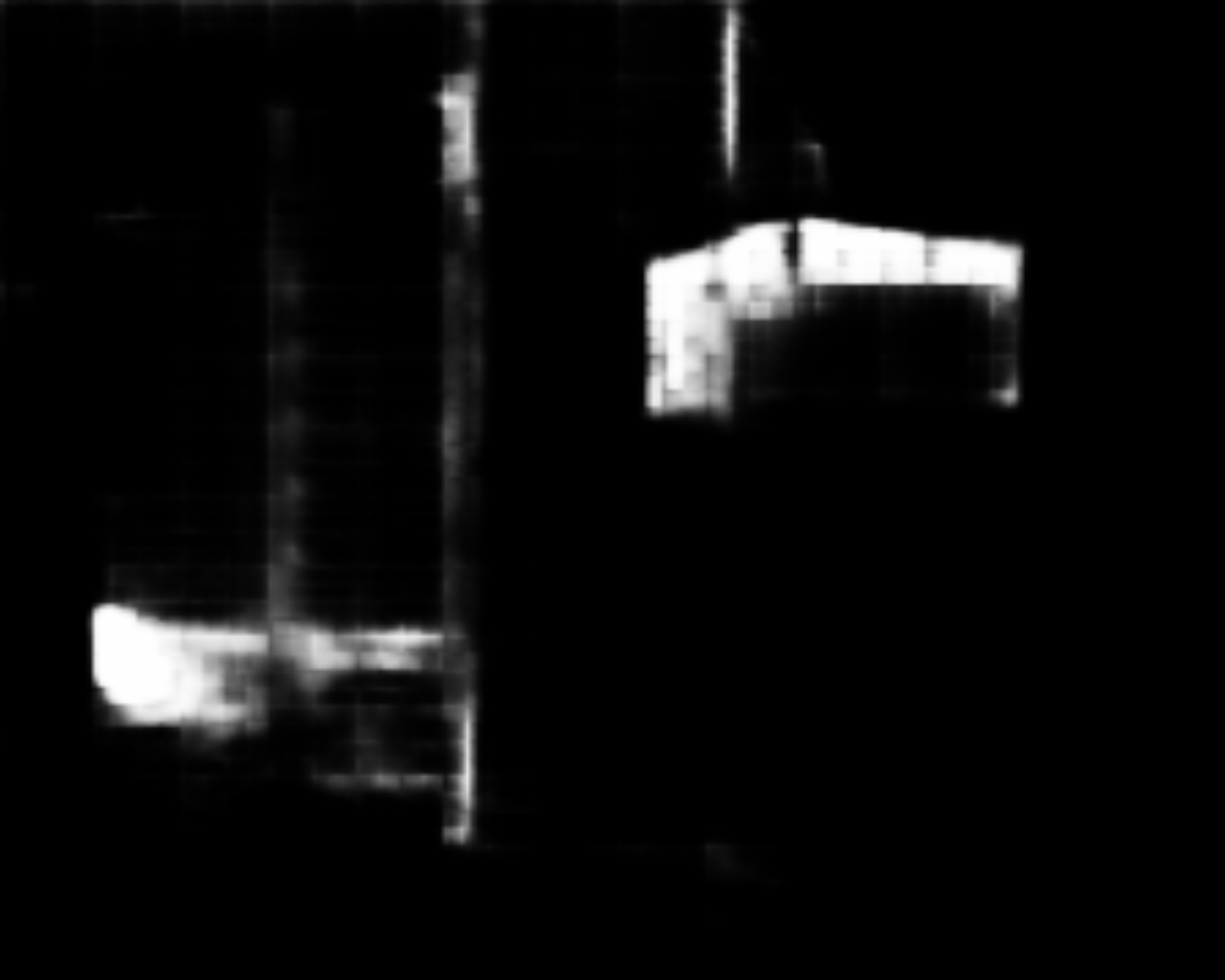} \hspace{-4mm} &
        \includegraphics[width=0.08\textwidth]{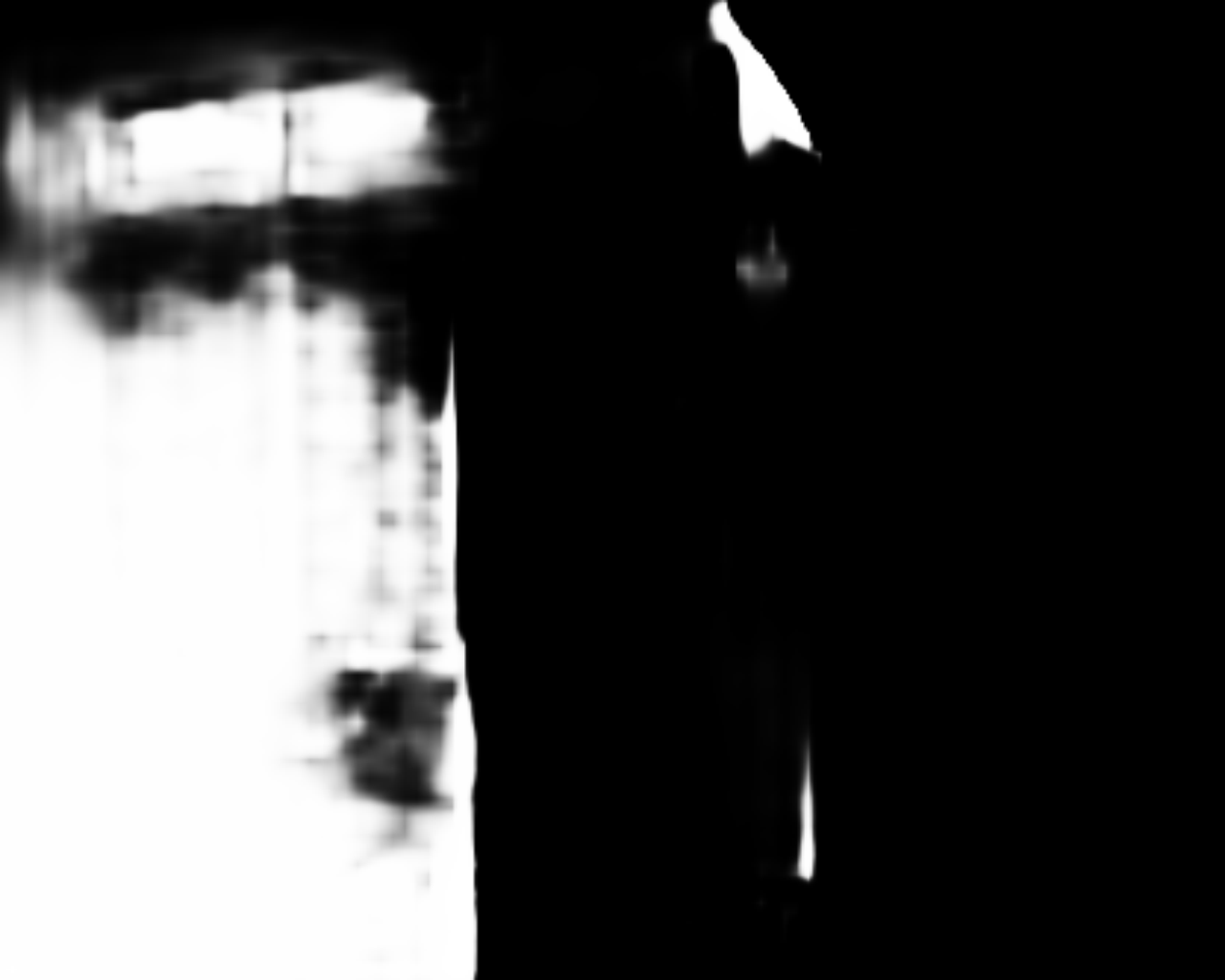} \hspace{-4mm} &
        \includegraphics[width=0.08\textwidth]{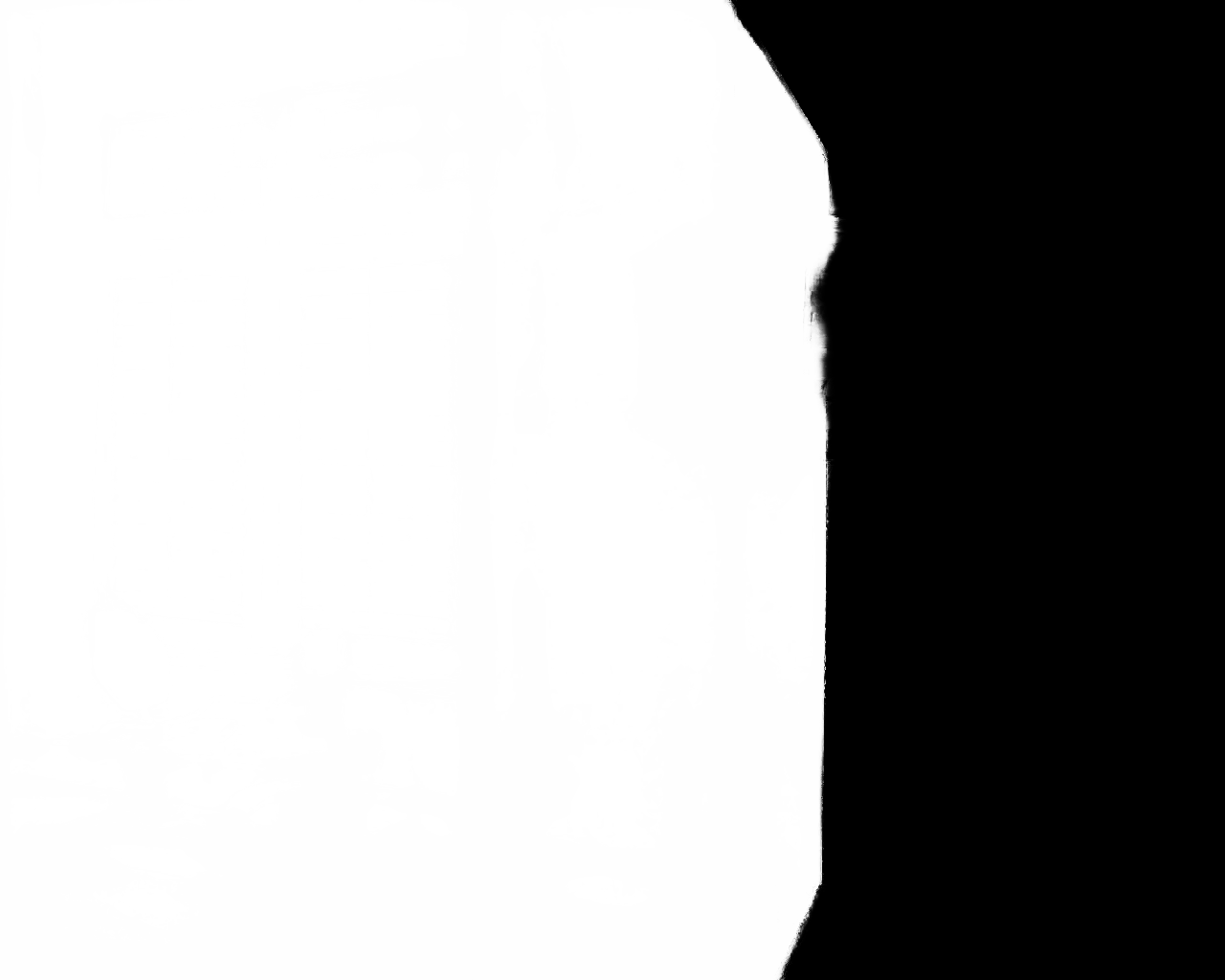} \hspace{-4mm} &
        \includegraphics[width=0.08\textwidth]{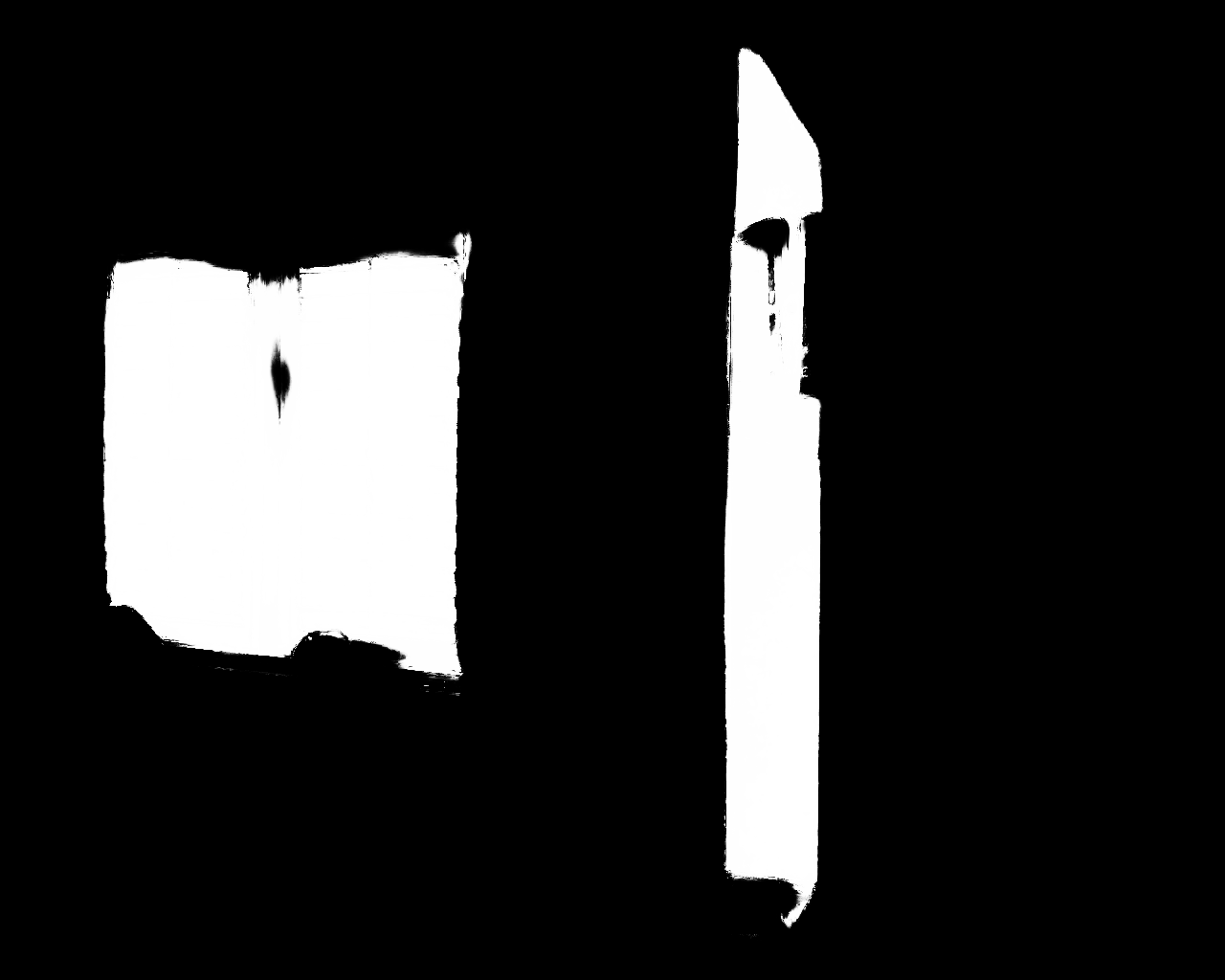} \hspace{-4mm} &
        \includegraphics[width=0.08\textwidth]{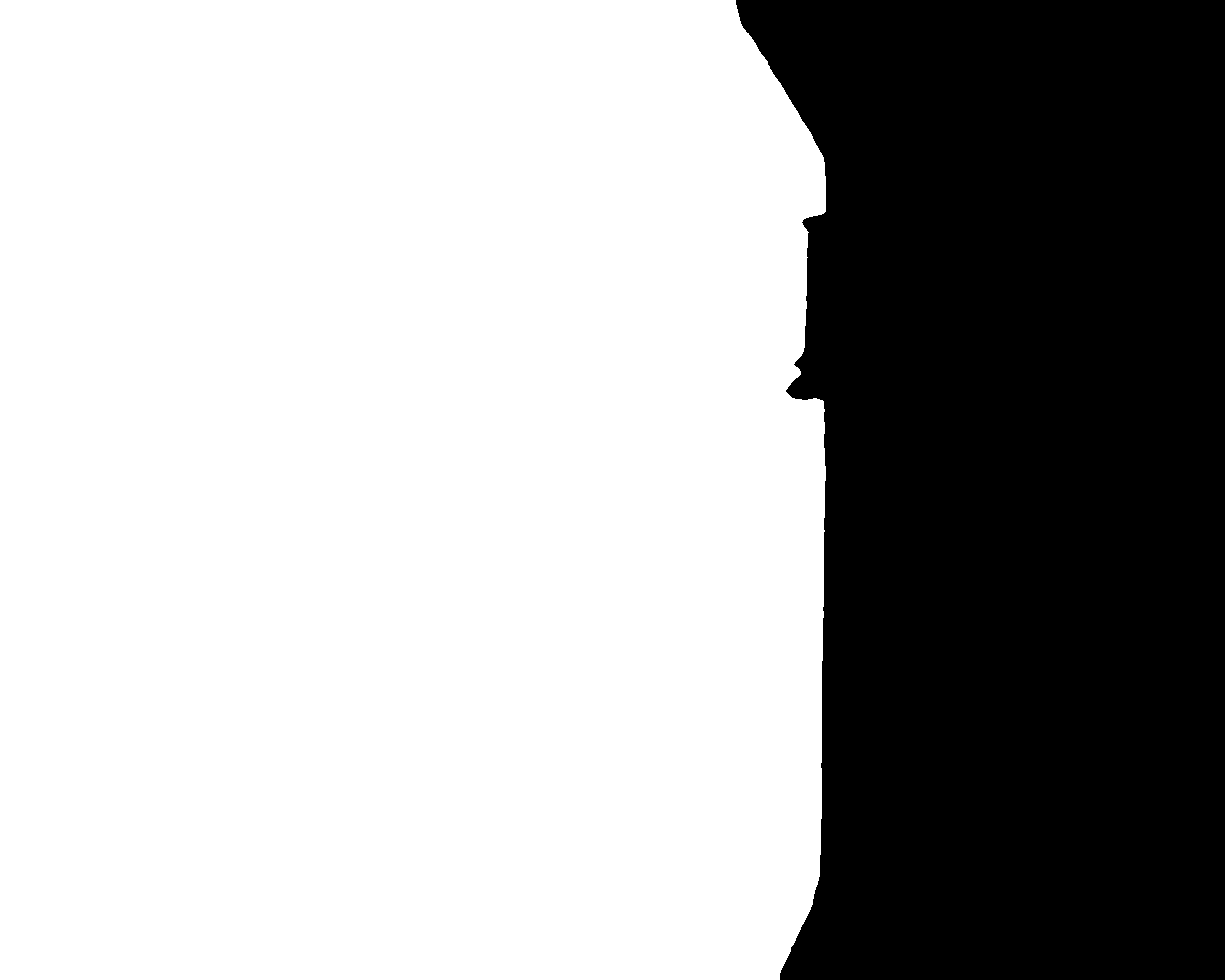} \hspace{-4mm} &
        \includegraphics[width=0.08\textwidth]{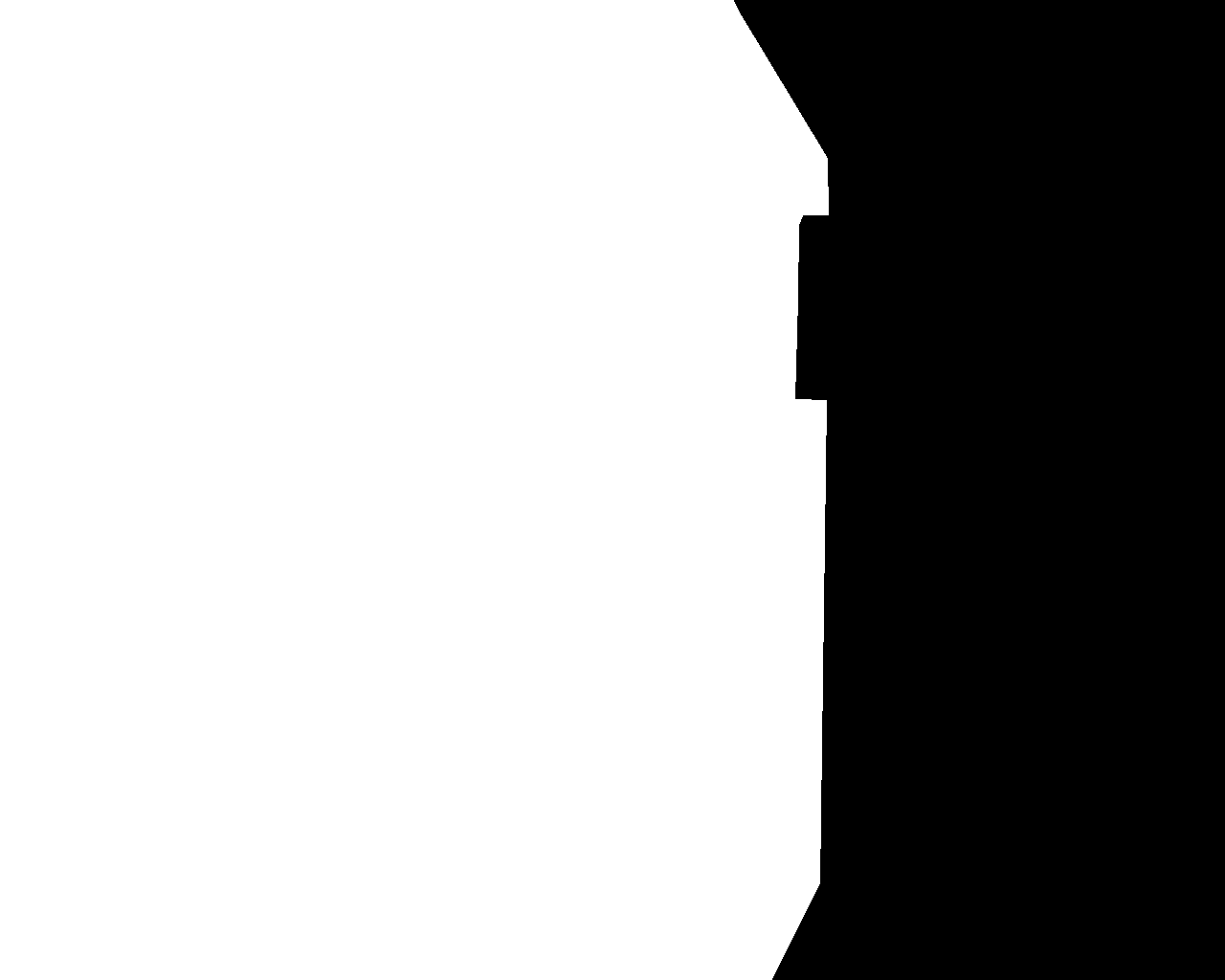} \\
        \includegraphics[width=0.08\textwidth]{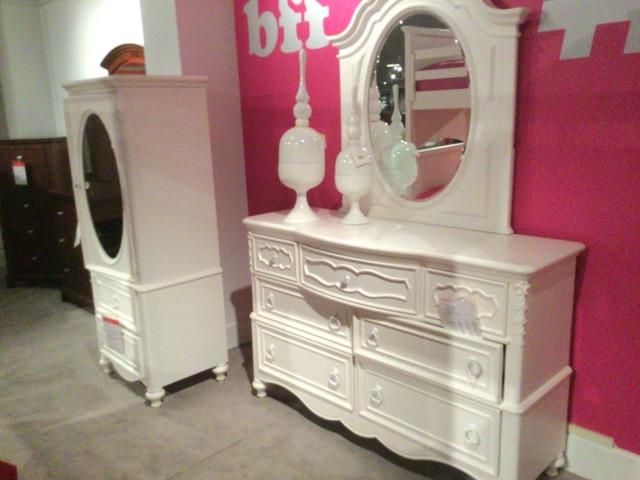} \hspace{-4mm} &
        \includegraphics[width=0.08\textwidth]{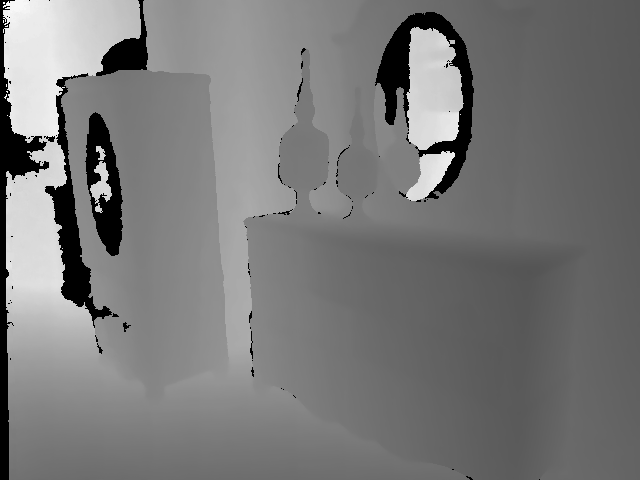} \hspace{-4mm} &
        \includegraphics[width=0.08\textwidth]{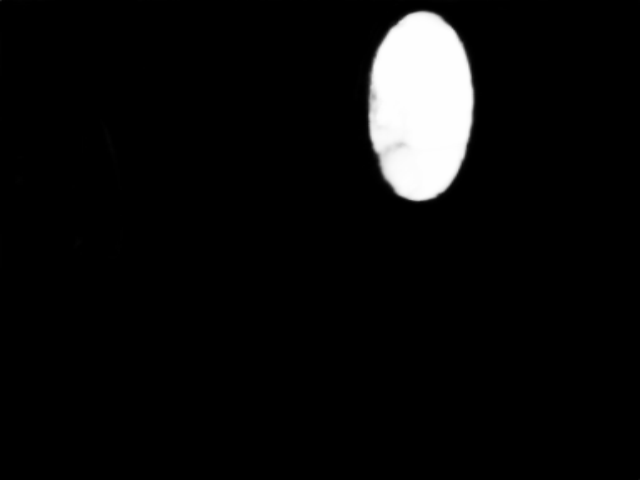} \hspace{-4mm} &
        \includegraphics[width=0.08\textwidth]{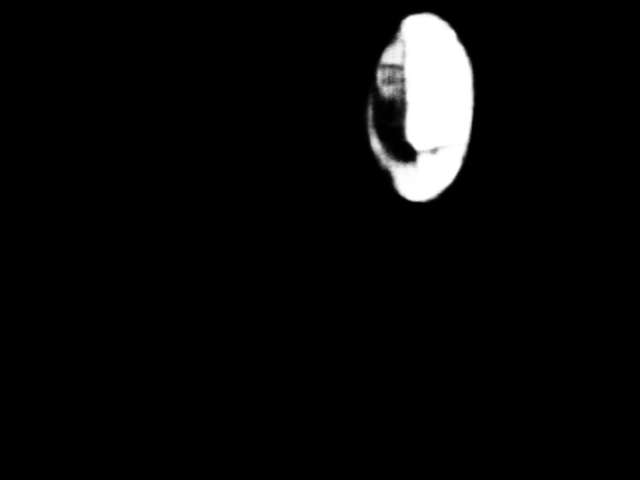} \hspace{-4mm} &
        \includegraphics[width=0.08\textwidth]{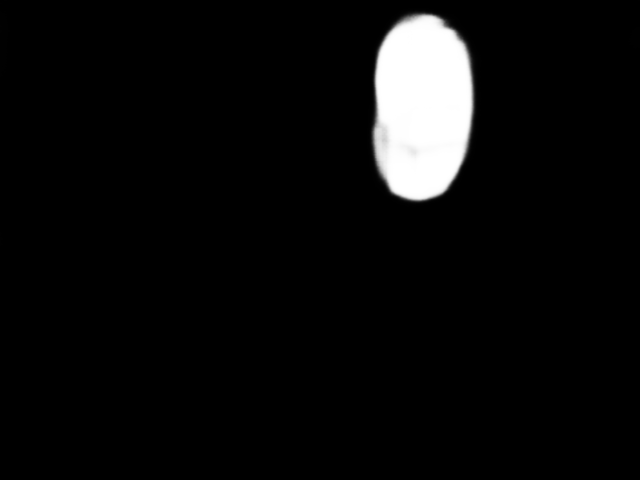} \hspace{-4mm} &
        \includegraphics[width=0.08\textwidth]{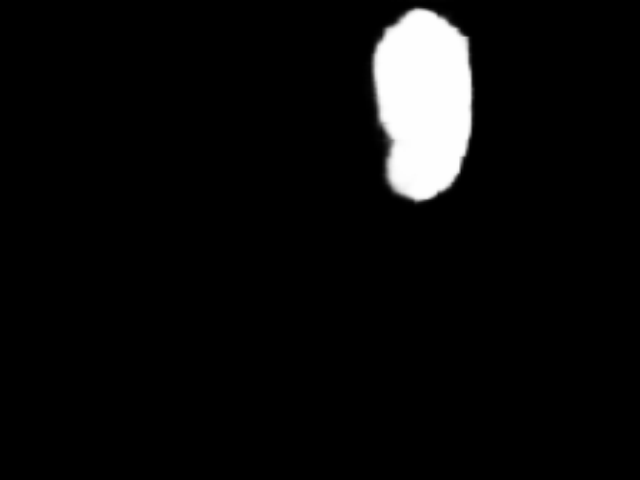} \hspace{-4mm} &
        \includegraphics[width=0.08\textwidth]{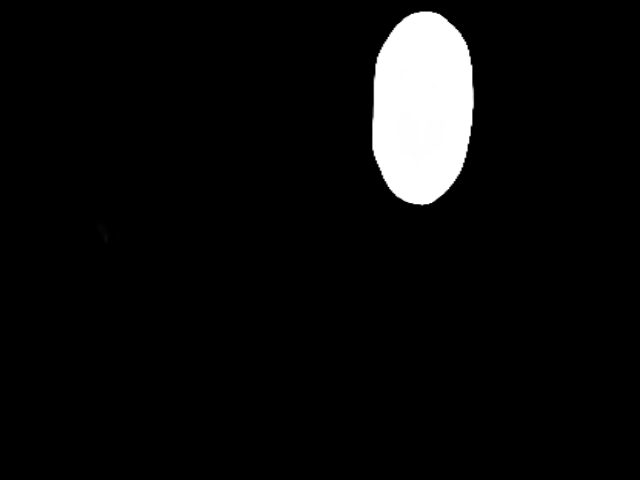} \hspace{-4mm} &
        \includegraphics[width=0.08\textwidth]{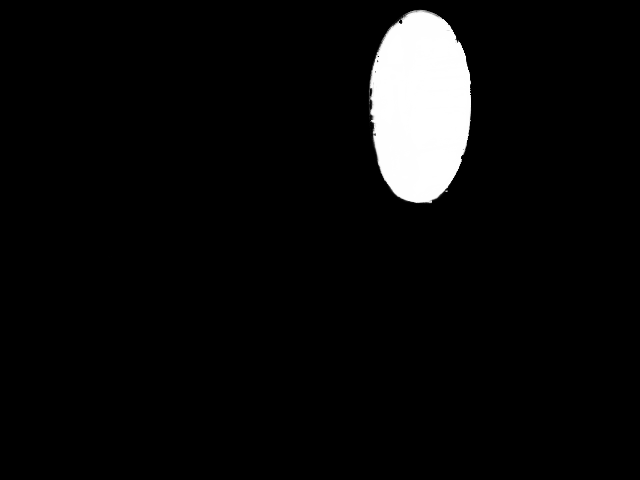} \hspace{-4mm} &
        \includegraphics[width=0.08\textwidth]{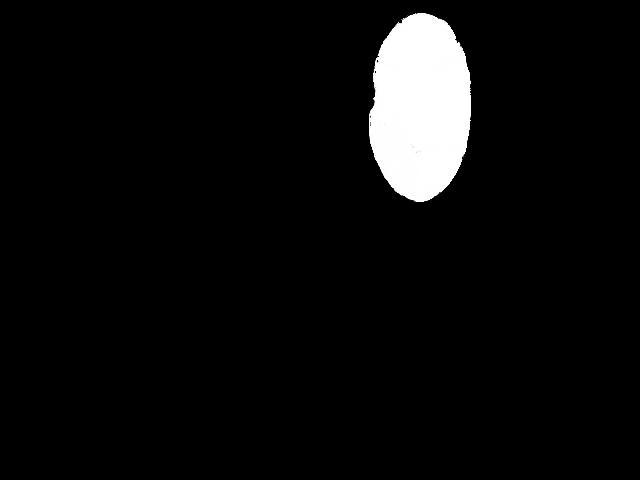} \hspace{-4mm} &
        \includegraphics[width=0.08\textwidth]{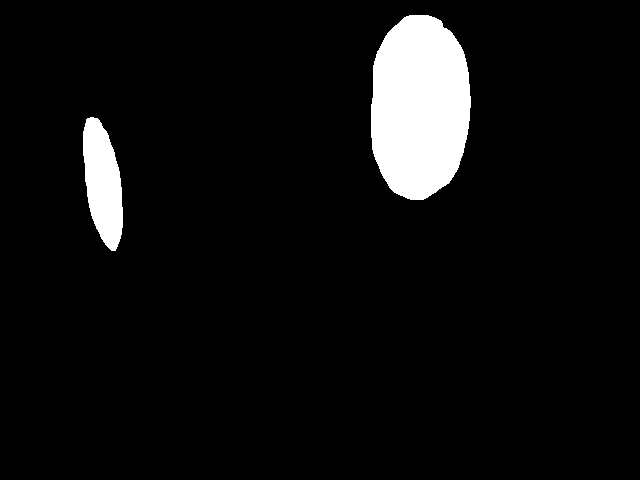} \hspace{-4mm} &
        \includegraphics[width=0.08\textwidth]{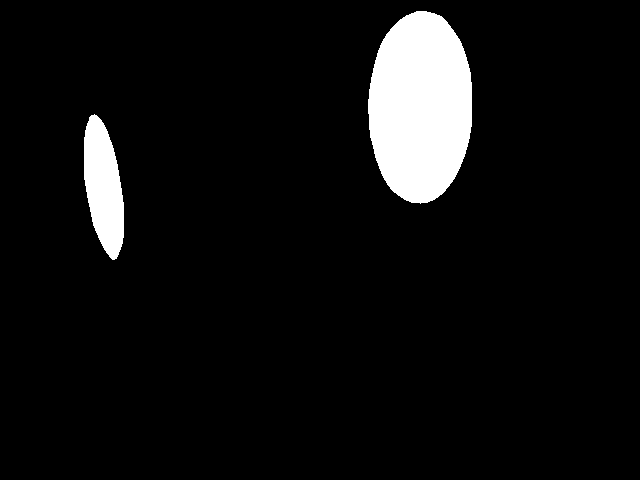} \\
        \includegraphics[width=0.08\textwidth]{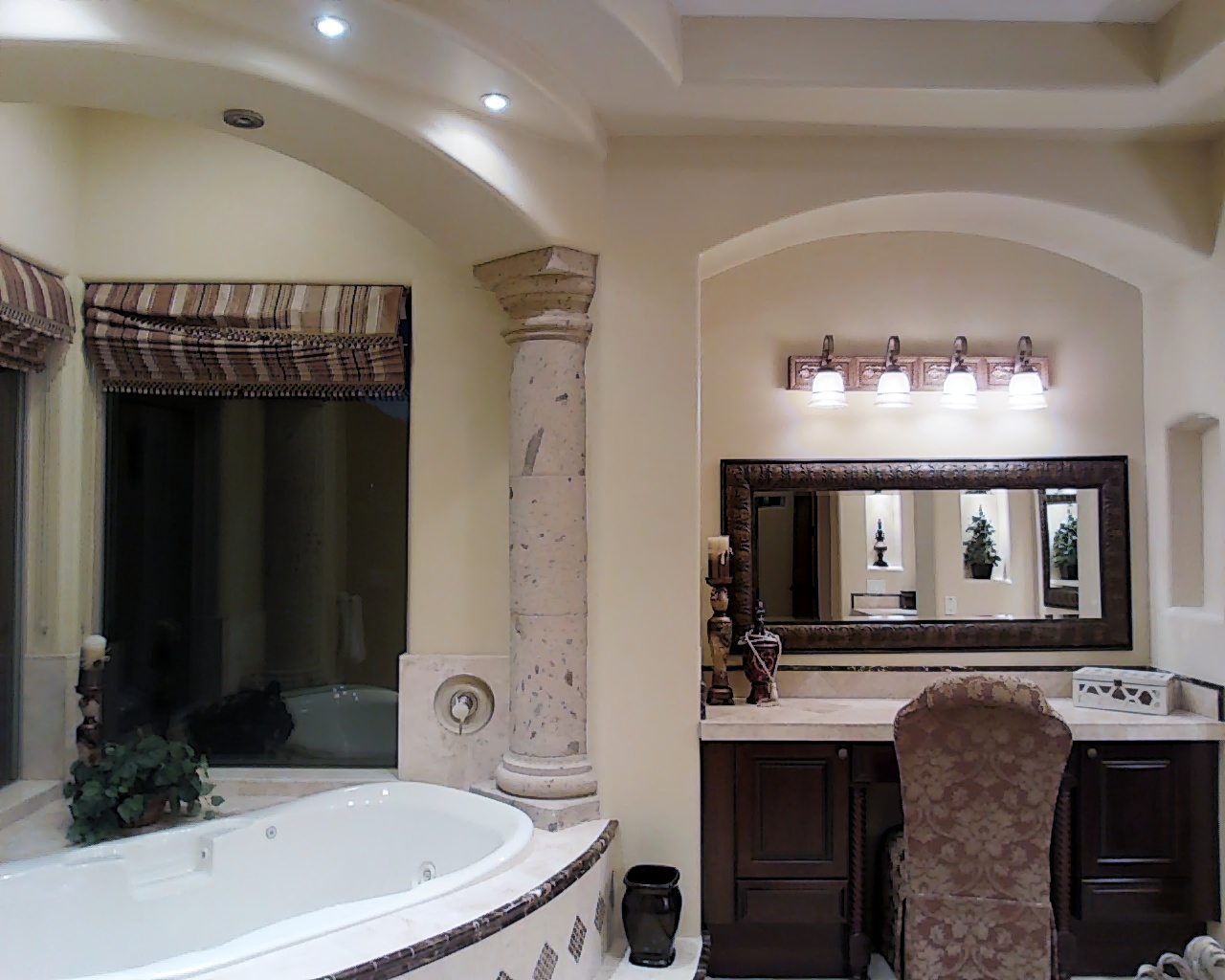} \hspace{-4mm} &
        \includegraphics[width=0.08\textwidth]{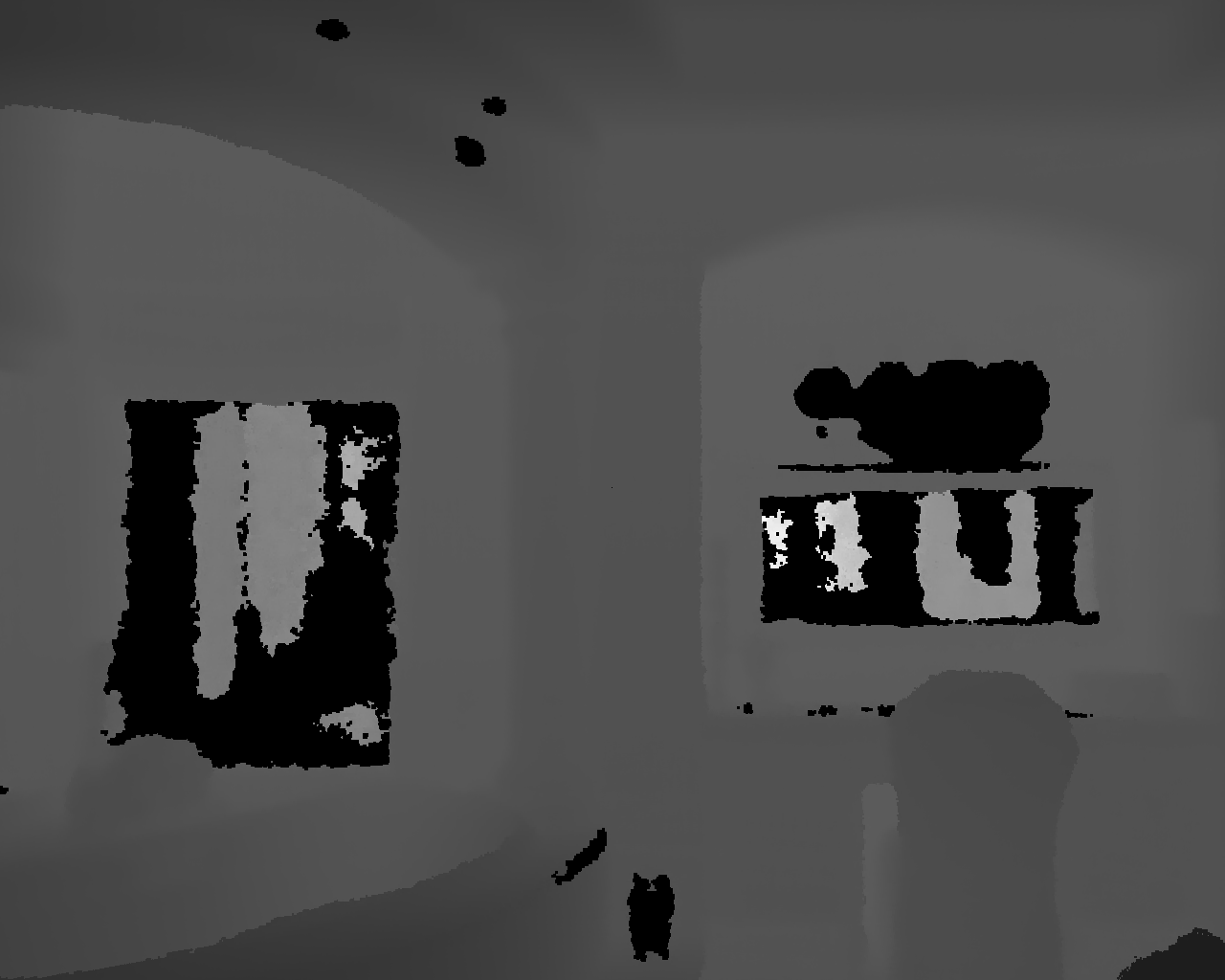} \hspace{-4mm} &
        \includegraphics[width=0.08\textwidth]{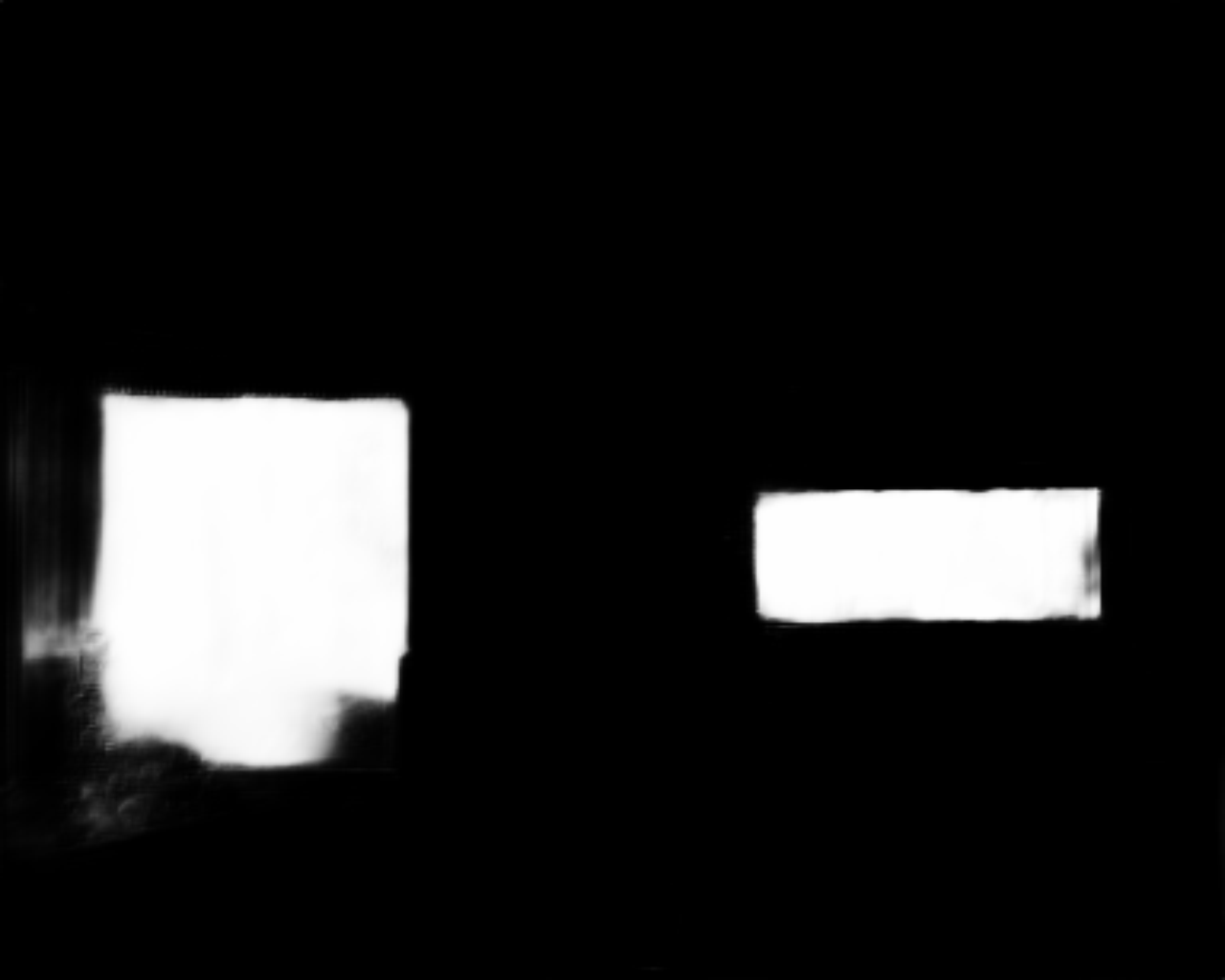} \hspace{-4mm} &
        \includegraphics[width=0.08\textwidth]{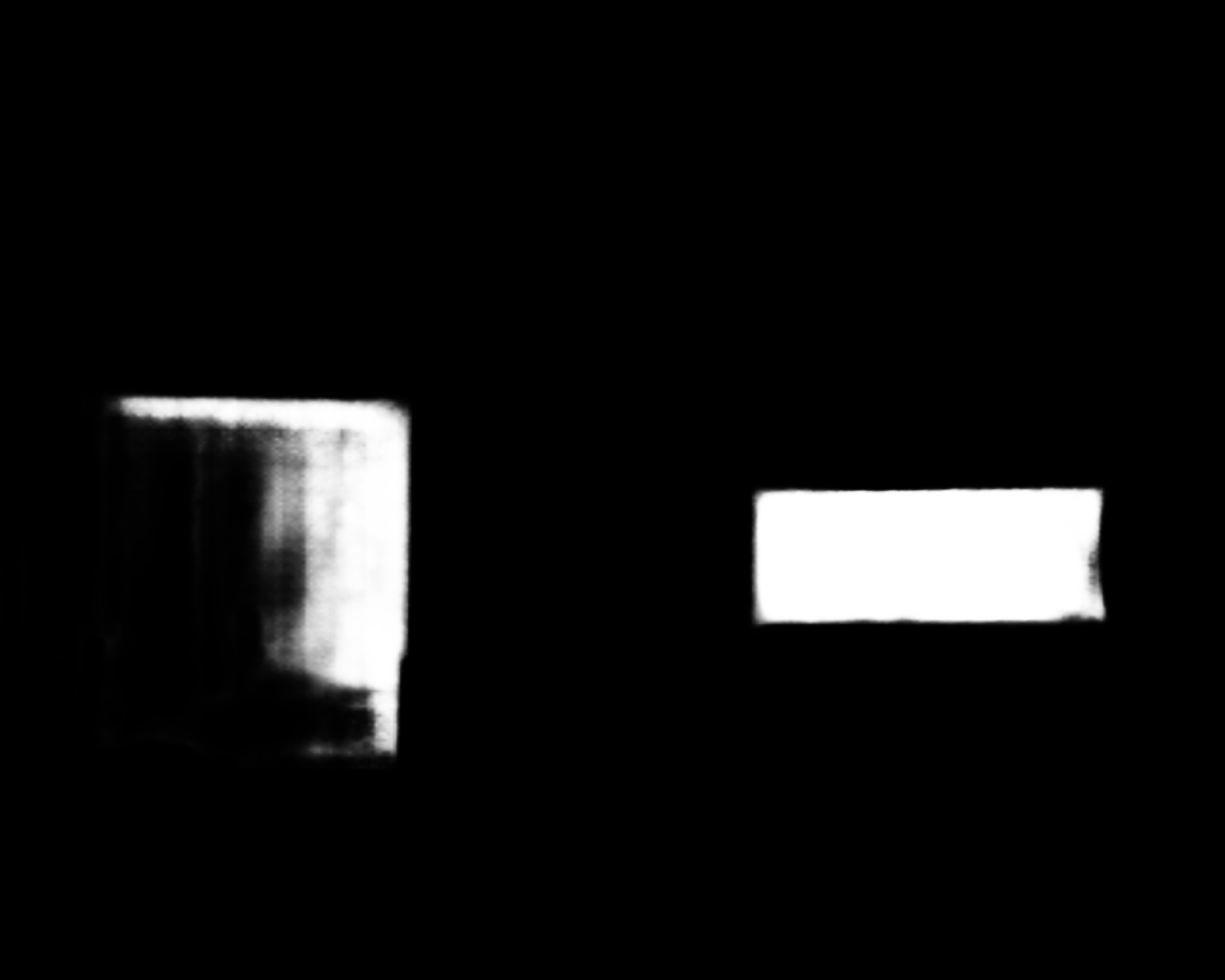} \hspace{-4mm} &
        \includegraphics[width=0.08\textwidth]{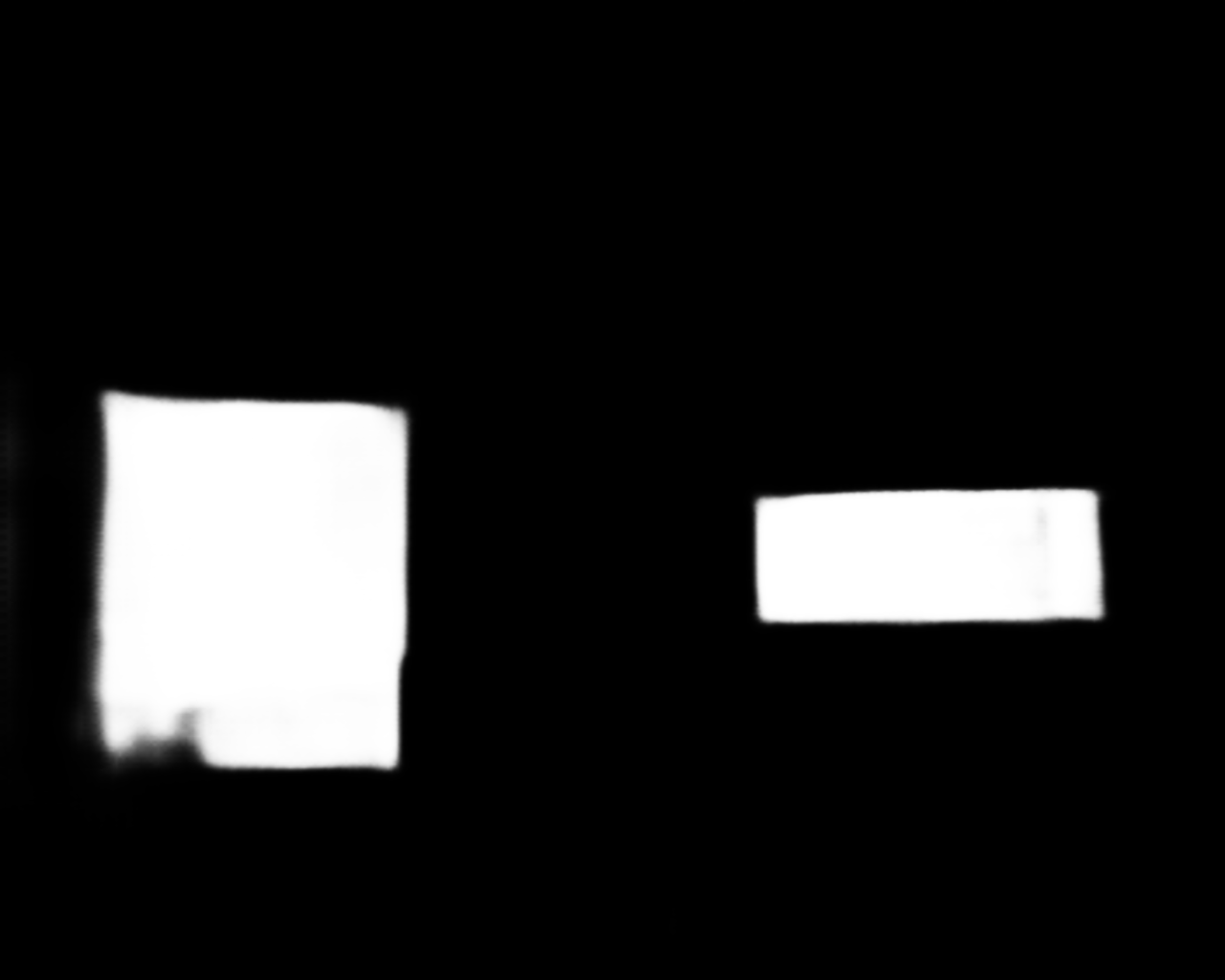} \hspace{-4mm} &
        \includegraphics[width=0.08\textwidth]{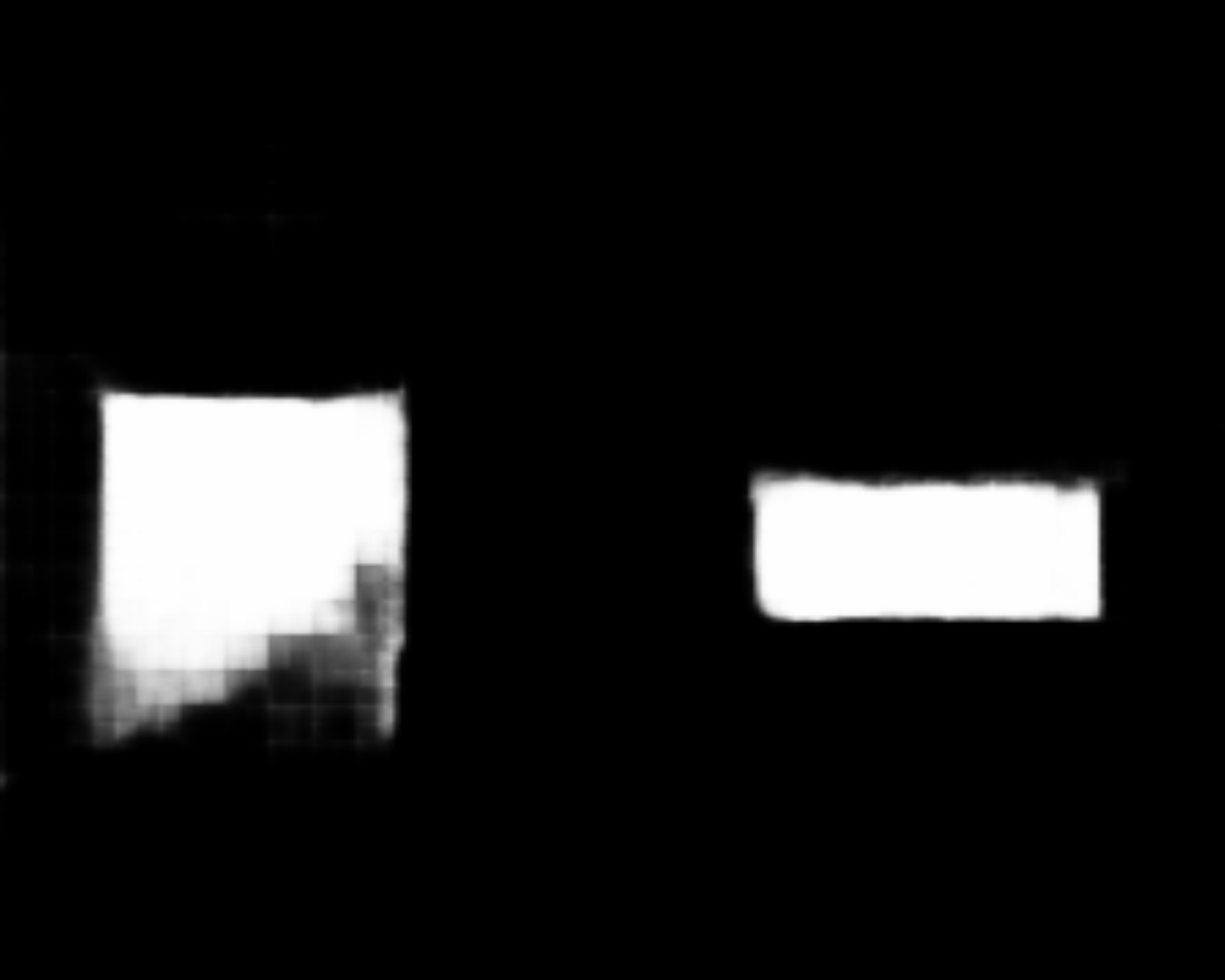} \hspace{-4mm} &
        \includegraphics[width=0.08\textwidth]{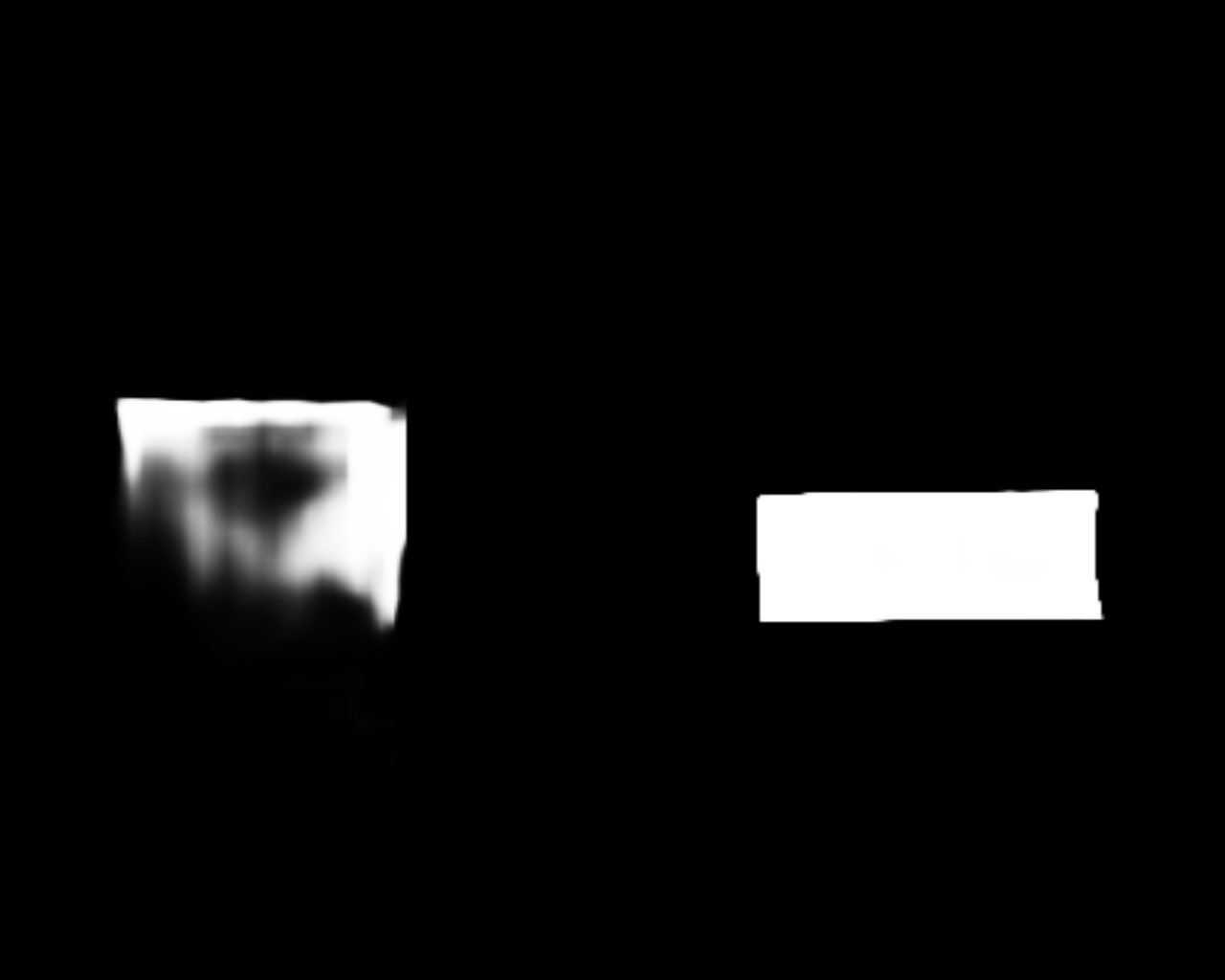} \hspace{-4mm} &
        \includegraphics[width=0.08\textwidth]{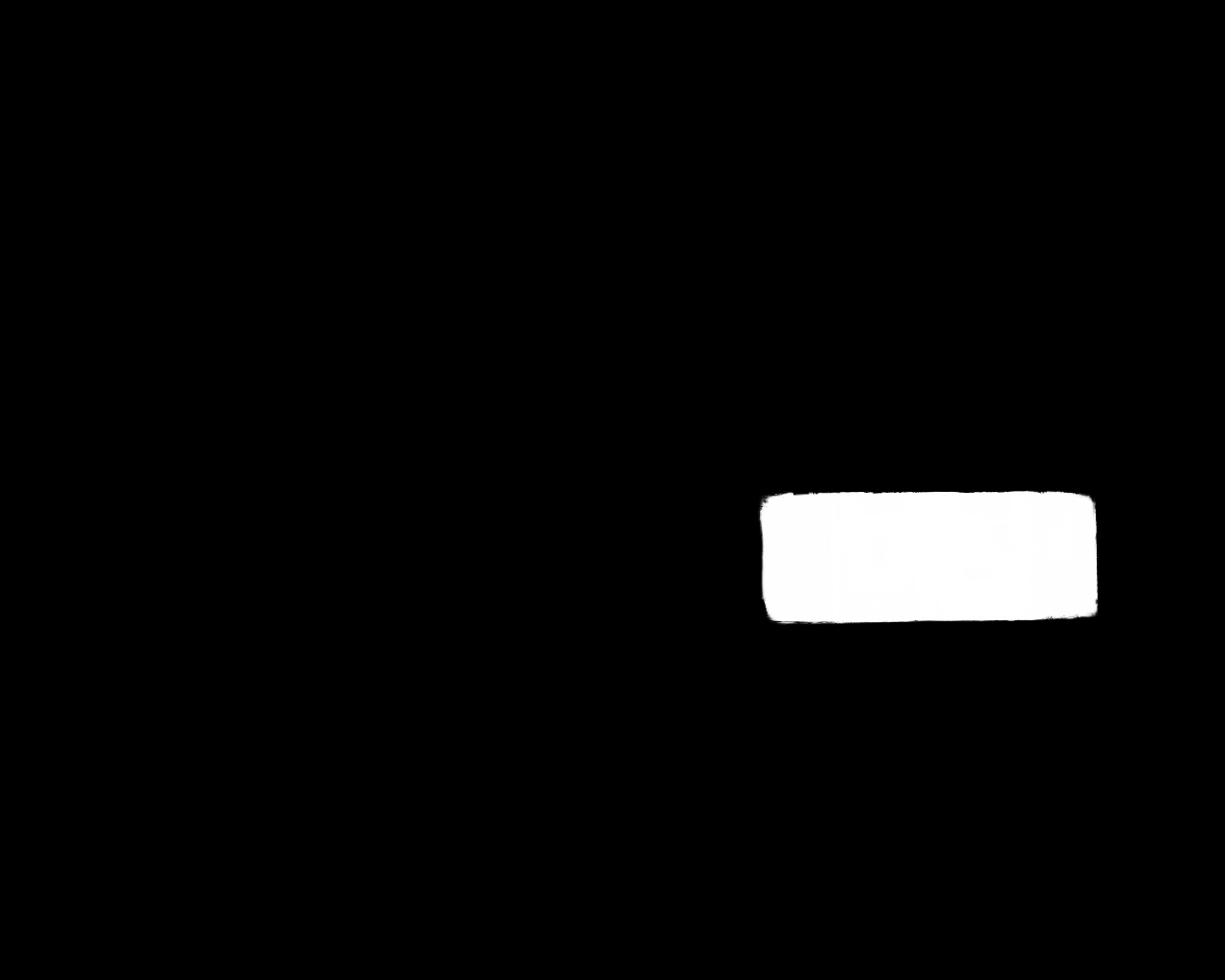} \hspace{-4mm} &
        \includegraphics[width=0.08\textwidth]{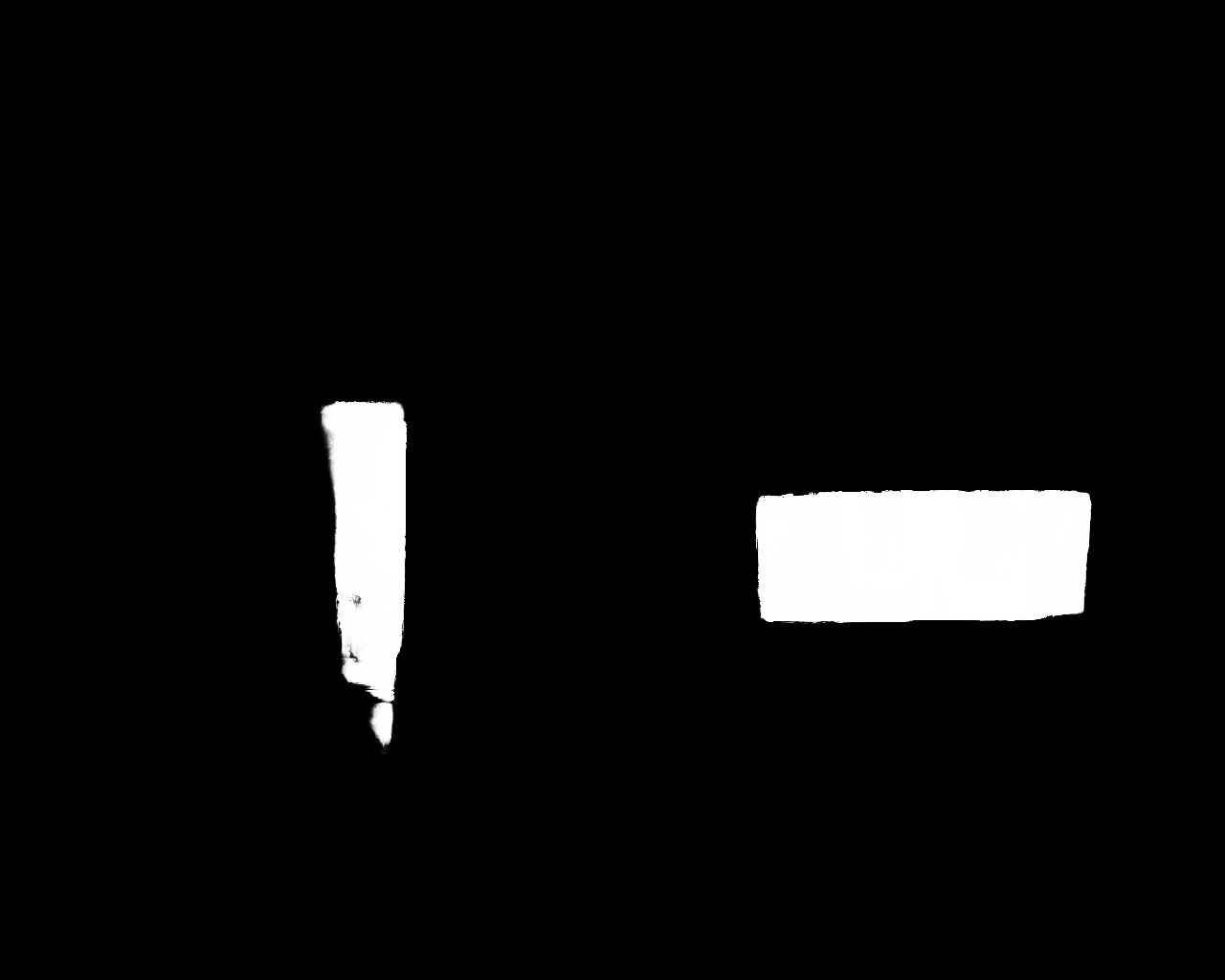} \hspace{-4mm} &
        \includegraphics[width=0.08\textwidth]{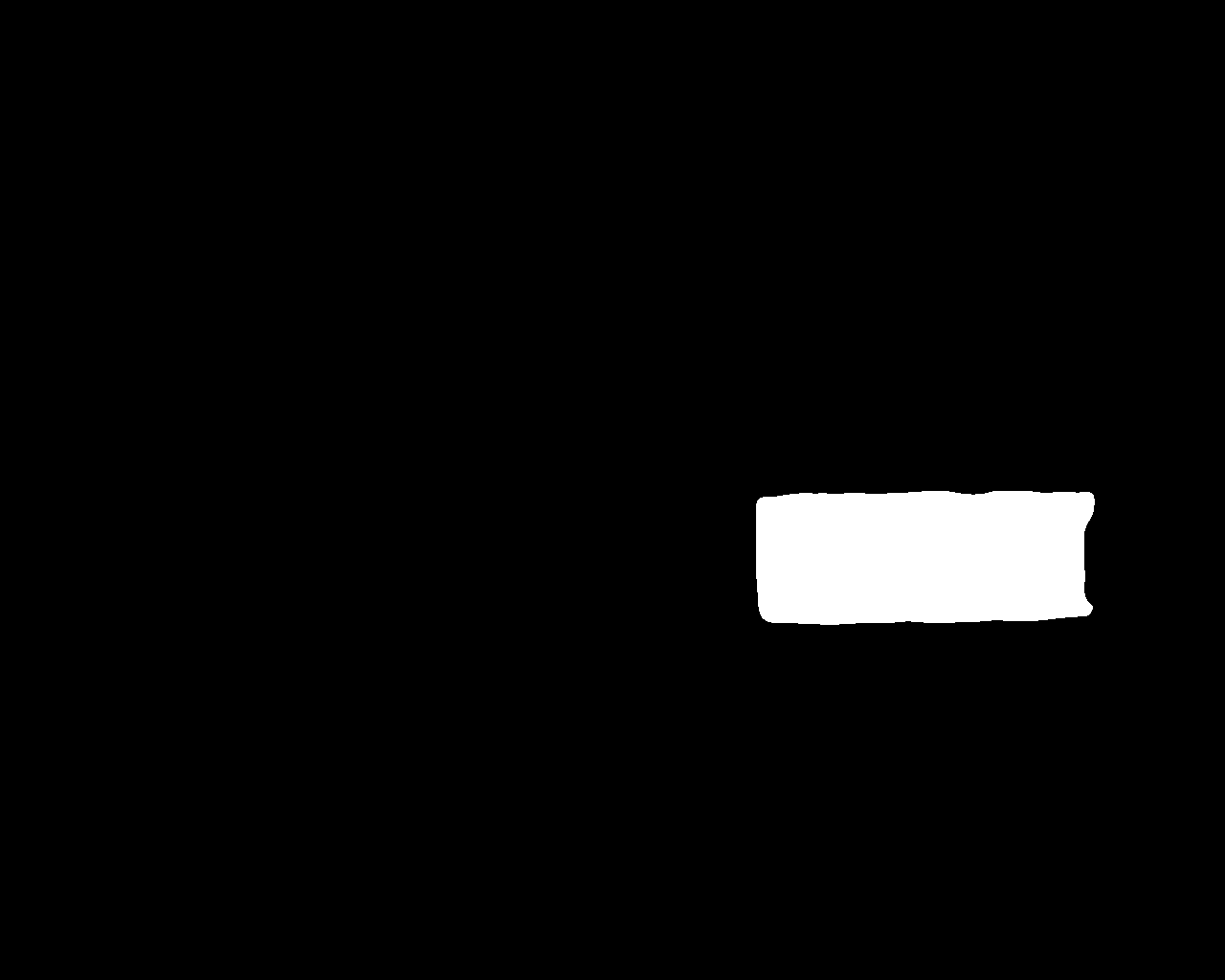} \hspace{-4mm} &
        \includegraphics[width=0.08\textwidth]{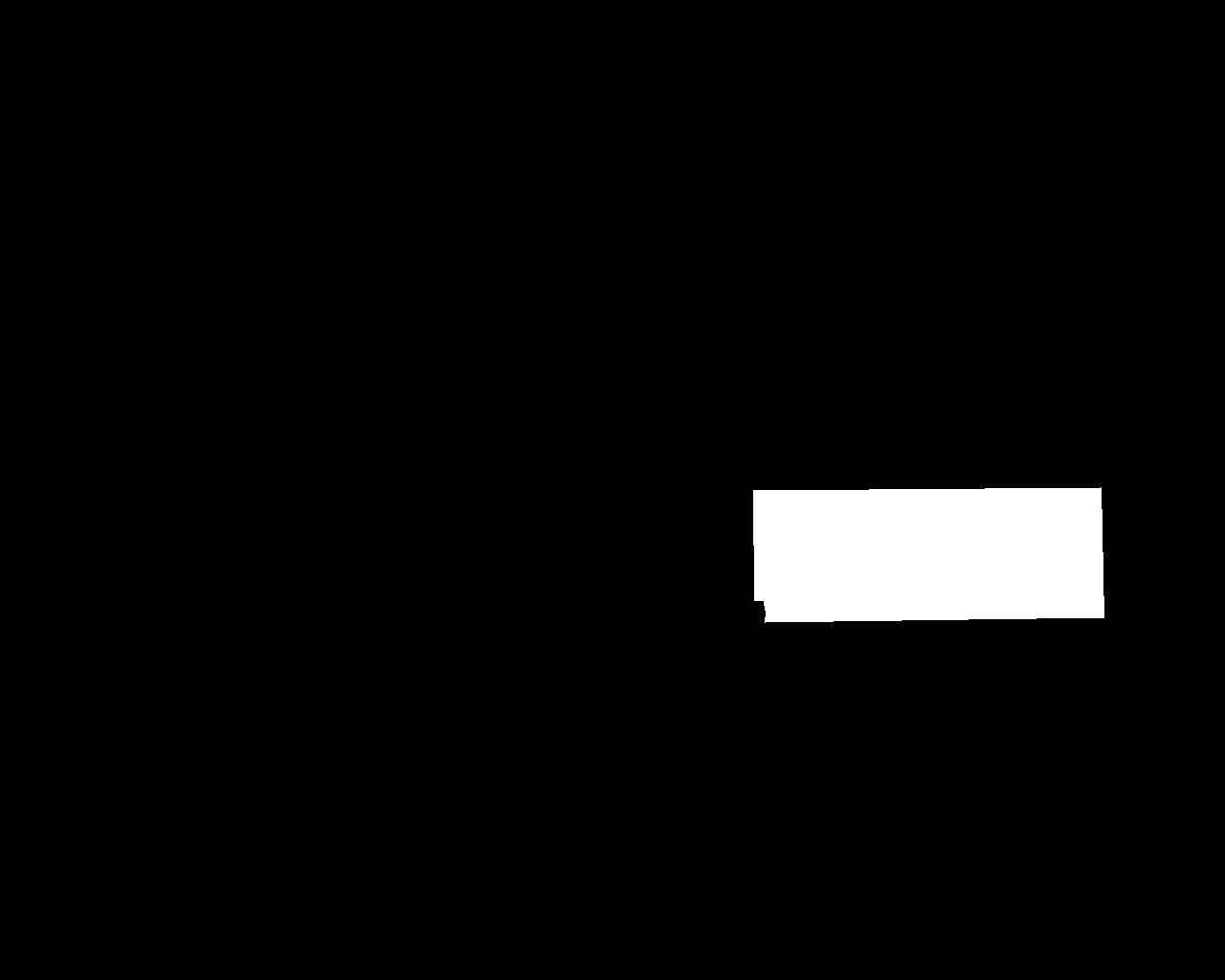} \\
        Image \hspace{-4mm} & Detph \hspace{-4mm} & DANet \hspace{-4mm} & JL-DCF \hspace{-4mm} & BBSNet \hspace{-4mm} & VST \hspace{-4mm} & PDNet \hspace{-4mm} & SANet \hspace{-4mm} & VCNet \hspace{-4mm} & Ours \hspace{-4mm} & GT
    \end{tabular}
    \caption{Visualization results on RGB-D dataset. In the first row, changes in depth can easily affect the judgement of RGB-D methods. The second row contains a pair of symmetric objects inside and outside mirrors. The third row represents mirrors that can hardly be recognized. And the last row is a scene including both glasses and mirrors.}
    \label{fig:rgbd}
\end{figure*}

Our method is also compared with 4 RGB-D salient object detection methods 
JL-DCF~\cite{Fu_2020_CVPR}, DANet~\cite{zhao2020single}, BBSNet~\cite{fan2020bbs} and VST~\cite{Liu_2021_ICCV}, and 3 mirror detection methods PDNet~\cite{mei2021depth}, SANet~\cite{guan2022learning} and VCNet~\cite{tan2022mirror} on the RGBD-Mirror dataset.
As shown in Table~\ref{tab:rgbd}, our method does not leverage depth information, and can still achieve the best performance in terms of all the evaluation metrics.
Visualization results are shown in Fig.~\ref{fig:rgbd}.
%
%
In all the four examples, RGB-D methods are likely misled by depth information.
Especially in the first row, they wrongly judge the depth changes as the existence of mirrors.
In the second row, our method correctly detects the mirror region by exploiting the loose symmetry relationship between the television and its reflection, while some competing methods even fail to detect the correct side of the mirror.
In the third row, there is a mirror that can be easily missed.
All the competing methods ignore the left mirror, although the depth map has an obvious change in that area.
Our method can still discover the mirror as the scene in it have a kind of symmetry relation with the nearby cabinet.
In the last row, we note that our method does not mis-detect the glasses region as a mirror region, while the competing methods can hardly tell the subtle differences between mirrors and glasses.
Different from mirrors, glasses can transmit most of the lights, which weakens the reflection effects.
It shows that our method can identify corresponding reflection features from mirrors.
All the cases show that utilizing symmetry information can greatly benefit the performance of mirror detection, especially in complex scenes.

\subsection{Ablation study}
\begin{table}[t]
  \caption{Ablation study results on MSD. Swin-S denotes our baseline method, which is decoded by UperNet. Dual-Path denotes the dual-path Swin Transformer. SAAM denotes our Symmetry-Aware Attention Module on scale 3. SAAMs denotes SAAM on both scale 2 and scale 3. CFDM denotes our Contrast and Fusion Decoder Module.}
  \centering
  \begin{tabular}{l|c c c}
    \hline
    Method & $IoU\uparrow$ & $F_\beta\uparrow$ & $MAE\downarrow$ \\
    \hline
    Baseline & 80.46 & 0.901 & 0.045 \\
    Dual-Path & 79.59 & 0.903 & 0.044 \\
    Dual-Path + SAAM & 80.01 & 0.918 & 0.042 \\
    Dual-Path + SAAMs & 80.03 & 0.903 & 0.043 \\
    Dual-Path + CFDM & 81.98 & 0.918 & 0.039 \\
    Dual-Path + SAAM + CFDM & 82.96 & 0.911 & 0.039 \\
    SATNet(Ours) & \textbf{85.41} & \textbf{0.922} & \textbf{0.033} \\
    \hline
  \end{tabular}
  \label{tab:ablation}
\end{table}

\noindent\textbf{Benefits of Dual-Path Structure. }
To better analyze the benefits of our dual-path structure, we conduct two experiments:
One is a pure Swin Transformer decoded by UperNet~\cite{xiao2018unified} (1st row); the other is a dual-path Swin Transformer, where features are trained and supervised separately in two paths (2nd row).
Results in the first two rows show that, with extra features and supervision, the second method has no clear advantage when compared against the first one.
That is to say, we cannot simply attribute the improvement of our method to the extra features we extract.
Albeit we introduce the dual-path structure to enhance the symmetry semantics,
the extra features are more like a repeated computation of the original ones if there are no appropriate fusing and matching mechanisms for the two paths.

\begin{figure}
    \centering
    \begin{tabular}{cccc}
        \includegraphics[width=0.085\textwidth]{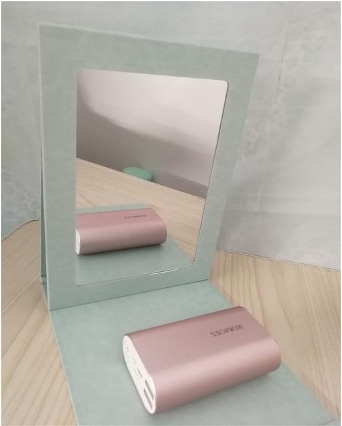} \hspace{-4mm} &
        \includegraphics[width=0.085\textwidth]{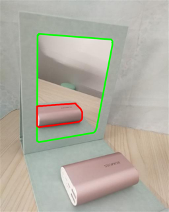} \hspace{-4mm} &
        \includegraphics[width=0.085\textwidth]{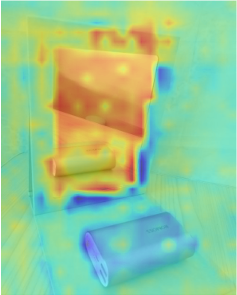} \hspace{-4mm} &
        \includegraphics[width=0.085\textwidth]{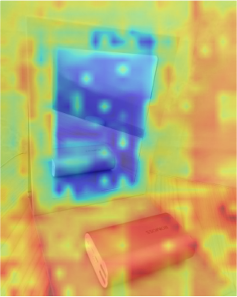} \\
        (a) \hspace{-4mm} & (b) \hspace{-4mm} & (c) \hspace{-4mm} & (d)
    \end{tabular}
    \caption{Visualization of attention maps in SAAM. (a)-(d) denote image, region of interest, attention of mirrors, and attention of objects in mirrors. While attention of the mirror region focuses on the mirror itself, attention of the power bank in mirrors lies in the corresponding real object.}
    \label{fig:attn_map}
\end{figure}

\noindent\textbf{Effect of SAAM. }
To evaluate the effect of our attention module, we conduct another two experiments:
One is a dual-path Swin-S with a SAAM in the highest level (3-rd row), and the other is the same structure, but with SAAMs in the highest two levels (4-th row).
Comparing the third row with the second row, we discover that ``Dual-Path + SAAM" gets better results in all the three metrics, which is reasonable as our SAAM models symmetry relationships in high-level features.
However, $F_\beta$ in the fourth row drops back to 0.903, indicating that directly applying SAAM to features in lower levels may not work well.
We further visualize the attention map in SAAM. In Fig.~\ref{fig:attn_map}(c), the mirror region (green contour in (b)) of the attention map focuses on the mirror itself.
While in Fig.~\ref{fig:attn_map}(d), the highest attention signal of the power bank region inside the mirror (red contour in (b)) is located on the real power bank in the image.
This observation supports that SAAM is able to model loose symmetry relations.

\noindent\textbf{Effect of CFDM. }
In the fifth row, we conduct an experiment based on the second row, replacing the UperNet decoder with our CFDM.
Comparing results of the two rows, "Dual-Path + CFDM" have a gain of 2.39\%, 1.50\%, and 0.5\% in $IoU$, $F_\beta$, and $MAE$, respectively.
The improvement proves that our decoder module can properly fuse features in the two paths, and is more suitable for the mirror detection task.

\noindent\textbf{Combination of SAAM and CFDM. }
To explore the best way to combine our SAAM and CFDM, we conduct two experiments:
one is a dual-path Swin Transformer with a SAAM in the highest level and CFDMs as the decoder (6th row), and the other is our final network SATNet, which has SAAMs in the highest two levels (last row).
Analyzing the last three rows, we conclude that applying SAAM before CFDM is effective as the three evaluation metrics have progressively improved to 85.41\%. 0.922 and 0.033.
On the other hand, comparing the network in the fourth row with SATNet, the improvement from ``Dual-Path + SAAMs" to SATNet is even larger, which means our CFDM contributes to the fusion of dual-path features, especially for symmetry semantics in high levels.

\begin{figure}
    \centering
    \begin{tabular}{cccccc}
        \includegraphics[width=0.07\textwidth]{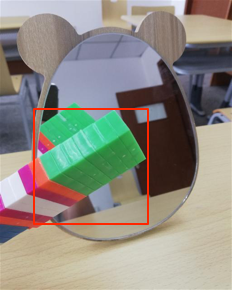} \hspace{-4mm} &
        \includegraphics[width=0.07\textwidth]{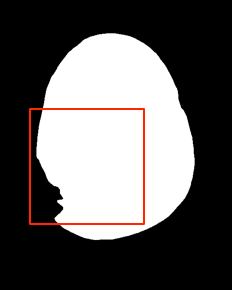} \hspace{-4mm} &
        \includegraphics[width=0.07\textwidth]{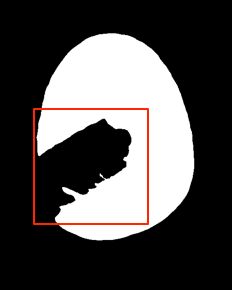} \hspace{-4mm} &
        \includegraphics[width=0.07\textwidth]{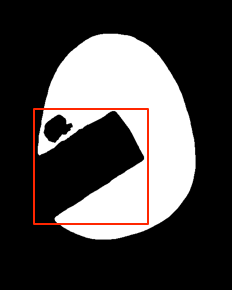} \hspace{-4mm} &
        \includegraphics[width=0.07\textwidth]{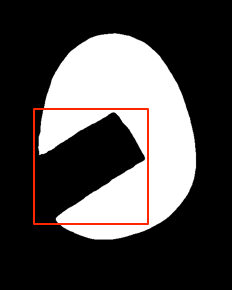} \hspace{-4mm} &
        \includegraphics[width=0.07\textwidth]{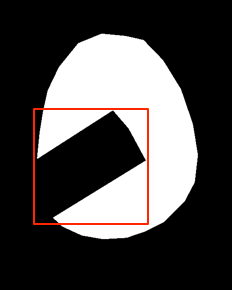} \\
        Image \hspace{-4mm} & Baseline \hspace{-4mm} & +SAAM \hspace{-4mm} & +CFDM \hspace{-4mm} & SatNet \hspace{-4mm} & GT
    \end{tabular}
    \caption{Visualization results for the ablation study. In this example, our baseline Swin-S cannot perceive the symmetry relationship. The network embedded with SAAM does not outline a precise boundary. And when adding CFDM, the network is still confused about the symmetry relationship. Only SATNet can correctly detect the mirror region.}
    \label{fig:vis4}
\end{figure}

\noindent\textbf{Visualization results for the ablation study. }
To further analyze the effectiveness of each component, we visualize the prediction results of Swin-S, Dual-Path + SAAM, Dual-Path + CFDM, and SATNet in Fig.~\ref{fig:vis4}.
Swin-S can provide the approximate location of the mirror, but is not sensitive to symmetry relationships, which demonstrates that current baselines can hardly model loose symmetry relationships.
Equipped with our attention module SAAM, the network can exclude the real-world object which shades the mirror from the mirror region, showing the ability of perceiving symmetry relationships.
However, its prediction map is not precise enough, especially near the boundary of mirrors.
In comparison to our SAAM, our decoder module CFDM refines mirror boundaries well, but it wrongly excludes the symmetry area in the mirror region.
Analogous to Swin-S, it cannot handle symmetry relationships well.
Only with both two modules, SATNet marks the mirror region correctly.
The visualization result is basically consistent with the corresponding effects of the components we expect.

\section{Conclusion}
In this paper, we proposed a dual-path  Symmetry-Aware Transformer-based mirror detection network (SATNet) for better mirror detection.
We presented a new perspective on detecting mirrors by leveraging loose symmetry relationships.
Then, we suggested a novel dual-path network, introducing a transformer pipeline to enhance the ability of long-range dependencies understanding for mirror detection.
Furthermore, we proposed the Symmetry-Aware Attention Module (SAAM) to aggregate better feature representation of symmetry relations, while exploiting Contrast and Fusion Decoder Module (CFDM) to generate refined prediction maps progressively.
Experimental results on multiple datasets demonstrate the benefit of loose symmetry relationships in mirror detection. 
Our network can effectively model such relationships and greatly improve the performance of mirror detection in comparison to state-of-the-arts.

\bibliography{aaai23}
\end{document}